\documentclass[10pt,journal,compsoc]{IEEEtran}

\ifCLASSOPTIONcompsoc
  \usepackage[nocompress]{cite}
\else
\usepackage{cite}
\fi
\usepackage{booktabs} 
\usepackage{xcolor}
\usepackage{soul}
\usepackage{epsfig}
\usepackage{graphicx}
\usepackage{amsmath}
\usepackage{amssymb}
\usepackage{courier}
\usepackage{helvet}
\usepackage{courier}
\usepackage{algorithm}
\usepackage{algorithmic}
\usepackage{multirow}
\usepackage{url}
\usepackage{booktabs}
\usepackage{array}
\usepackage{paralist,algorithmic,algorithm}
\usepackage{balance}
\usepackage[numbers]{natbib}

\newcommand{\x}{{\bf x}}

\newcommand{\y}{{\bf y}}

\newcommand{\D}{\mathcal{D}}

\newcommand{\R}{\mathbb{R}}

\newtheorem{defn}{Definition}

\newcolumntype{L}[1]{>{\raggedright\let\newline\\\arraybackslash\hspace{0pt}}m{#1}}

\ifCLASSINFOpdf
\else
\fi

\begin{document}
\title{Learning Adaptive Embedding Considering Incremental Class}

\author{Yang Yang,~\IEEEmembership{}
	    Zhen-Qiang Sun,~\IEEEmembership{}
	    HengShu Zhu,~\IEEEmembership{Senior Member, IEEE}
	    Yanjie Fu,~\IEEEmembership{}
        Hui Xiong,~\IEEEmembership{Senior Member, IEEE}
        and Jian Yang,~\IEEEmembership{Member, IEEE}
\IEEEcompsocitemizethanks{\IEEEcompsocthanksitem Yang Yang and Jian Yang are with the Nanjing University of Science and Technology, Nanjing 210094, China. E-mail: {yyang,csjyang}@njust.edu.cn}
\IEEEcompsocitemizethanks{\IEEEcompsocthanksitem Zhen-Qiang Sun is with the Nanjing Normal University, Nanjing 210023, China. E-mail: enderman19980125@outlook.com}
\IEEEcompsocitemizethanks{\IEEEcompsocthanksitem Hengshu Zhu is with Baidu Talent Intelligence Center, Baidu Inc, Beijing 100000, China. E-mail:zhuhengshu@baidu.com}
\IEEEcompsocitemizethanks{\IEEEcompsocthanksitem Yanjie Fu is with the Missouri University of Science and Technology, Rolla, MO 65401. E-mail: fuyan@mst.edu.}
\IEEEcompsocitemizethanks{\IEEEcompsocthanksitem  Hui Xiong is with the Management Science and Information Systems Department, Rutgers Business School, Rutgers University, Newark, NJ 07102, USA. E-mail: hxiong@rutgers.edu}
\thanks{Yang Yang and Jian Yang are with PCA Lab, Key Lab of Intelligent Perception and Systems for High-Dimensional Information of Ministry of Education, and Jiangsu Key Lab of Image and Video Understanding for Social Security, School of Computer Science and Engineering, Nanjing University of Science and Technology. Yang Yang is the corresponding author.}}


\IEEEtitleabstractindextext{%
\begin{abstract}
Class-Incremental Learning (CIL) aims to train a reliable model with the streaming data, which emerges unknown classes sequentially. Different from traditional closed set learning,  CIL has two main challenges: 1) Novel class detection. The initial training data only contains incomplete classes, and streaming test data will accept unknown classes. Therefore, the model needs to not only accurately classify known classes, but also effectively detect unknown classes; 2) Model expansion. After the novel classes are detected, the model needs to be updated without re-training using entire previous data. However, traditional CIL methods have not fully considered these two challenges, first, they are always restricted to single novel class detection each phase and embedding confusion caused by unknown classes. Besides, they also ignore the catastrophic forgetting of known categories in model update. To this end, we propose a Class-Incremental Learning without Forgetting (CILF) framework, which aims to learn adaptive embedding for processing novel class detection and model update in a unified framework. In detail, CILF designs to regularize classification with decoupled prototype based loss, which can improve the intra-class and inter-class structure significantly, and acquire a compact embedding representation for novel class detection in result. Then, CILF employs a learnable curriculum clustering operator to estimate the number of semantic clusters via fine-tuning the learned network, in which curriculum operator can adaptively learn the embedding in self-taught form. Therefore, CILF can detect multiple novel classes and mitigate the embedding confusion problem. Last, with the labeled streaming test data, CILF can update the network with robust regularization to mitigate the catastrophic forgetting. Consequently, CILF is able to iteratively perform novel class detection and model update. We verify the effectiveness of our model on four streaming classification task, empirical studies show the superior performances of the proposed method.
\end{abstract}
}

\maketitle

\IEEEdisplaynontitleabstractindextext

\IEEEpeerreviewmaketitle

\begin{figure}[t]
	\begin{center}
		\begin{minipage}[h]{88mm}
			\centering
			\includegraphics[width=88mm]{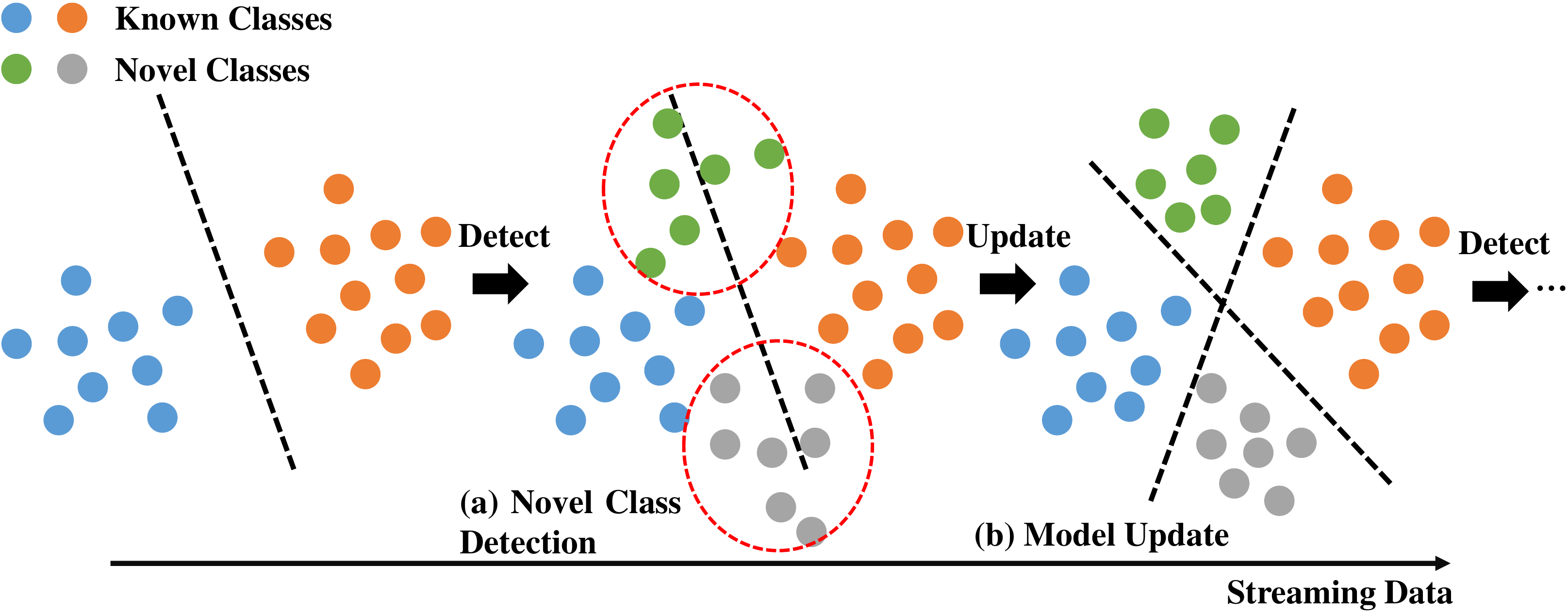}
		\end{minipage}
	\end{center}
	\caption{Schematic of class-incremental learning. Unknown categories occur with the streaming data, (a) the model first detect novel class with pre-trained model; (b) the model is then updated with newly labeled instances from unknown classes, without or with limited examples from known classes.}\label{fig:f1}
\end{figure}

\IEEEraisesectionheading{\section{Introduction}\label{sec:introduction}}
Traditional closed set recognition (CSR) assumes that training and testing data are draw from the same space, i.e., the label and feature spaces, and various methods have achieved significant success in different applications~\cite{Masi0HN18,Schmarje2020}. However, the real-world is dynamically changing, and many applications are non-stationary, which always receive the data as streaming form and many unknown classes will emerge sequentially, for example, driverless cars need to identify unknown objects, face recognition needs to distinguish unseen personal pictures, image retrieval often appears with new categories, etc. This is defined as \emph{class-incremental learning (CIL)} in literature, which is more challenging and practical than CSR. As shown in Figure \ref{fig:f1}, CIL includes two key components: novel class detection (NCD) and incremental model update (IMU). The main difficulty of NCD is to effectively distinguish the known and unknown classes, i.e., the instances from novel classes during testing, which are unknown in training phase as shown in Figure \ref{fig:f1} (a). Meanwhile, after novel class detection, we also need to consider the IMU, which aims to re-train the model with newly labeled instances from unknown classes as shown in Figure \ref{fig:f1} (b). Consequently, streaming test data continue to present novel classes, and we need to conduct the NCD and IMU operators iteratively.

To address the NCD issue, zero-shot learning (ZSL) is firstly proposed~\cite{PalatucciPHM09,FuXJXSG18}, which aims to classify instances from unknown categories, by merely utilizing seen class examples and semantic information about unknown classes. Whereas the standard ZSL methods only test unknown classes, rather than test both known and unknown classes. Thus, generalized zero-shot learning (GZSL) is proposed, which automatically detect known and unknown classes simultaneously. For example, \citeauthor{LampertNH14} introduced an attribute-based classification method, which detected new objects based on a high-level description in terms of semantic attributes~\cite{LampertNH14}; \citeauthor{ChangpinyoCGS16} proposed a GZSL method via manifold learning, which was to align the semantic space with visual features~\cite{ChangpinyoCGS16}; \citeauthor{LiJLZYH19} introduced the feature confusion GAN, which proposed a boundary loss to maximize the decision boundary of seen categories and unseen ones~\cite{LiJLZYH19}. However, both ZSL and GZSL assume that semantic information (for example, attributes or descriptions) of the unknown classes is given, which is limited to detect with prior knowledge, and have no ability to detect incrementally.

\begin{figure}[t]
	\begin{center}
		\begin{minipage}[h]{40mm}
			\centering
			\includegraphics[width=40mm]{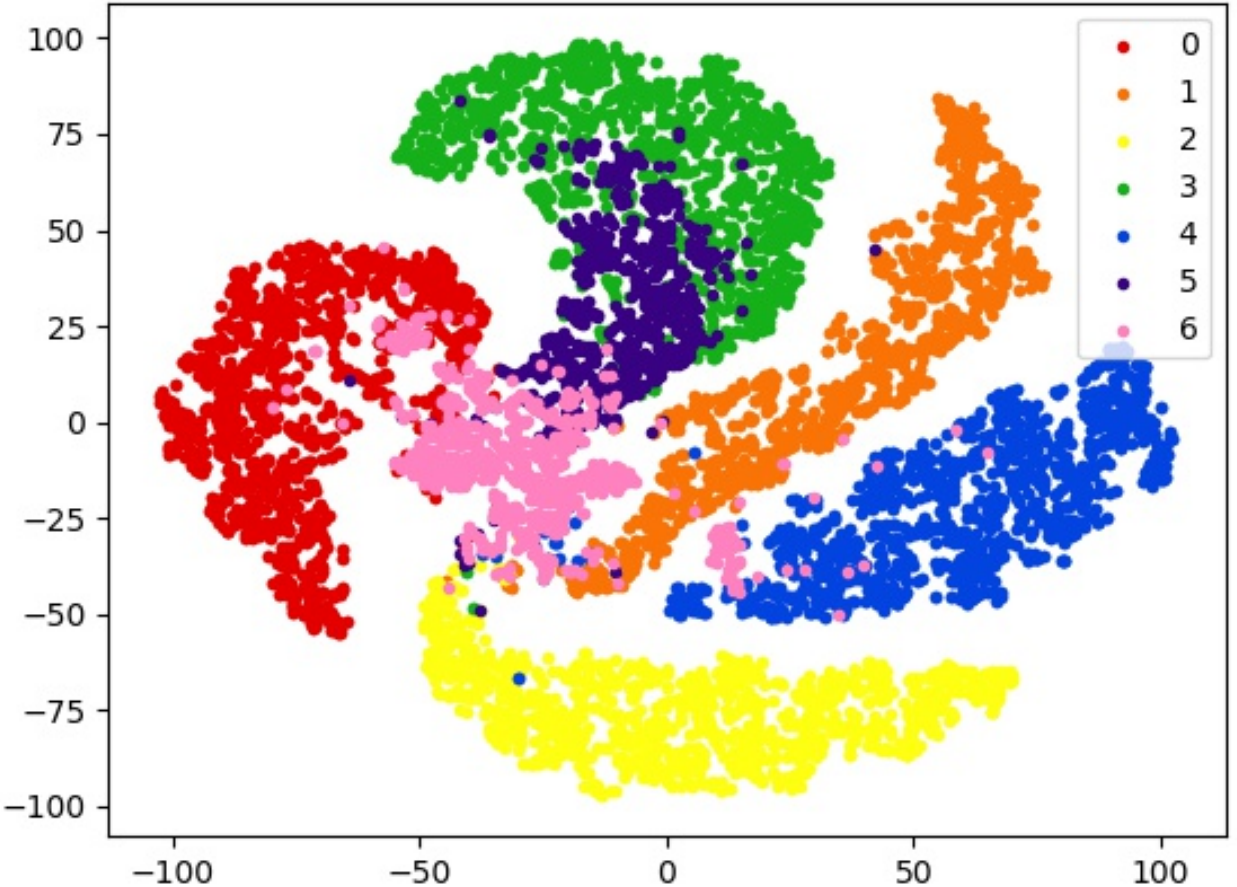}\\
			\mbox{ \;\;\;\; ({\it a}) {DM on MNIST}}
		\end{minipage}
		\begin{minipage}[h]{40mm}
			\centering
			\includegraphics[width=40mm]{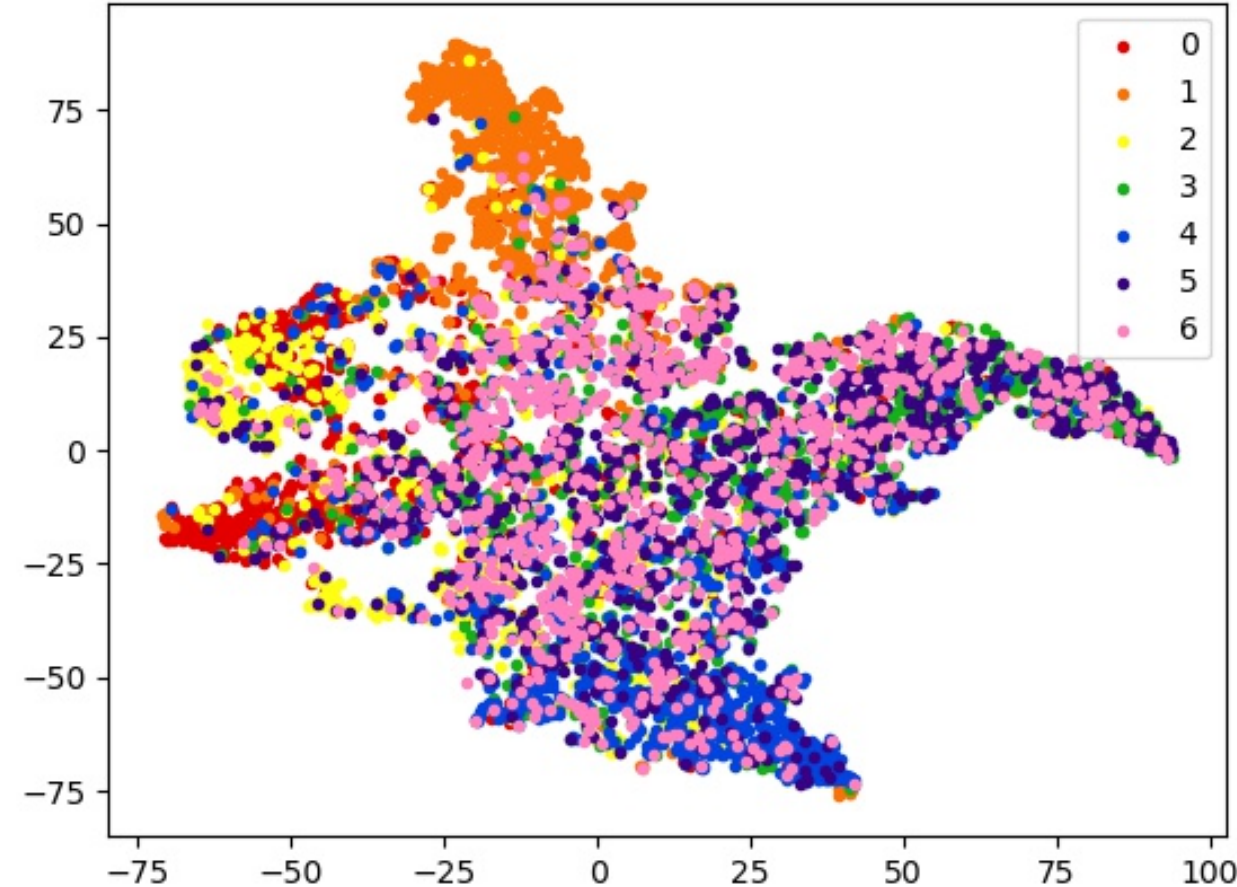}\\
			\mbox{ \;\;\;\; ({\it b}) {DM on CIFAR-10}}
		\end{minipage}\\
		\begin{minipage}[h]{40mm}
			\centering
			\includegraphics[width=40mm]{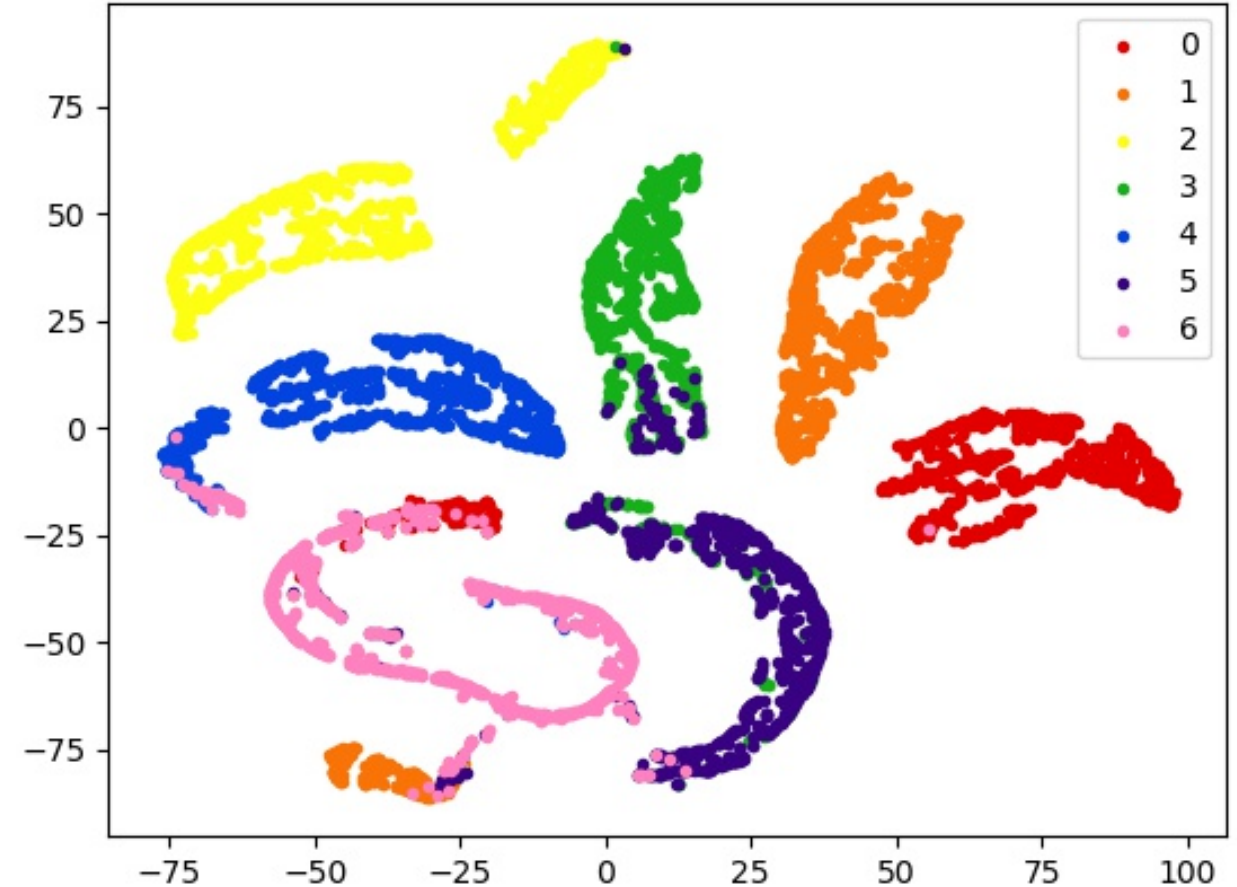}\\
			\mbox{ \;\;\;\; ({\it c}) {GM on MNIST}}
		\end{minipage}
		\begin{minipage}[h]{40mm}
			\centering
			\includegraphics[width=40mm]{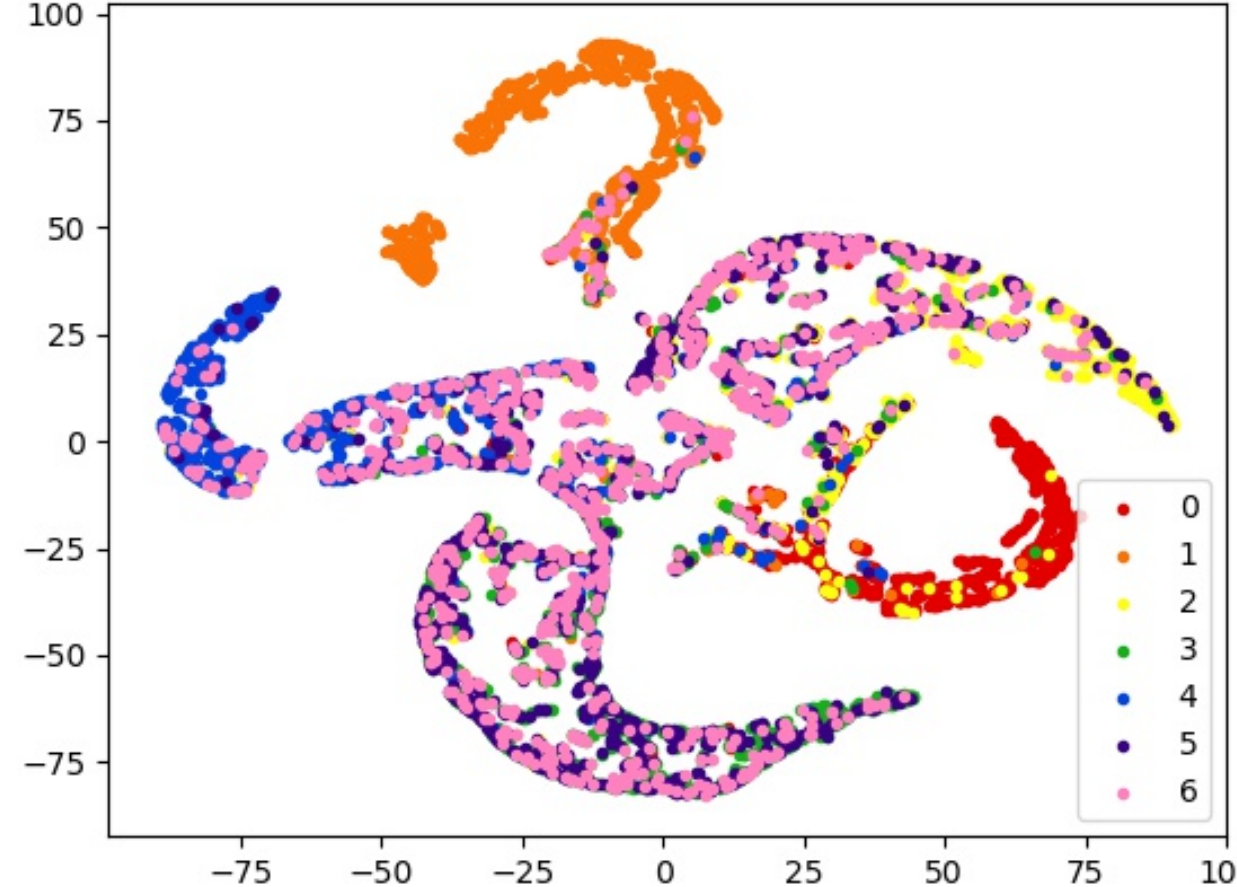}\\
			\mbox{ \;\;\;\; ({\it d}) {GM on CIFAR-10}}
		\end{minipage}
	\end{center}
	\caption{T-SNE of discriminative model (DM)~\cite{WangKCTK19} and generative model (GM)~\cite{NealOFWL18} on simple (MNIST) and complex (CIFAR-10) datasets. In detail, we train the two models with five classes (i.e., 0-4) in the training stage, then utilize the pre-trained model to achieve the feature embeddings of data from two unknown classes (i.e., 5, 6) and known classes (i.e., 0-4) appearing in the testing stage, and give the T-SNE in (a)-(d).}\label{fig:f2}
\end{figure}

Therefore, a more realistic prediction is to detect unknown classes without any information, either instances or side-information during training. Recent NCD approaches usually leverage the powerful deep neural networks, and can mainly be divided into two aspects: discriminative and generative models. Discriminative models mainly utilize the powerful feature learning and prediction capabilities of deep models to design corresponding distance or prediction confidence measures. For example, \citeauthor{BendaleB16} proposed the OpenMax model, which trained a deep neural network with the normal SoftMax layer by using the Weibull distribution fitting score~\cite{BendaleB16}, yet it failed to recognize the adversarial images which are visually indistinguishable from training instances; \citeauthor{Hassen2018} learned the neural network based representation, which restricted the inter-class and intra-class distances during training, thus to lead larger spaces for novelty detection~\cite{Hassen2018}. However, as shown in Figure \ref{fig:f2} (a) and (b), it is notable that on simple visual datasets such as MNIST, known and unknown classes have strong separability on the trained model, thereby the distance measure is more effective. Whereas in more complex visual datasets such as CIFAR-10, unknown and known categories have embedding confusion in feature space, i.e, instances of unknown and known classes are mixed, thus the performance of distance measure based methods will greatly reduce. On the other hand, Generative models mainly employ the adversarial learning to generate instances that can fool the discriminative model, thus to detect the novel class. \citeauthor{GeDG17} utilized the generative model to generate unknown class instances near the decision margin, which can provide explicit probability estimation over generated unknown class~\cite{GeDG17}. However, as shown in Figure \ref{fig:f2} (c) and (d), we can get similar conclusions to discriminative model, i.e., generated instances on complex datasets such as CIFAR-10 are almost uselesss, which is also mentioned in~\cite{NealOFWL18}. Besides, most existing detection methods are limited to detect single novel class, i.e., they assume that only one novel class appears at each period.

Furthermore, we need incremental model update (IMU) with the newly labeled instances of novel classes after detecting. Different from re-training with all previous known data, IMU aims to re-train the model only referring no or limited known data, which can ensure the efficiency of incremental update. Therefore, a big challenge for IMU is the catastrophic forgetting phenomenon~\cite{ratcliff1990connectionist}, i.e., we can find that the knowledge learned from the previous task (known classes classification) will be lost when information relevant to the current task (novel class classification) is incorporated. To mitigate the catastrophic forgetting, there are many attempts, including replay-based methods that explicitly re-train on stored examples while training on new tasks~\cite{RebuffiKSL17,IseleC18}, and regularization-based methods that utilize extra regularization term on output or parameters to consolidate previous knowledge~\cite{NealOFWL18,WangKCTK19,Cuong2018}.

To this end, we propose a Class-Incremental Learning without Forgetting (CILF) framework, which aims to process new class detection and model update iteratively. In detail, we firstly develop a novel decoupled prototype based network to train the known classes, which employs the constrained clustering loss to regularize the inter-class and intra-class structure. In testing, considering emergence of single or multiple novel classes, we develop the curriculum operator for learning adaptive embedding, which aims to conduct learnable clustering from easy to hard instances and overcome the embedding confusion. Then, with the limited memory data of known classes, CILF updates the network with robust regularization to mitigate the catastrophic forgetting. In summary, the main contributions are summarized as follows:
\begin{itemize}
	\item Propose the `` Class-Incremental Learning without Forgetting'' (CILF) framework, which considers both novel class detection and incremental model update;
	\item Propose a novel decoupled prototype based network, which can conduct novel class detection and model update effectively;
	\item Propose curriculum clustering operator for better multiple novel classes detection and robust regularization to mitigate catastrophic forgetting.
\end{itemize}

In remaining sections of this paper, section 2 introduces the related work. Section 3 presents the proposed method. Section 4 evaluates the proposed method. Finally, the whole work is concluded in Section 5.

\section{Related Work}
Our work aims to detect novel classes in streaming data, and update the model with limited known data without forgetting. Therefore, our work is related to: novel class detection and incremental model update.  

Traditional novel class detection approaches mainly restrict intra-class and inter-class distance property in training data, then detect novel class by identifying outliers. For example, \citeauthor{DaYZ14} developed a SVM-based method, which learned the concept of known classes while incorporating the structure presented in the unlabeled data collected from open set~\cite{DaYZ14}; \citeauthor{MuZDLZ17} proposed to dynamically maintain two low-dimensional matrix sketches for detecting novel classes~\cite{MuZDLZ17}. However, these approaches are difficult to process high dimensional space considering complex matrix operations. Recently, with the development of deep learning techniques, several studies have applied convolutional neural network (CNN) on the detection scenario. \citeauthor{HendrycksG17} verified that CNN trained on the MNIST images can predict high confidence (90$\%$) on gaussian noise instances, thus we can use the softmax output probabilities to distinguish known/unknown class~\cite{HendrycksG17}. Furthermore, \citeauthor{LiangLS18} directly utilized temperature scaling or added small perturbations to separate the softmax score distributions between in- and out-of-distribution images~\cite{LiangLS18}; \citeauthor{NealOFWL18} introduced a novel augmentation technique, which adopted an encoder-decoder GAN architecture to generate the synthetic instances similar to known class~\cite{NealOFWL18}; \citeauthor{WangKCTK19} proposed a cnn-based prototype ensemble method, which adaptively update the prototype for robust detection~\cite{WangKCTK19}. Nevertheless, these methods always limited to detect single novel class in once time. Therefore, \citeauthor{HanVZ19} proposed an extended deep transfer embedded clustering method for multiple novel class detection~\cite{HanVZ19}. Nevertheless,  existing NCD methods usually have superior detection performance on simple datasets, but are easily interfered by embedding confusion on complex datasets. 

Incremental learning is always applied for streaming data. In most situations, only a few examples from known classes/features/distributions are available in the beginning and data with new classes/features/distributions emerge thereafter. Incremental learning methods aim to update the models from streaming data sequentially only with newly coming data and limited previous data, without re-training with all previous data~\cite{ZhangYJXZ16}. As a matter of fact, incremental deep learning can directly apply with online backpropagation, yet with one important drawback: catastrophic forgetting, which is the tendency for losing the learned knowledge of previous distribution (previously known classes/features/distributions). To solve this problem, there are many attempts, for example, \citeauthor{RebuffiKSL17} stored a subset of examples per class,  which are selected to best approximate the mean of each class in the feature space~\cite{RebuffiKSL17}, \citeauthor{PazR17} projected the estimated gradient direction on the feasible region outlined by previous tasks through a first order Taylor series approximation~\cite{PazR17}; \citeauthor{LiH16} utilized the output of previous model as soft labels for previous tasks~\cite{LiH16}; \citeauthor{KirkpatrickPRVD16} proposed the elastic weight consolidation to reduce catastrophic forgetting~\cite{KirkpatrickPRVD16}; \citeauthor{LeeKJHZ17} proposed to incrementally match the moment of posterior distribution of the neural network~\cite{LeeKJHZ17}.

\section{Proposed Method}
In this section, we formalize the problem of class-incremental learning with streaming data, and give the details of proposed framework.
\subsection{Problem Definition}
Without any loss of generality, at initial time, we have a supervised training set $D^0=\{(\x_i^0,\y_i^0)\}_{i=1}^N$, where $\x_i^0 \in \R^d$ denotes the $i-$th instance, and $\y_i^0 \in Y^0 = \{1,2,\cdots,C\}$ denotes corresponding label, $0$ represents initial time. Then, we receive a non-stationary unlabeled testing data $D^1 = \{(\x_j)\}_{j=1}^{N_1}$, where $\x_j \in \R^d$ denotes the $j-$th instance, and label $\y_j \in \hat{Y} = \{1,2,\cdots,C,C+1,\cdots,C+K^1\}$ is unknown, $K^1$ is the number of unknown classes. Thus, novel class detection can be defined as:

\begin{defn}\label{def:d1}
	\textbf{Novel Class Detection (NCD)}  With the initial training set $D^0=\{(\x_i^0,\y_i^0)\}_{i=1}^N$, we aim to construct a model, i.e., $f^0: X^0 \rightarrow Y^0$. Then with the pre-trained model $f^0$, novel class detection is to classify the known and unknown classes in $D^1$ accurately.
\end{defn}

On the other hand, it is notable that streaming data with novel classes has two characteristics: (1) Data window. At time window $t$, we only get the data of current time window, not the full amount of streaming data for detection; and (2) Novel class continuity. At time window $t$, only partial novel classes will appear, or even no novel classes. Therefore, we need to incrementally detect novel classes, i.e., with the streaming data, every time after receiving the data of time window $t$, NCD is performed~\cite{geng2018}. Specifically, the streaming test data $D$ can be denoted as $D = \{D^t\}_{t=1}^T$, where $D^t = \{\x_j^t\}_{j=1}^{N_t}$ is with $N_t$ unlabeled instances, and the underlying label $\y_j^t \in \hat{Y}^t$ is unknown, $\hat{Y}^t = \hat{Y}^{t-1} \cup Y^t$, where $\hat{Y}^{t-1}$ is the cumulative known classes until $(t-1)-$th time window and $Y^t$ is the new classes in $t-$th window. $\hat{Y}^T = \hat{Y} = \{1,2,\cdots,C,C+1,\cdots,C+K\}$. Therefore, we can give the definition of class-incremental learning:

\begin{defn}\label{def:d2}
	\textbf{Class-Incremental Learning (CIL)} At time $t \in \{1,2, \cdots,T\}$, we have pre-trained model $f^{t-1}$ and finite stored instance set $M^{t-1}$ from known classes until $(t-1)-$th time, and receive streaming data $D^t$. First, we aim to classify known and unknown classes in $D^t$ as Definition~\ref{def:d1}. Then, with the newly labeled data from novel classes and stored data $M^{t-1}$, we update the model while mitigating forgetting to acquire $f^t$. Cycle this process until terminated.
\end{defn}


Note that there exist two labeling cases after novel class detection, i.e., manually labeling or self-taught labeling~\cite{MuZDLZ17}. We consider manually labeling following most approaches~\cite{geng2018,MuZDLZ17,WangKCTK19} to avoid label noise accumulation, and is more in line with real-world applications.

\begin{figure}[t]
	\begin{center}
		\begin{minipage}[h]{90mm}
			\centering
			\includegraphics[width=90mm]{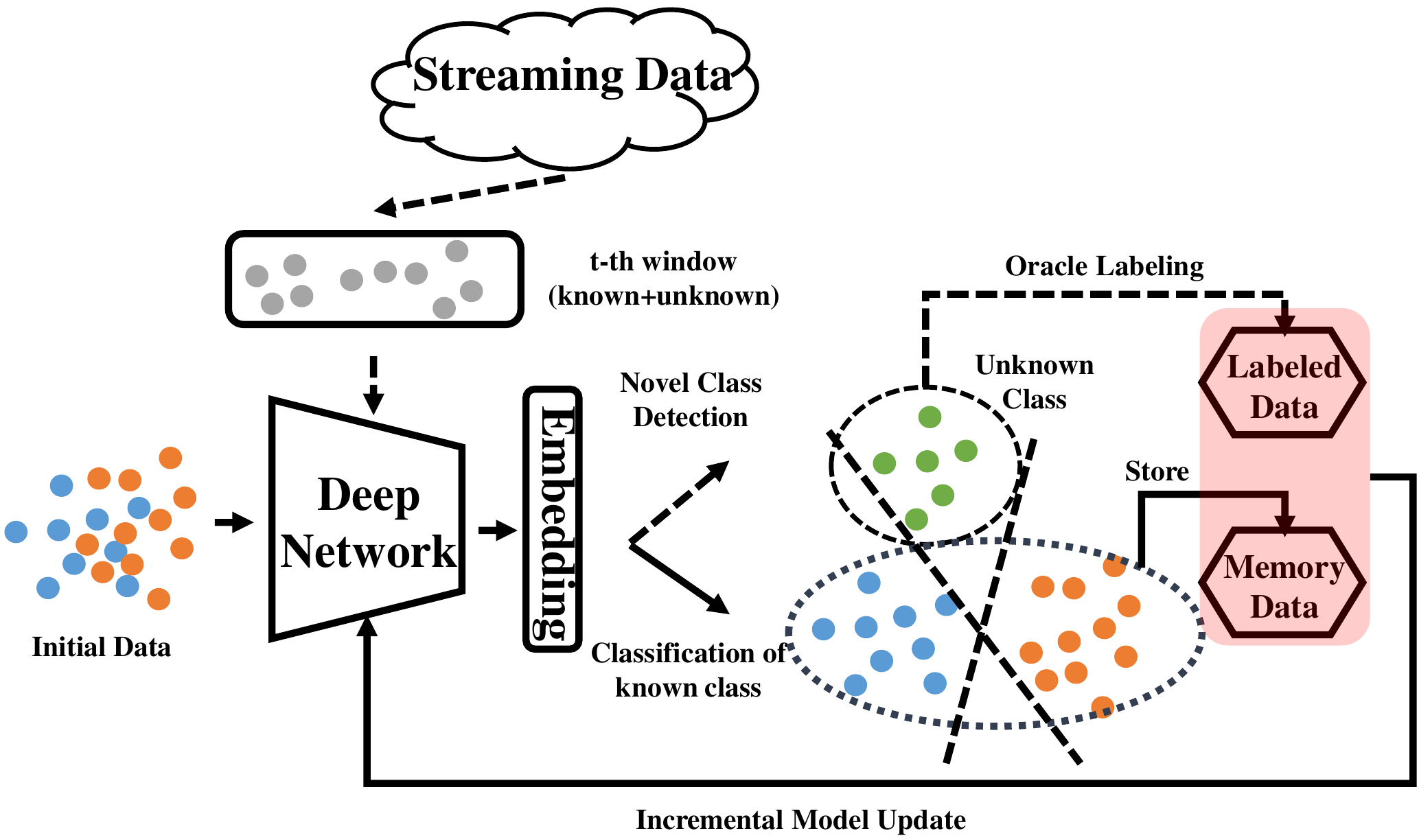}
		\end{minipage}
	\end{center}
	\caption{Overview of the CILF framework. The blue and orange dots are initial training data for developing the deep network. While the gray dots are unlabeled testing data of $t-$th time window, which are received from the stream data. With the trained deep network, CILF aims to classify the known and novel classes, then query the ground-truths of novel class instances for updating the network continuously.}\label{fig:fra}
\end{figure}

\subsection{CILF Framework}
The main idea of CILF is to learn the feature embeddings such that instances exhibit distinguishing characteristics for label prediction, novel class detection, and subsequent model update over the non-stationary streaming data. Therefore, the most critical parts of CILF are: (1) feature embedding network; (2) novel class detection operator; and (3) model update mechanism.

\begin{itemize}
	\item \textbf{Feature Embedding Network:} With the initial training data, i.e., the blue and orange dots as shown in figure \ref{fig:fra}, the decoupled neural network model is trained using the labeled initial data with the prototype based loss, which concerns the intr-class/inter-class structure and can be easily transformed for NCD;  
	\item \textbf{Novel Class Detection Operator:} At time $t$, we receive a set of unlabeled data from the streaming data, i.e., gray dots as shown in figure \ref{fig:fra}, which includes known (blue and orange dots) and unknown (green dots) classes. The observed instances set $D^t$ are transformed through learned network, and achieve feature representations. Then we employ the pre-trained $f^{t-1}$ for curriculum clustering operating, which can detect multiple unknown classes from easy to difficult in self-taught form;   
	\item \textbf{Model Update Mechanism:} After NCD, true labels of instances from novel classes are queried (partially or fully), then IMU is performed on $f^{t-1}$ with the newly labeled data, while regularizing the performance of stored data from known classes to mitigate forgetting. The updated model is then used to further classify incoming instances along the stream.
\end{itemize}

This process is repeated until the end of streaming data. Figure \ref{fig:fra} illustrates the overall streaming data classification performed by CILF framework.  And Table \ref{tab:tab} provides the definition of symbols used in this paper.

\begin{table}[t]{
		\centering
		\caption{Description of symbols.}
		\label{tab:tab}
		\begin{tabular}{l|l}
			\toprule
			Sym. &Definition\\
			\midrule
			$D^0=\{(\x_i^0,\y_i^0)\}_{i=1}^N$ & initial supervised training data\\
			$D^t=\{\x_j^t\}_{j=1}^{N_t}$ & set of unlabeled data at $t-$th time window\\
			$D_{new}^t$ & set of labeled new class data at time $t$\\
			$\hat{Y}^t$ & label set at $t-$th time window\\
			$f^t$ & trained model $t-$th time window\\
			$M^t$ & stored memory data of known classes \\
			&until $t-$th time window\\
			$f^t(\x)$ & feature embedding of instance $\x$\\
			$p_{i}^t$ & prediction of $i-$th instance at time $t$\\
			$\mu_c^t$ & prototype of $c-$th class at time $t$\\
			$w_j^t$ & weight of each instance at time $t$\\
			$g(l)$ & pacing function to determine the number of \\
			&selected instances in each mini-batch\\
			\bottomrule	
	\end{tabular}}
\end{table}

\subsection{Feature Embedding Network}
Given the initial training data $D^0$, our primary objective is to build an effective model $f^0$ for subsequent classification. Recent researches have demonstrated the effectiveness of deep model on feature embedding and subsequent tasks, thereby we employ the deep models for building $f^0$, for example, convolution neural networks for images. Importantly, the built deep model needs to consider two aspects: (1) Distance measure. The model needs to emphasize the exploitation of feature embeddings considering intra-class compactness and inter-class separability, thus leave larger space for novel class detection; (2) Model scalability. The model needs to effectively learn novel class knowledge and incrementally update model with the emergence of novel class data. However, traditional deep models using cross entropy cannot consider the distance measure effectively, and are difficult to conduct the model update (the prediction layer is coupled to the fully connected layer, and is difficult to expand). Consequently, we develop a decoupled deep embedding network with prototype based loss to improve the inter-class and intra-class structure.

Particularly, for a given input $\x_i^0$, the output feature representations are denoted as $f^0(\x_i^0,\theta)$, $\theta$ is the correspond network parameters, and we utilize notation $f^0(\x_i^0)$ for clarity. Inspired from the topic of metric learning~\cite{Kulis13}, the loss can be defined as: 
\begin{equation}\label{eq:all}
\begin{split}
L = L_{intra} + \lambda L_{inter}
\end{split}
\end{equation}
where $L_{intra}$ aims to pull data towards their neighbor from same class, and $L_{inter}$ is to push data away from different classes. $\lambda$ is the balance parameter.

\subsubsection{Intra-Class Compactness}
$L_{intra}$ can be obtained by calculating the distance between each instance and corresponding prototype, here we utilize the class center. Similar to cross entropy loss~\cite{BishopN07}, i.e., $\sum_{i=1}^N -\y_i log(g(f(\x_i,\theta)))$, where $\y_i$ is the ground truth of $i-$th instance, $g(\cdot)$ denotes the fully connected layer with softmax function. $L_{intra}$ is depending on the feature output $f^0(\x^0)$. Consequently, we define the prototype-based cross entropy loss as following:   
\begin{equation}\label{eq:eq2}
\begin{split}
L_{intra} = \sum_{i=1}^N \sum_{c=1}^C -y_{i_c}^0 log(p_{i_c}^0)
\end{split}
\end{equation}
where $p_{i_c}^0$ is the probability of $\x_i^0$ being classified as $y_c^0$, which is negatively related to the distance between instance and prototype of $c-$th class, i.e., the probability is larger if the distance is closer, otherwise is smaller. Thus the $p_{i_c}^0 \propto -\|\x_i^0 - \mu_c^0\|_2^2$, where $\mu_c^0$ is the representations of $c-$th class prototype, and can be defined as following:
\begin{equation}\label{eq:eq1}
\begin{split}
p_{i_c}^0 = \frac{exp(-\alpha\|f^0(\x_i^0)-\mu_c^0\|_2^2)}{\sum_{m=1}^C exp(-\alpha\|f^0(\x_i^0)-\mu_m^0\|_2^2)}
\end{split}
\end{equation}
where $C$ is the class number, and $\alpha$ is a hyperparameter that controls the strength of distance similar to large margin cross-entropy~\cite{LiuWYY16}. Note that Eq. \ref{eq:eq1} minimizes loss via maximizing the probability of $\x_i^0$ being associated with the prototype $\mu_{y_i}^0$. Moreover, it is crucial to initialize and update each class prototype effectively. The labels of initial training data $\D^0$ are given, thereby we use the output representation $f^0(\x^0)$ , for each prototype initialization:
\begin{equation}
\begin{split}
\mu_c^0 =  \frac{1}{|\pi_c|} \sum_{\x^0 \in \pi_c} f^0(\x^0)\nonumber
\end{split}
\end{equation}
where $|\pi_c|$ is the size of $c-$th class. On the other hand, the key idea of prototype update is to anneal clusters slowly to eliminate the biased instances in each mini-batch. Thus we propose to smooth the annealing process via temporal ensemble~\cite{LaineA17}:
\begin{equation}\label{eq:eq}
\begin{split}
\mu_c^{0_e} = \beta\mu_c^{0_{e-1}}+(1-\beta)\mu_c^{0_e}
\end{split}
\end{equation}
where $\beta$ is a momentum term controlling the ensemble, and ${0_e}$ indicates the $e-$th epoch for the initial training.

\subsubsection{Inter-Class Separability}
The prototype-based cross entropy loss guarantees the local intra-class compactness, while neglects the inter-class separability. To make the projection of instances robust in distance measure, $L_{inter}$ focuses on improving global separation between different classes. Particularly, $L_{inter}$ aims to transform instances from similar classes to be closer than those from different classes, i.e., $d(f^0(\x_i^0),f^0(\x_p^0))<d(f^0(\x_i^0),f^0(\x_n^0))$, where $\x_i^0, \x_p^0$ share same class and $\x_n^0$ is from different class, $d(f^0(\x_i^0), f^0(\x_j^0))$ is a metric function measuring distances in the embedding space, and we use notation $d_{i,j}$ for clarity. This is known as the triplet loss with a pre-specified margin value $m$, i.e., $\big[m+d_{a,p}-d_{a,n}\big]_+ = \max\{0,m+d_{a,p}-d_{a,n}\}$. It is notable that triplet loss always suffers from slow convergence, thus triplet construction is central for improving the performance. Inspired by~\cite{YuT19}, we consider the hard triplet to fully explore multiple negative examples from different classes in each mini-batch, which can further improve the inter-class distances. In result, hard triplet is denoted as:
\begin{equation}\label{eq:eq3}
\begin{split}
\Omega_i = (\x_i^0,\x_p^0,\x_{n_1}^0,\cdots,\x_{n_C-1}^0) 
\end{split}
\end{equation}
where $C$ is the class number and $C-1$ negative examples $\x_{n_c}$ is from different classes. Eq. \ref{eq:eq3} can better consider the global inter-class distances. Thereby the $L_{inter}$ can be defined as:
\begin{equation}\label{eq:eq4}
\begin{split}
L_{inter} = \sum_{\Omega_i \in \mathbb{T}} \big[m+d(i,p)-\min\limits_{\x_{n_c}^0 \in \Omega_i}d_{i,n_c}\big]_+
\end{split}
\end{equation}
where $\mathbb{T}$ denotes the hard triplet set. Here we utilize euclidean distance to evaluate the distance between two examples:
\begin{equation}
\begin{split}
d_{i,j} =  \|f^0(\x_i^0)- f^0(\x_j^0)\|_2^2
\end{split}
\end{equation}
Consequently, we can learn discriminative feature embedding, and boost the performance of classification and detection via optimizing Eq. \ref{eq:all}: (1) Prototype-based loss highlights the compactness of representation, i.e., the intra-class would be more compact and inter-class would be more distant. This property is suited for distinguishing the known and unknown classes; and (2) Prototype-based loss is based on the feature output embedding, which is independent of prediction layer. Therefore, it is easy to update the model and learn novel classes, without the expansion of model structure (prediction layer). The details are shown in Algorithm \ref{alg:alg1}.


{\begin{algorithm}[t]
		\caption{Feature Embedding Network}
		\label{alg:alg1}
		\begin{compactitem}
			\item \textbf{Input}:
			\item Data set: $D^0=\{(\x_i^0,\y_i^0)\}_{i=1}^N$
			\item Parameter: $\lambda$, $\alpha$, Learning rate parameter: $\eta$
		\end{compactitem}
		\begin{compactitem}
			\item \textbf{Output}:
			\item Decoupled deep clustering network: $f^0$
		\end{compactitem}
		\begin{algorithmic}[1]{
				\STATE Initialize model parameters $\theta$;
				\STATE Initialize the prototype $\mu$ for each class;
				\WHILE {stop condition is not triggered}
				\FOR{instance mini-batch}
				\STATE Calculate $L_{intra}$ according to Equation \ref{eq:eq2};
				\STATE Calculate $L_{inter}$ according to Equation \ref{eq:eq4};
				\STATE Calculate loss $L = L_{intra} + \lambda L_{inter}$ according to Equation \ref{eq:all};
				\STATE Update model parameters using gradient descent;
				\ENDFOR
				\STATE Update the prototype $\mu$ according to Equation \ref{eq:eq};
				\ENDWHILE
			}
		\end{algorithmic}
\end{algorithm}}

\subsection{Novel Class Detection}

Traditional closed-set methods predict the known classes of training phase, in which the number of possible labels at testing is known and fixed. However, in class-incremental setting, instances belonging to unknown classes may appear with the streaming test data. Therefore, we need to distinguish the known and unknown classes. Specifically, we receive a set of unlabeled data $D^t$ at $t-$th time, and there may occur $K^t$ novel classes, where $K^t \geq 0$. However, most current detection methods either assume that only one novel class appears per time, i.e., $K^t = 1$, or classify multiple novel classes as a super-class, which is impractical and difficult to operate considering efficiency. To solve this problem, we aim to fine-tune the deep clustering network $f^{t-1}$ of last time for multiple novel class detection. As shown in Figure \ref{fig:f2}, a key challenge is that adversarial instances of novel classes are mixing with known classes in complex scenario, leading the embedding confusion and greatly affecting the clustering effect, i.e., biased prototypes for known and unknown classes. To solve this problem, we employ a learnable curriculum clustering operator, which aims to conduct clustering from easy (distinguishable) to difficult (confused) instance via curriculum learning~\cite{BengioLCW09}. Consequently, we can acquire more reliable prototype and novel class detection result.

In detail, considering the model training in Algorithm \ref{alg:alg1} is entirely supervised, whereas $D^t$ is unsupervised, we aim to discover novel classes in $D^t$ by unsupervised clustering, which fine-tunes the $f^{t-1}$ trained on $t-1$ phase with easy instances first, and then cluster the mixed ones. We address this challenge by decomposing the learnable curriculum clustering into two closely related sub-tasks as curriculum learning. The first is \emph{weighting function} to calculate the weight of each instances, and initialize the prototype with weighted k-means. The second is \emph{pacing function} to determine the pace for which data are presented to fine-tune the model, thus conduct curriculum clustering. 

\subsubsection{Weighting Function}
Inspired by~\cite{HacohenW19}, we evaluate the weight of each instance by self-taught weighting function. In detail, we compute confidence score for each instance $\x_j^t$ in $D^t$ using existing model $f^{t-1}$. We first obtain the statistic confidence by applying intra-class distance using Eq. \ref{eq:eq2}, i.e., $u_j^t = \sum_{c=1}^{\hat{Y}^{t-1}} -y_{j_c}^{t-1} log(p_{j_c}^t)$. It is notable that $u_j^t$ of the instance near prototype are smaller, and $u_j^t$ of the instances away from all class prototypes are larger. Therefore, the weight of each instance can be denoted as $w_j^t = (u_j^t - \gamma)^2$, where $\gamma$ is the threshold parameter. Thereby the highly confident and unsure instances have larger weights, and confusing ones have lower weights. 

On the other hand, Algorithm \ref{alg:alg1} requires initial setting for prototypes $\mu_c^t, c \in \hat{Y}^t$. Thus we initialize prototypes by running semi-supervised weighted k-means algorithm~\cite{DhillonGK07} by combing the unlabeled set $D^t$ and pre-trained $\mu_c^{t-1}$. In result, we can obtain more robust initial prototypes:
\begin{equation}\label{eq:weight}
\begin{split}
\mu_c^{t} = \left\{
\begin{aligned}
&\beta \mu_c^{t-1} + (1-\beta) \sum_{\x_j^t \in \pi_c} \frac{ w_j^t f^{t-1}(\x_j^t)}{\sum_{\x_j^t \in \pi_c} w_j^t}, {when \quad c \in \hat{Y}^{t-1}},\\
&\sum_{\x_j^t \in \pi_c} \frac{ w_j^t f^{t-1}(\x_j^t)}{\sum_{\x_j^t \in \pi_c} w_j^t}  ,\qquad\qquad\qquad\quad{\rm ~~} {when \quad c \in Y^t}   
\end{aligned}
\right.
\end{split}
\end{equation}
where $\pi_c$ is the set of $c-$the class, in which the pseudo-label $\arg\max\{p_{j_c}^t\}$ of each instance can be calculated by Eq. \ref{eq:eq1}.

\subsubsection{Pacing Function}
A direct way for classifying known and unknown classes in $D^t$ is to fine-tune $f^{t-1}$ using all the instances. However, considering the embedding confusion, the initialized prototypes are biased and pseudo-labels exist noises. If we randomly sample batches from the full amount of data to fine-tune model, the embedding confusion will further affect the update of prototypes and pseudo-labels. Therefore, we turn to sort the instances according to difficulty, and present from easier to harder instances for fine-tuning with the model capability increase, rather than giving a sequence of mini-batches uniformly from $D^t$ in most common training procedure, $L$ is the number of batches. 

In detail, the pacing function $h: [L] \rightarrow [N_t]$ is used to determine a sequence of subsets $\{B_1,B_2,\cdots,B_L\} \in D^t$, of size $|B_l| = h(l)$. The $l$-th subset $B_l$ includes the first $h(l)$ elements of the instances, which are sorted by the scoring function in ascending order. Here, we utilize the fixed exponential pacing, which has a fixed step length, and exponentially increasing size in each batch. Formally, it is given by:
\begin{equation}\label{eq:pacing}
\begin{split}
h(l) = min(\upsilon \cdot \delta^{\lfloor \frac{l}{\phi} \rfloor},1)\cdot {N_t}
\end{split}
\end{equation}
Where $\upsilon$ denotes the fraction of data in the initial step, $\delta$ is the exponential factor for increasing the size of sampled mini-batches in each step, $\phi$ is the number of iterations in each step, $\lfloor \cdot \rfloor$ denotes round down, $l$ is the index of batches, $N_t$ is the number of instances. Consequently, in each mini-batches, we select episodic data with variable length for reliable fine-tuning.

\subsubsection{Fine-tune Clustering}
With the sampled mini-batches $\{B_1,B_2,\cdots,B_L\}$ in each epoch, we aim to fine-tune the $f^{t-1}$ from easier to harder, and Eq. \ref{eq:eq2} can be reformulated as:
\begin{equation}\label{eq:finetune}
\begin{split}
L^t & = L_{intra}^t + \lambda_1 L_{inter}^t + \lambda_2 R^t \\
L_{intra}^t & = \sum_{l=1}^L \sum_{j=1}^{|B_l|} \sum_{c = 1}^{|\hat{Y}^{t}|} -\bar{y}_{l_{j_c}}^t log(p_{l_{j_c}}^t)\\
L_{inter}^t & = \sum_{l=1}^L \sum_{\Omega_j \in \mathbb{T}_l} \big[m+d_{l_{j,p}}-\min\limits_{\x_{n_c}^t \in \Omega_j}d_{l_{i,n_c}}\big]_+ \\
R^t & = \sum_{c=1}^{|\hat{Y}^{t-1}|} \|\mu_c^t - \mu_c^{t-1}\|_2^2
\end{split}
\end{equation}
where $\mathbb{T}_l$ denotes the triplet set of $l-$th batch. $R^t$ aims to constraint the updated prototypes of known classes approaching the pre-trained ones, which can regularize the embeddings of known classes. The pseudo-labels $\bar{\y}_j^t = \arg\max\{p_{j_c}^t\}$ for each instance can be calculated by Eq. \ref{eq:eq1}. So far, we assume that the number of classes $K^t$ is known, which is impractical in real applications. Thus we aim to estimate the number of classes in the unlabeled data. Specifically, we fine-tune clustering using $D^t$ by varying the number of unknown classes. The resulting clusters are then examined by computing cluster validity index (CVI), which concerns the intra-cluster cohesion vs inter-cluster separation. And we select the generally used Silhouette index~\cite{ArbelaitzGMPP13}:
\begin{equation}\label{eq:cvi}
\begin{split}
CVI = \sum_{\x \in D^t}\frac{b(\x)-a(\x)}{\max\{a(\x),b(\x)\}}
\end{split}
\end{equation}
where $a(\x)$ is the average distance between $\x$ and all other data instances within the same cluster, and $b(\x)$ is the smallest average distance of $\x$ to all instances in any other different cluster. The optimal number of categories is the inflection point of CVI with maximum curvature. The details are shown in Algorithm \ref{alg:alg2}.

{\begin{algorithm}[htb]
		\caption{Novel class Detection}
		\label{alg:alg2}
		\begin{compactitem}
			\item \textbf{Input}:
			\item Data set: $D^t=\{(\x_j^t\}_{ij=1}^{N_t}$
			\item Parameter: $\beta$, $\gamma$, $\upsilon$, $\delta$, $\phi$
		\end{compactitem}
		\begin{compactitem}
			\item \textbf{Output}:
			\item Novel Class Detection Network: $\hat{f}^t$
		\end{compactitem}
		\begin{algorithmic}[1]{
				\FOR{$0 \leq K \leq K_{max}^t$}
				\STATE Initialize prototypes $\mu_c^t$ according to Equation \ref{eq:weight};
				\WHILE {stop condition is not triggered}
				\STATE Generate mini-batches $\{B_1,B_2,\cdots,B_L\}$ according to Equation \ref{eq:pacing};
				\FOR{instance mini-batch $X_l$}
				\STATE Calculate $L^t$ using Eq. \ref{eq:finetune} similar to Algorithm \ref{alg:alg1}; 
				\STATE Fine-tune model parameters using gradient descent;
				\ENDFOR
				\STATE Update the prototype $\mu_c^t$ according to Equation \ref{eq:eq};
				\STATE Update the pseudo-labels $\bar{\y}$ according to Equation \ref{eq:eq1};
				\ENDWHILE
				\STATE Computer CVI for $D^t$ according to Eq. \ref{eq:cvi};
				\ENDFOR
				\STATE Let $\hat{f}^t$ as the $K^{*}$ with optimal CVI value.
			}
		\end{algorithmic}
\end{algorithm}}

\subsection{Incremental Model Update}
Ideally, the initial model training and novel class detection processes can identify the known and unknown classes. However, considering streaming data with unceasing novel class, we need reliable training data of novel classes to create new prototypes and update the model parameters in incremental fashion. Thus, we need to collect novel class data  for labeling, which can be used to re-train $f^{t-1}$. Similar to previous studies~\cite{HaqueKB16,MuZDLZ17}, after curriculum clustering operator for detection, we can achieve potential novel class instances $D_{new}^t$ for querying their true labels. Note that we can query full or only partial data of novel class. However, there exist catastrophic forgetting of known classes if we only use the new data to update the model.

To solve this problem, we develop a mechanism to incorporate the stored memory and novel class information incrementally, which can mitigate the forgetting of discriminatory characteristics about known classes. In detail, we utilize the exemplary data $M^{t-1}$ for regularization in re-training:
\begin{equation}\label{eq:update}
\begin{split}
L(D_{new}^t,M^{t-1}) & = \hat{L}_{intra}^t + \lambda_1 \hat{L}_{inter}^t + \lambda_2 R^t \\\\
\hat{L}_{intra}^t & = \sum_{\x_j \in D_{new}^t \cup M^{t-1}} - y_j^t log(p_j^t) + \\ & \sum_{\x_j \in M^{t-1}} f^{t-1}(\x_j) \log f^t(\x_j)\\
\hat{L}_{inter}^t & = \sum_{\Omega_i \in \mathbb{T}^t} \big[m+d_{l_{i,p}}-\min\limits_{\x_{n_c}^t \in \Omega_i}d_{l_{i,n_c}}\big]_+ \\
R^t & = \sum_{c=1}^{|\hat{Y}^{t-1}|} \|\mu_c^t - \mu_c^{t-1}\|_2^2
\end{split}
\end{equation}
The first term encourages the network to output the correct class indicator (classification loss) for all labeled examples, i.e., $D_{new}^t$ and $M^{t-1}$, and reproduces the scores calculated in the previous step (distillation loss) for stored in-class examples, i.e., $M^{t-1}$. $\mathbb{T}^t$ is constituted from $D_{new}^t$ and $M^{t-1}$. After re-training, we need to update the $M^{t-1}$ to store key points of novel classes, we randomly remove $\frac{|Y^t||M|}{|\hat{Y}^{t-1}||\hat{Y}^{t}|}$ instances for each known class, and sample $\frac{|M|}{|\hat{Y}^{t}|}$ instances for each novel class. The details are shown in Algorithm \ref{alg:alg3}.


{\begin{algorithm}[htb]
		\caption{Class-Incremental Learning}
		\label{alg:alg3}
		\begin{compactitem}
			\item \textbf{Input}:
			\item Data set: memory data $M^{t-1}$, labeled novel class data $D_{new}^t$
			\item Learning rate parameter: $\eta$
		\end{compactitem}
		\begin{compactitem}
			\item \textbf{Output}:
			\item Re-trained deep clustering Network: $f^t$
		\end{compactitem}
		\begin{algorithmic}[1]{
				\STATE Calculate the $f^{t-1}(\x_j)$ of the examples from $M^{t-1}$ and $D_{new}^t$;
				\WHILE {stop condition is not triggered}
				\FOR{instance mini-batch}
				\STATE Calculate $L(D_{new}^t,M^{t-1},f^t)$ according to Eq. \ref{eq:update}; 
				\STATE Re-train model parameters using gradient descent;
				\ENDFOR
				\STATE Update the prototype $\mu$ according to Equation \ref{eq:eq};
				\ENDWHILE
			}
		\end{algorithmic}
\end{algorithm}}

\section{Experiments}
In this section, we mainly verify the proposed CILF from two aspects: (1) classification of known and novel classes; and (2) forgetting of known classes. Considering that most large-scale datasets are concentrated on images, thus we empirically evaluate CILF comparing with the state-of-the-art approaches on four simulated stream image datasets.

\begin{figure}[htb]
	\begin{center}
		\begin{minipage}[h]{44mm}
			\centering
			\includegraphics[width=44mm]{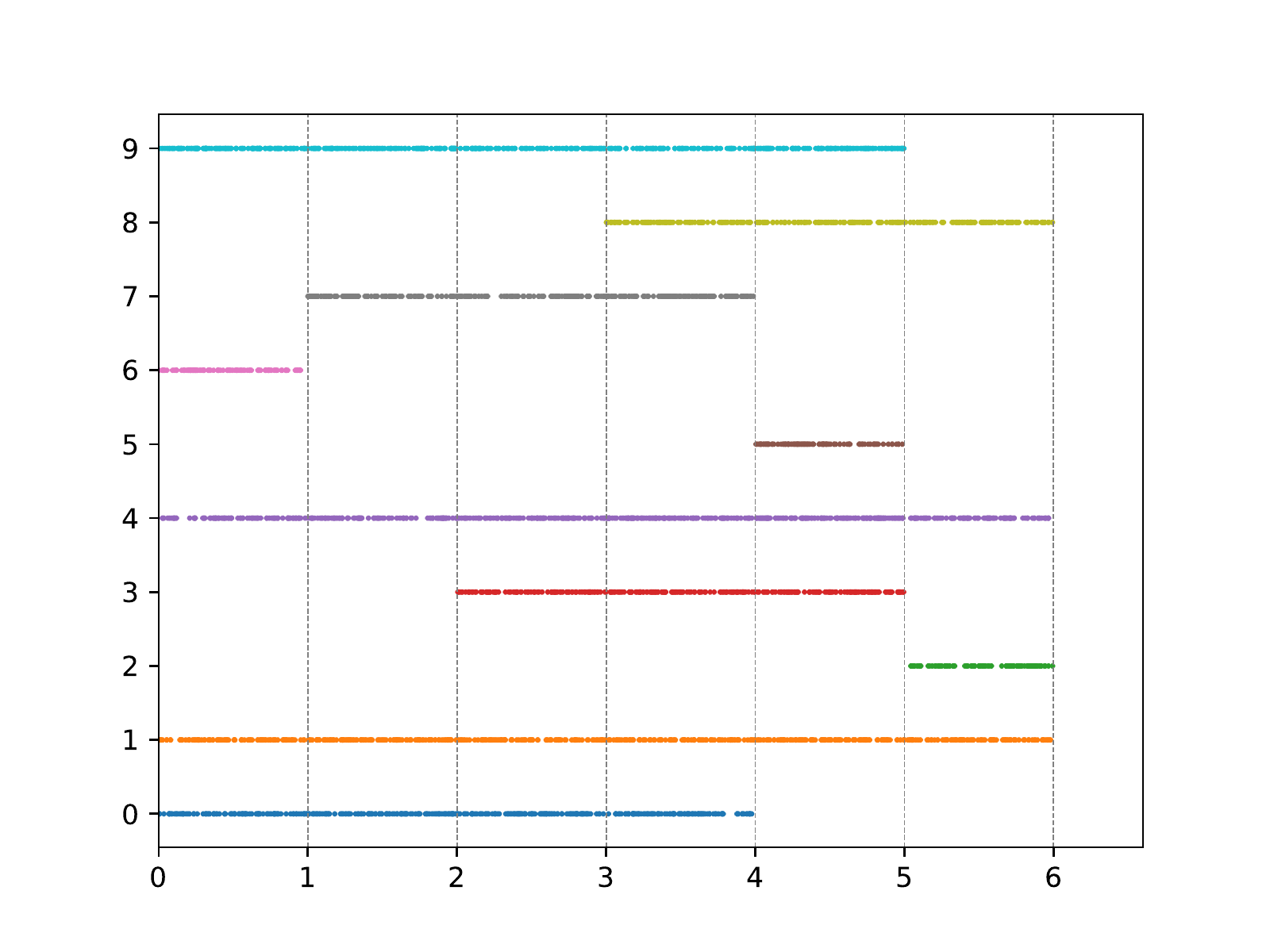}\\
			\mbox{ \;\;\;\; ({\it a}) {Single Novel Class}}
		\end{minipage}
		\begin{minipage}[h]{44mm}
			\centering
			\includegraphics[width=44mm]{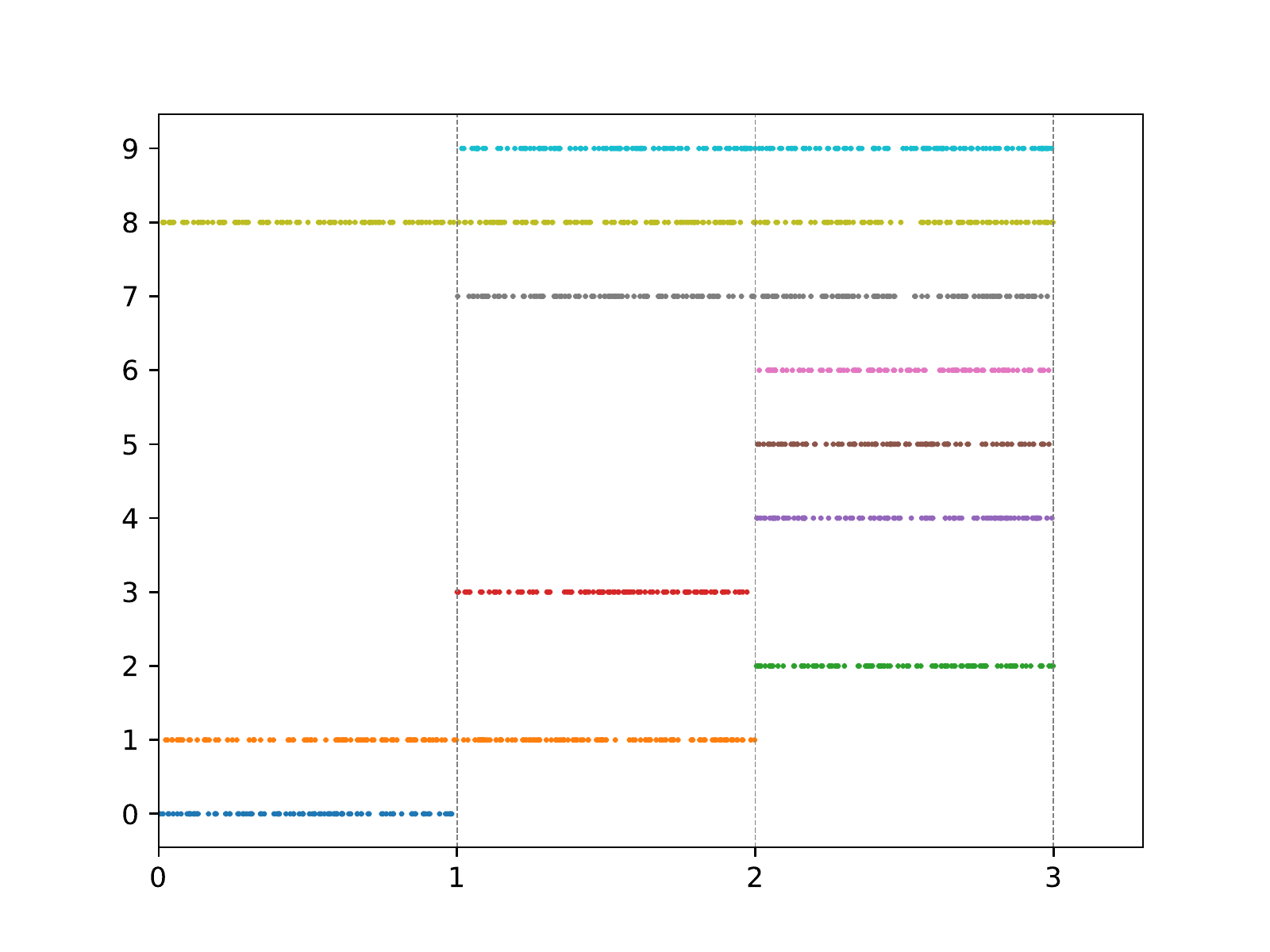}\\
			\mbox{ \;\;\;\; ({\it b}) {Multiple Novel Classes}}
		\end{minipage}
	\end{center}
	\caption{The class distribution of simulated stream on CIFAR-10 dataset as an example. (a) represents the single novel class case, and (b) denotes the multiple novel classes case. The X-axis denotes the streaming data and the Y-axis
		is the class information.}\label{fig:data}
\end{figure}

\begin{table*}[t]{
		\centering
		\caption{Classification of known classes and novel class detection performance over streaming data in single novel class case. The best results are highlighted in bold.}
		\label{tab:tab1}
		\begin{tabular*}{1\textwidth}{@{\extracolsep{\fill}}@{}l|c|c|c|c|c|c|c|c}
			\toprule
			\multirow{2}{*}{Methods} & \multicolumn{4}{c|}{Average NA $\uparrow$} & \multicolumn{4}{c}{Average Macro-F-Measure $\uparrow$}\\
			\cmidrule(l){2-9}
			& MNIST & CIFAR-10 & CIFAR-50 & CIFAR-100 & MNIST & CIFAR-10 & CIFAR-50 & CIFAR-100\\
			\midrule
			Iforest  &    .189$\pm$.021 &    .131$\pm$.023 &    .045$\pm$.006 &    .040$\pm$.004 &    .156$\pm$.023 &    .096$\pm$.011 &    .035$\pm$.005 &    .027$\pm$.004 \\
			One-SVM  &    .211$\pm$.032 &    .136$\pm$.031 &    .043$\pm$.007 &    .039$\pm$.000 &    .135$\pm$.021 &    .090$\pm$.014 &    .032$\pm$.003 &    .024$\pm$.005 \\
			LACU-SVM &    .222$\pm$.034 &    .141$\pm$.020 &    .045$\pm$.005 &    .039$\pm$.004 &    .170$\pm$.018 &    .096$\pm$.010 &    .034$\pm$.004 &    .026$\pm$.006 \\
			SENC-MAS &    .203$\pm$.030 &    .134$\pm$.033 &    .043$\pm$.006 &    .038$\pm$.001 &    .149$\pm$.031 &    .082$\pm$.013 &    .031$\pm$.003 &    .022$\pm$.005 \\
			\midrule
			ODIN-CNN &    .293$\pm$.011 &    .194$\pm$.020 &    .068$\pm$.001 &    .034$\pm$.027 &    .323$\pm$.029 &    .156$\pm$.033 &    .031$\pm$.002 &    .055$\pm$.045 \\
			CFO      &    .276$\pm$.008 &    .202$\pm$.015 &    .037$\pm$.001 &    .022$\pm$.008 &    .292$\pm$.022 &    .167$\pm$.025 &    .045$\pm$.002 &    .033$\pm$.013 \\
			CPE      &    .298$\pm$.013 &    .250$\pm$.020 &    .055$\pm$.001 &    .047$\pm$.017 &    .337$\pm$.034 &    .281$\pm$.033 &    .109$\pm$.002 &    .087$\pm$.031 \\
			\midrule
			DEC      &    .289$\pm$.058 &    .245$\pm$.152 &    .035$\pm$.001 &    .027$\pm$.001 &    .357$\pm$.081 &    .261$\pm$.344 &    .048$\pm$.001 &    .018$\pm$.001 \\
			\midrule
			CILF     & \bf.371$\pm$.022 & \bf.288$\pm$.034 & \bf.092$\pm$.008 & \bf.055$\pm$.004 & \bf.365$\pm$.012 & \bf.302$\pm$.033 & \bf.158$\pm$.005 & \bf.092$\pm$.008 \\
		\end{tabular*}
		\begin{tabular*}{1\textwidth}{@{\extracolsep{\fill}}@{}l|c|c|c|c|c|c|c|c}
			\toprule
			\multirow{2}{*}{Methods} & \multicolumn{4}{c|}{Average Micro-F-Measure $\uparrow$} & \multicolumn{4}{c}{Average AUROC $ \uparrow$}\\
			\cmidrule(l){2-9}
			& MNIST & CIFAR-10 & CIFAR-50 & CIFAR-100 & MNIST & CIFAR-10 & CIFAR-50 & CIFAR-100\\
			\midrule
			Iforest  &    .234$\pm$.036 &    .142$\pm$.021 &    .066$\pm$.008 &    .052$\pm$.008 &    .083$\pm$.014 &    .140$\pm$.343 &    .077$\pm$.014 &    .105$\pm$.009 \\
			One-SVM  &    .214$\pm$.074 &    .147$\pm$.031 &    .064$\pm$.008 &    .046$\pm$.007 &    .117$\pm$.009 &    .101$\pm$.021 &    .096$\pm$.011 &    .074$\pm$.020 \\
			LACU-SVM &    .258$\pm$.042 &    .148$\pm$.022 &    .064$\pm$.007 &    .050$\pm$.008 &    .123$\pm$.013 &    .068$\pm$.012 &    .089$\pm$.016 &    .117$\pm$.081 \\
			SENC-MAS &    .216$\pm$.065 &    .140$\pm$.036 &    .062$\pm$.008 &    .044$\pm$.006 &    .104$\pm$.036 &    .082$\pm$.026 &    .043$\pm$.010 &    .079$\pm$.096 \\
			\midrule
			ODIN-CNN &    .285$\pm$.021 &    .188$\pm$.040 &    .035$\pm$.001 &    .058$\pm$.041 &    .259$\pm$.189 &    .185$\pm$.063 &    .102$\pm$.045 &    .123$\pm$.096 \\
			CFO      &    .351$\pm$.016 &    .304$\pm$.030 &    .054$\pm$.001 &    .047$\pm$.017 &    .208$\pm$.180 &    .162$\pm$.073 &    .076$\pm$.030 &    .102$\pm$.092 \\
			CPE      &    .336$\pm$.026 &    .299$\pm$.040 &    .110$\pm$.001 &    .092$\pm$.033 &    .270$\pm$.184 &    .193$\pm$.060 &    .114$\pm$.037 &    .119$\pm$.100 \\
			\midrule
			DEC      &    .323$\pm$.073 &    .289$\pm$.305 &    .064$\pm$.001 &    .025$\pm$.001 &    .264$\pm$.187 &    .190$\pm$.058 &    .108$\pm$.036 &    .114$\pm$.097 \\
			\midrule
			CILF     & \bf.355$\pm$.025 & \bf.310$\pm$.025 & \bf.122$\pm$.008 & \bf.103$\pm$.008 & \bf.317$\pm$.018 & \bf.255$\pm$.039 & \bf.125$\pm$.012 & \bf.130$\pm$.027 \\
			\bottomrule
	\end{tabular*}}
\end{table*}

\begin{figure*}[t]
	\begin{center}
	\begin{minipage}[h]{28mm}
		\centering
		\includegraphics[width=28mm]{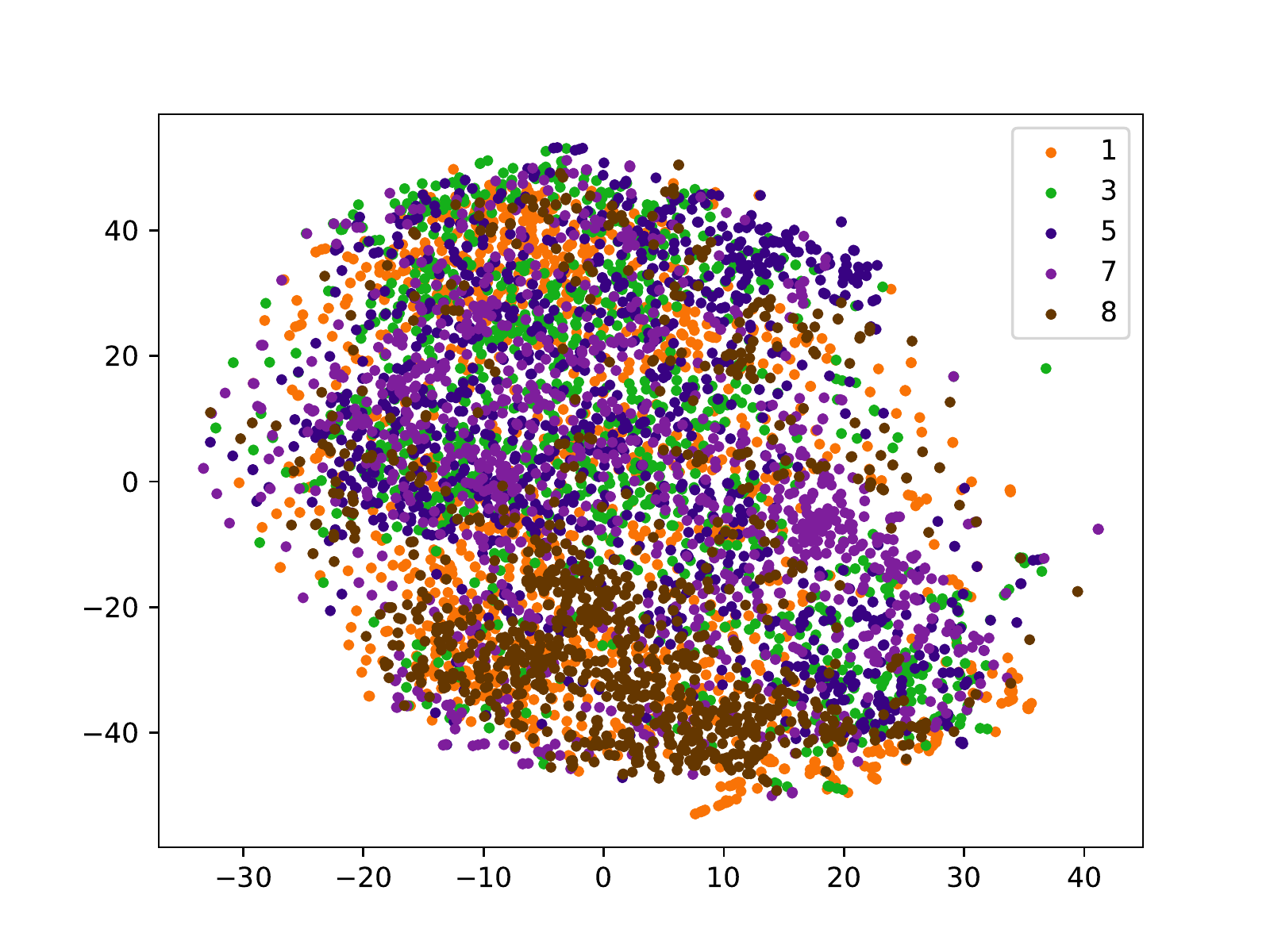}\\
		\mbox{ ({\it a-1}) {Original-1}}
	\end{minipage}
	\begin{minipage}[h]{28mm}
		\centering
		\includegraphics[width=28mm]{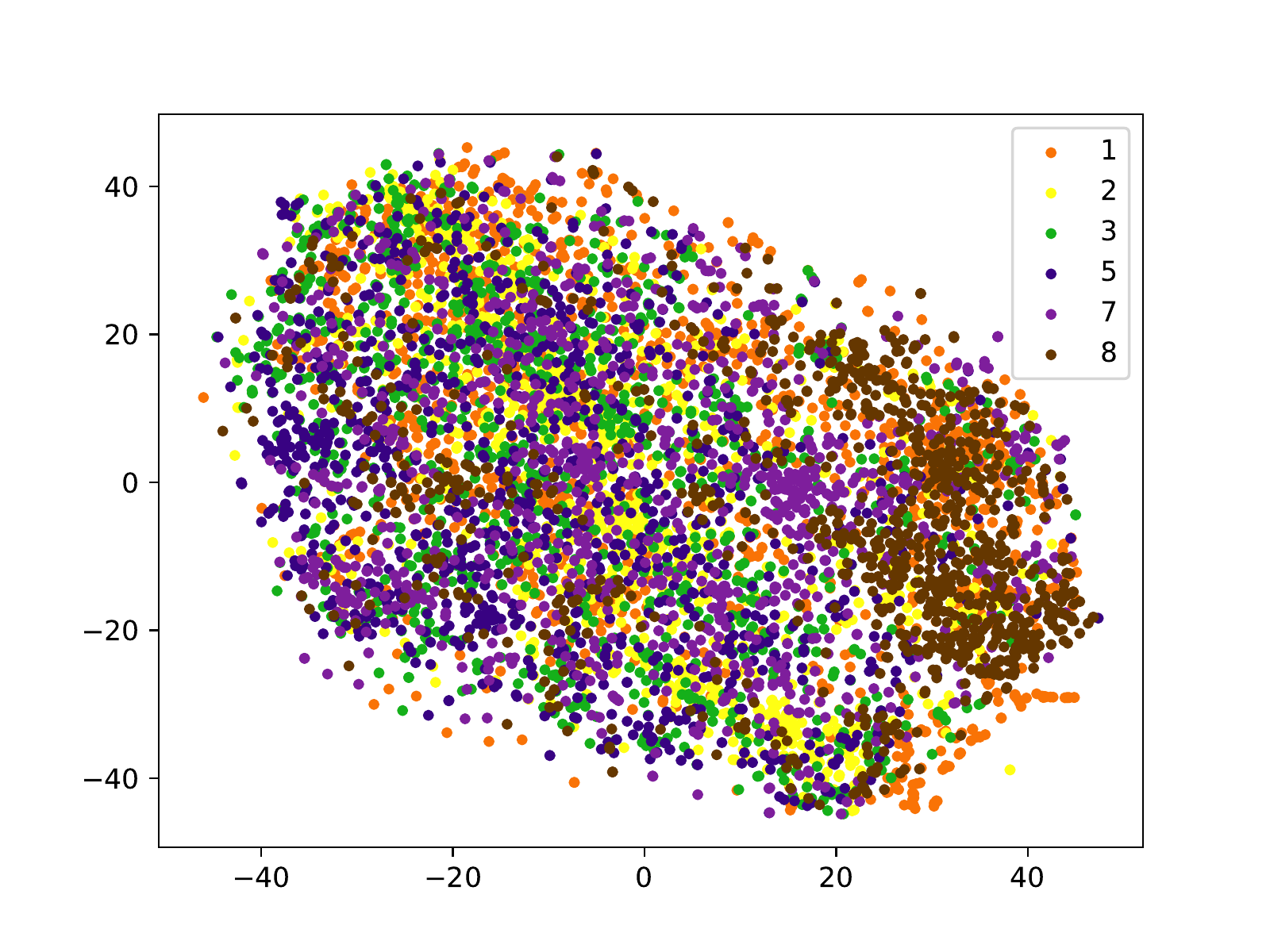}\\
		\mbox{ ({\it a-2}) {Original-2}}
	\end{minipage}
	\begin{minipage}[h]{28mm}
		\centering
		\includegraphics[width=28mm]{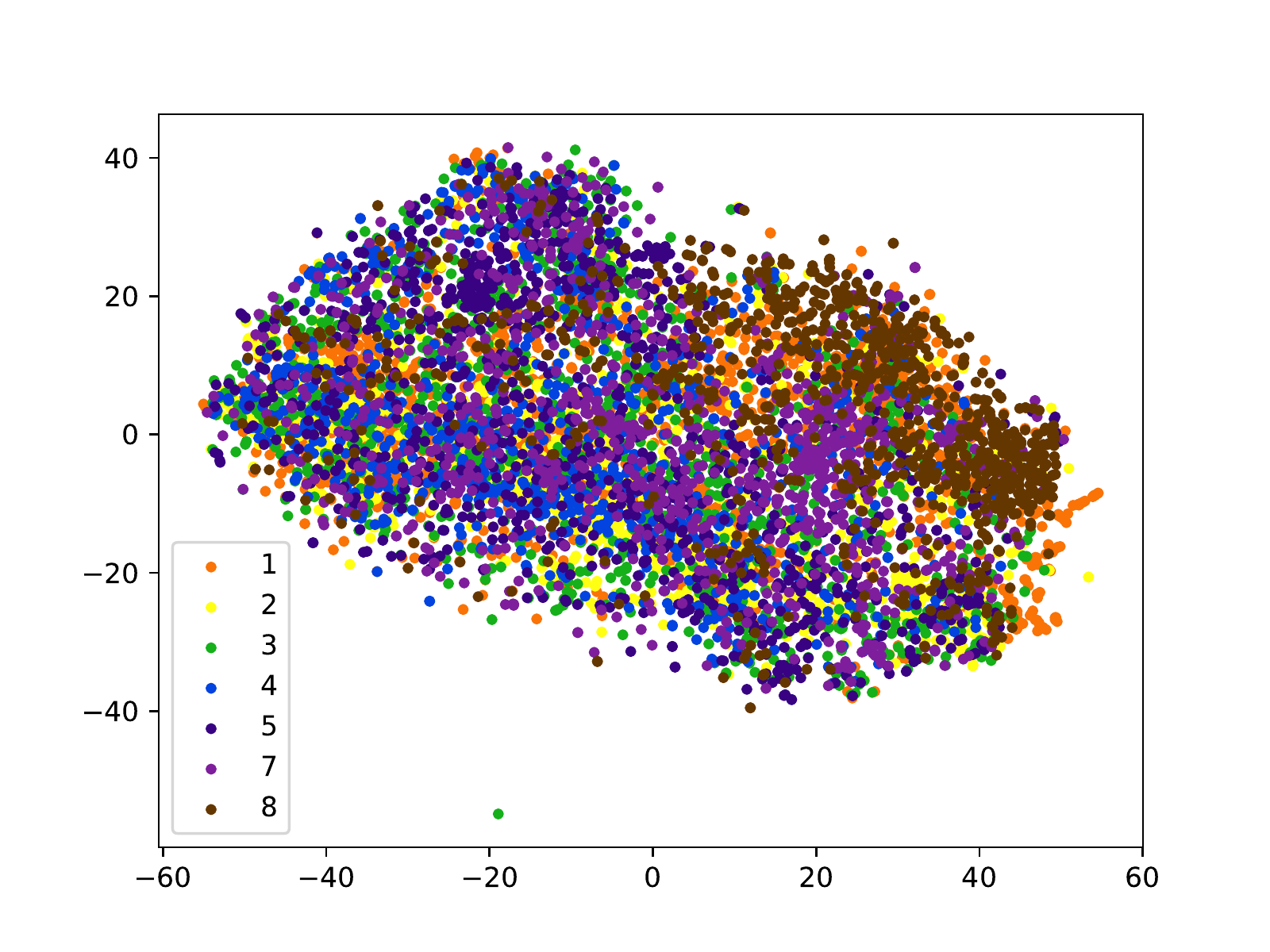}\\
		\mbox{ ({\it a-3}) {Original-3}}
	\end{minipage}
	\begin{minipage}[h]{28mm}
		\centering
		\includegraphics[width=28mm]{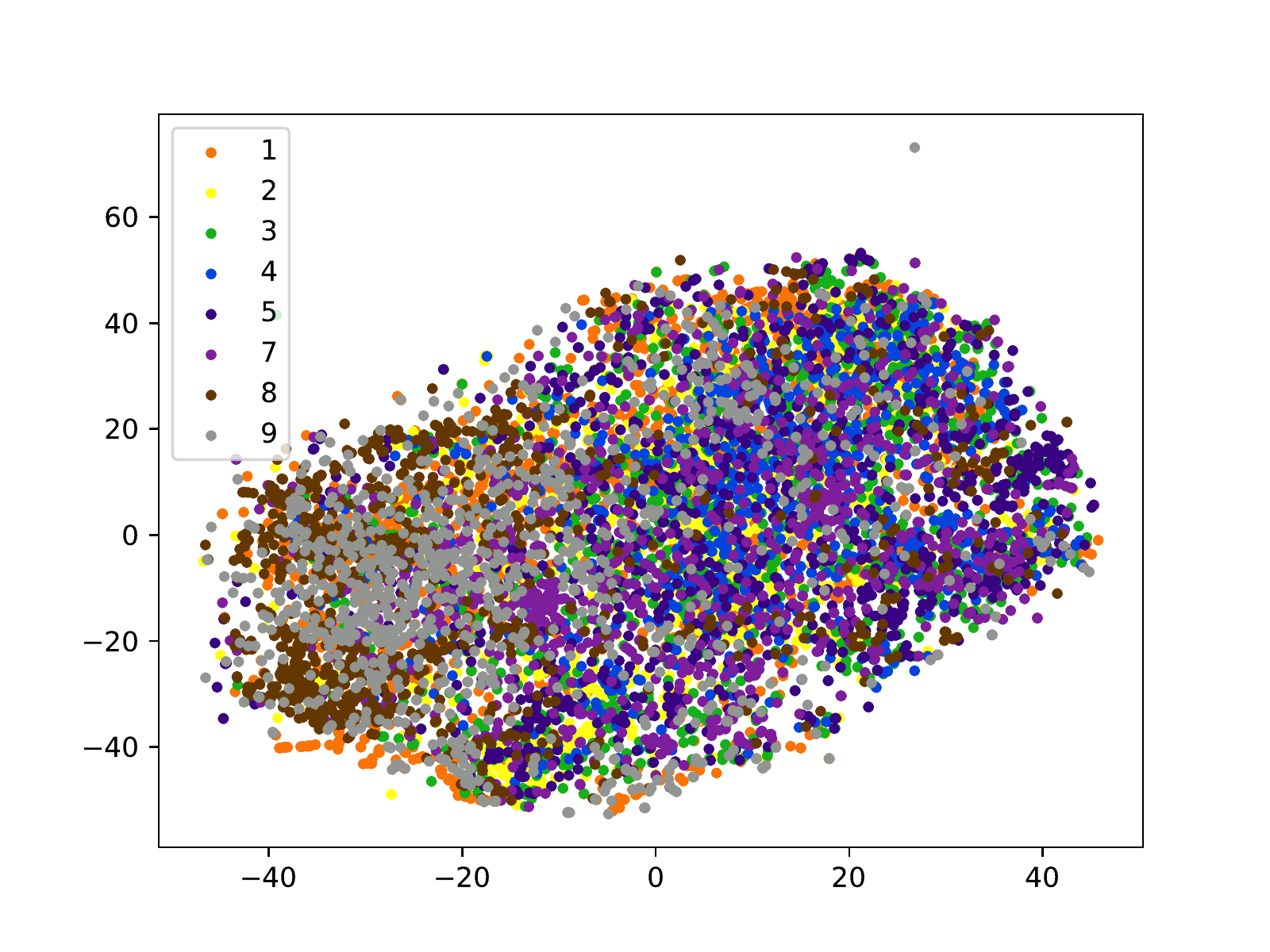}\\
		\mbox{ ({\it a-4}) {Original-4}}
	\end{minipage}
	\begin{minipage}[h]{28mm}
		\centering
		\includegraphics[width=28mm]{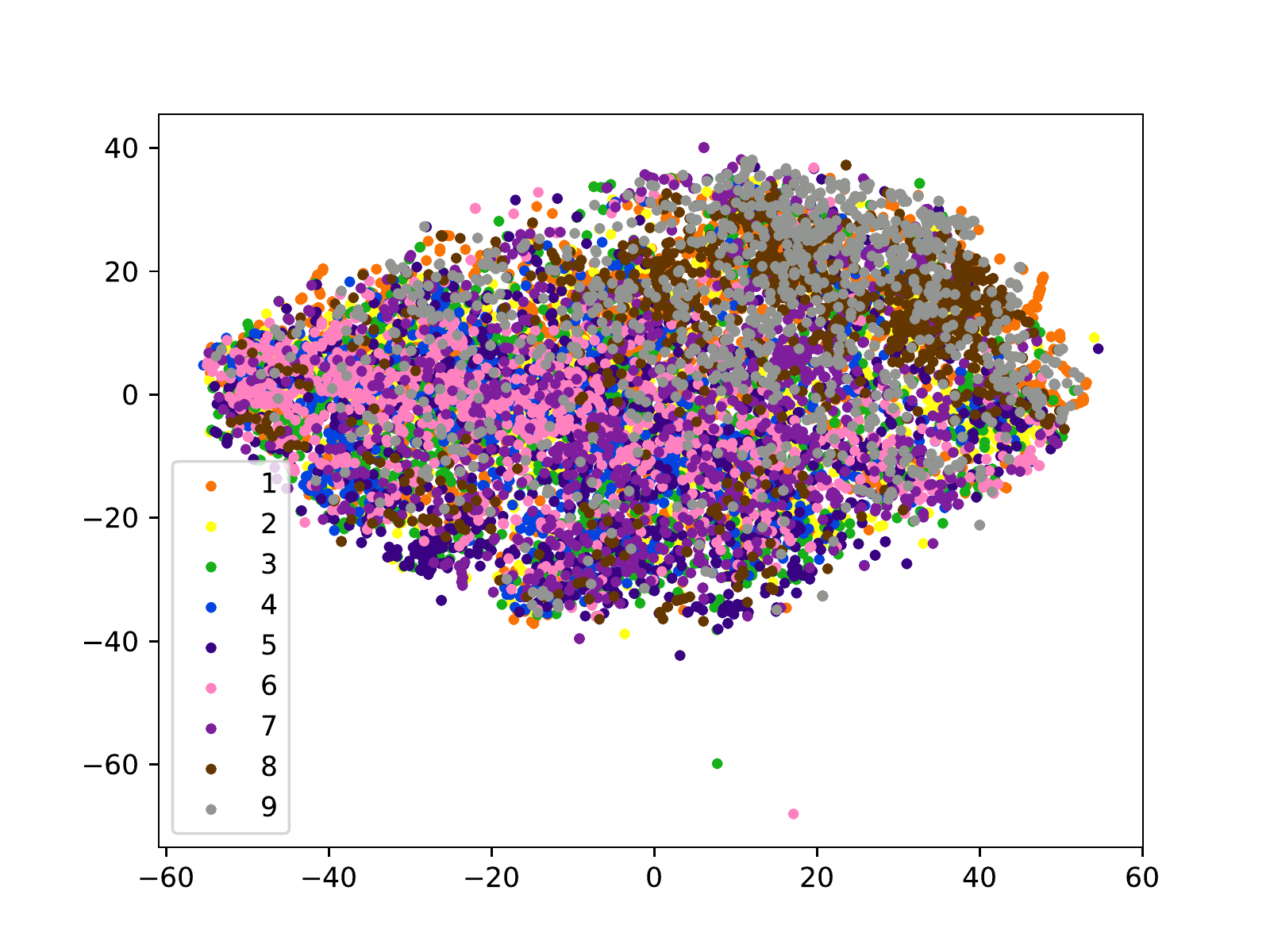}\\
		\mbox{ ({\it a-5}) {Original-5}}
	\end{minipage}
	\begin{minipage}[h]{28mm}
		\centering
		\includegraphics[width=28mm]{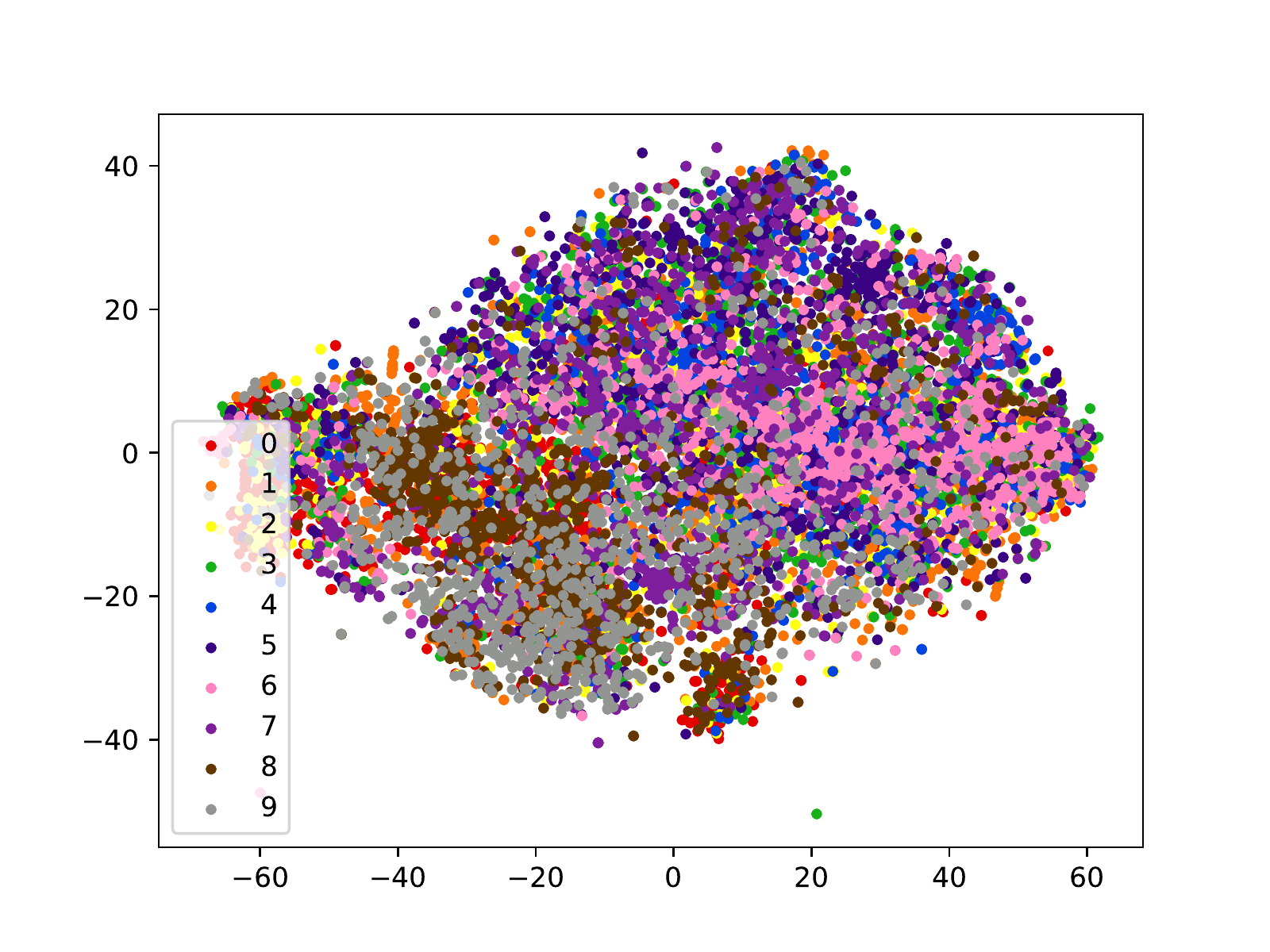}\\
		\mbox{ ({\it a-6}) {Original-6}}
	\end{minipage}
	
	\begin{minipage}[h]{28mm}
		\centering
		\includegraphics[width=28mm]{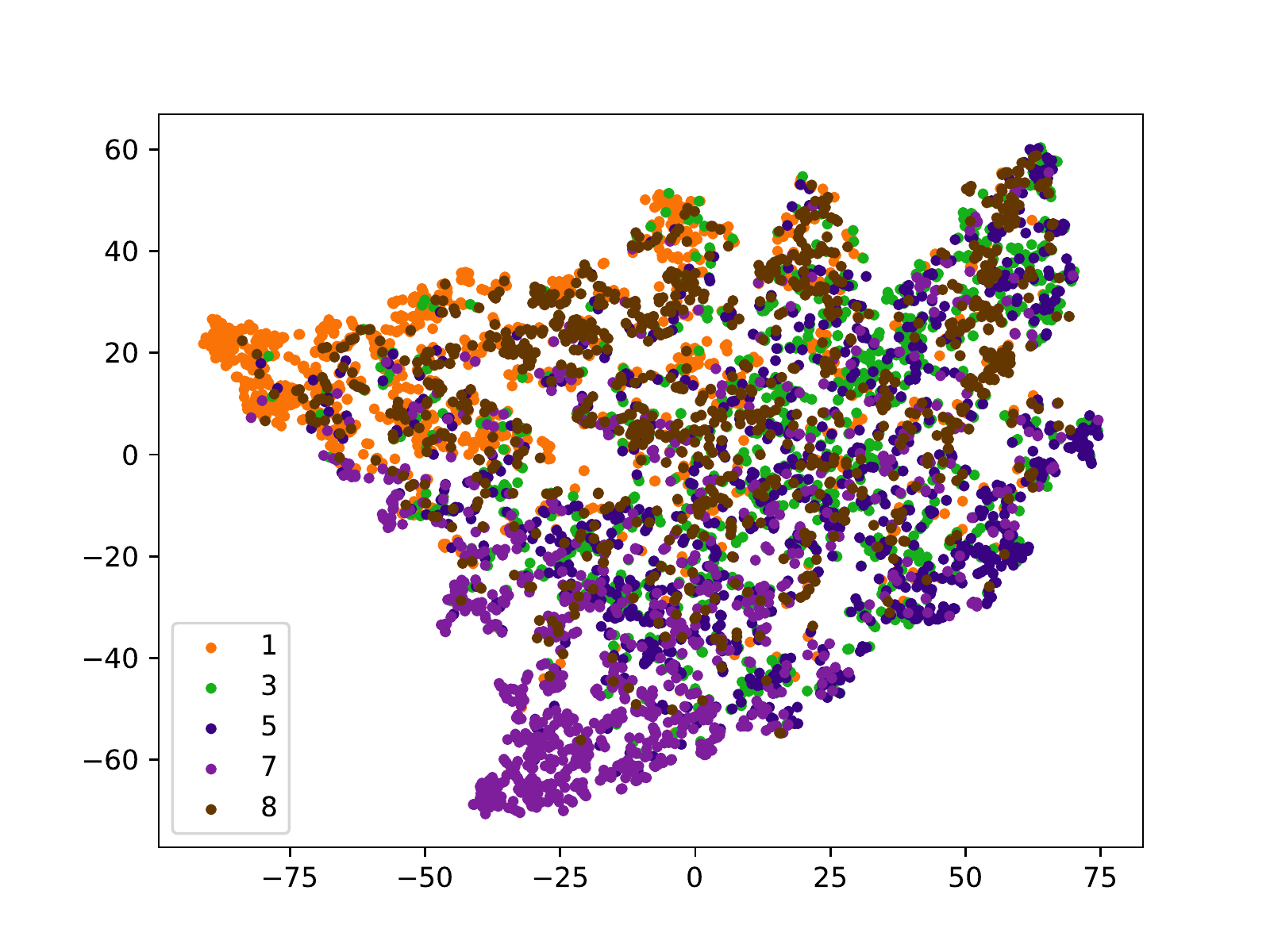}\\
		\mbox{ ({\it b-1}) {CPE-1}}
	\end{minipage}
	\begin{minipage}[h]{28mm}
		\centering
		\includegraphics[width=28mm]{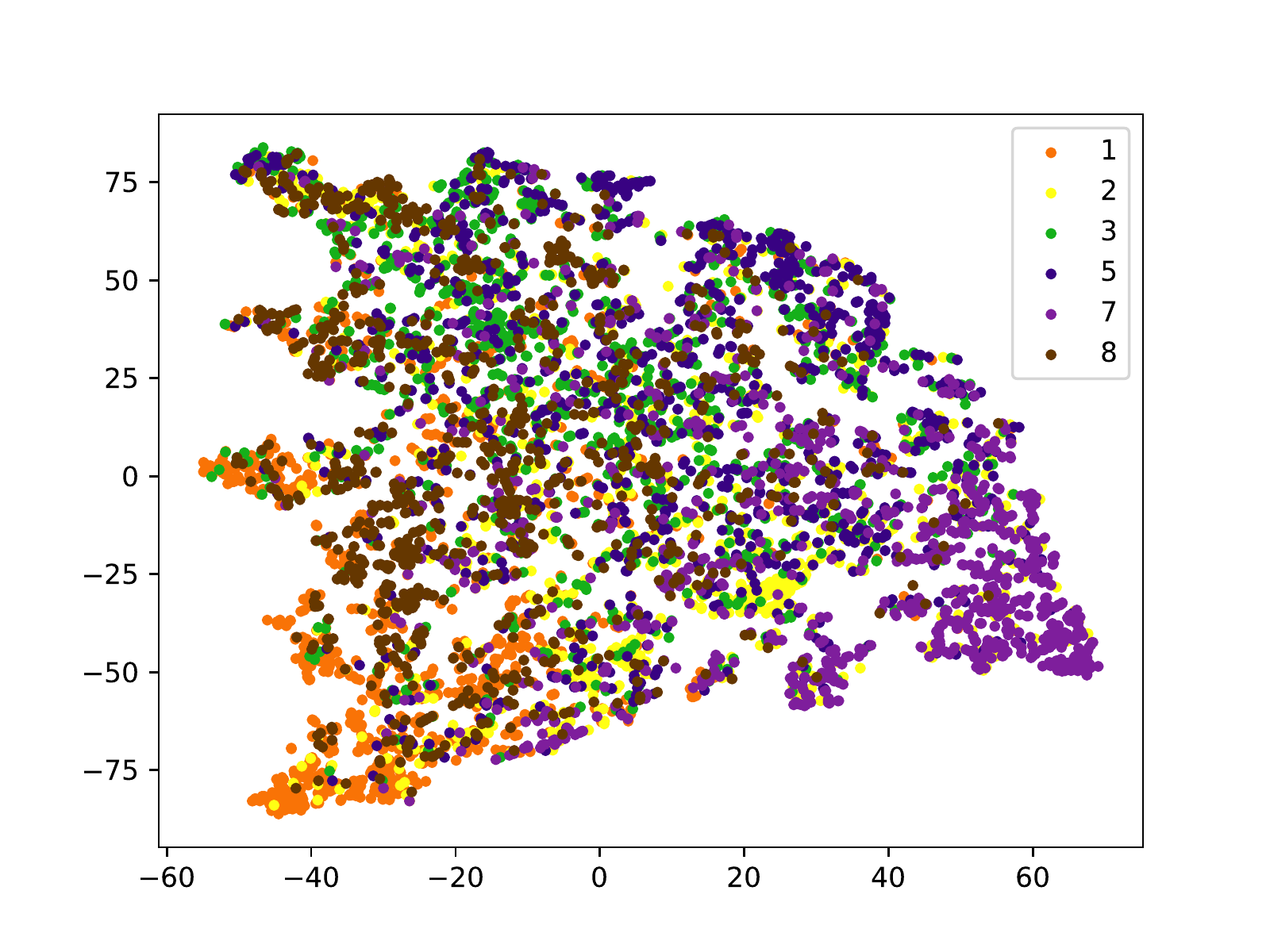}\\
		\mbox{ ({\it b-2}) {CPE-2}}
	\end{minipage}
	\begin{minipage}[h]{28mm}
		\centering
		\includegraphics[width=28mm]{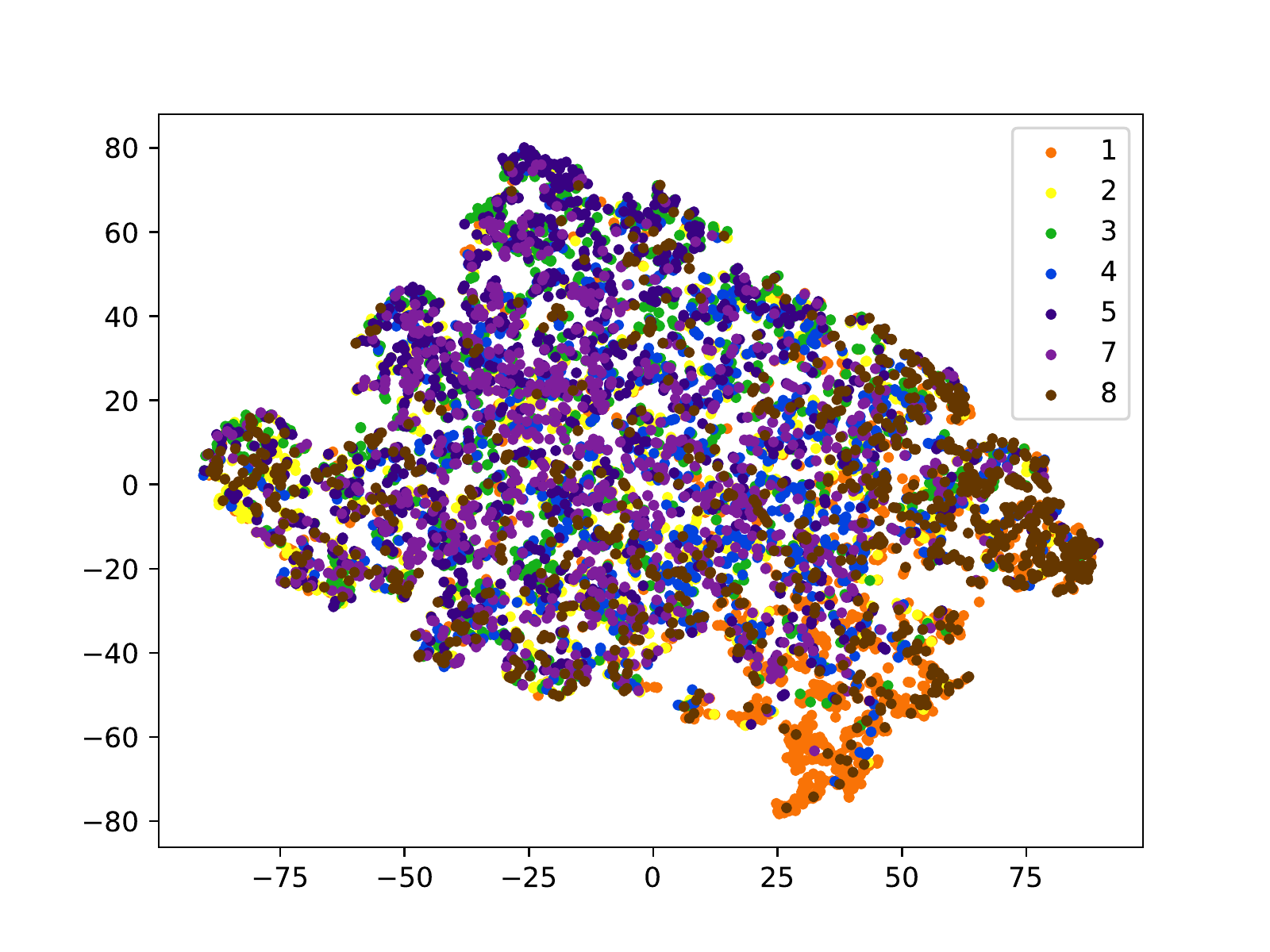}\\
		\mbox{ ({\it b-3}) {CPE-3}}
	\end{minipage}
	\begin{minipage}[h]{28mm}
		\centering
		\includegraphics[width=28mm]{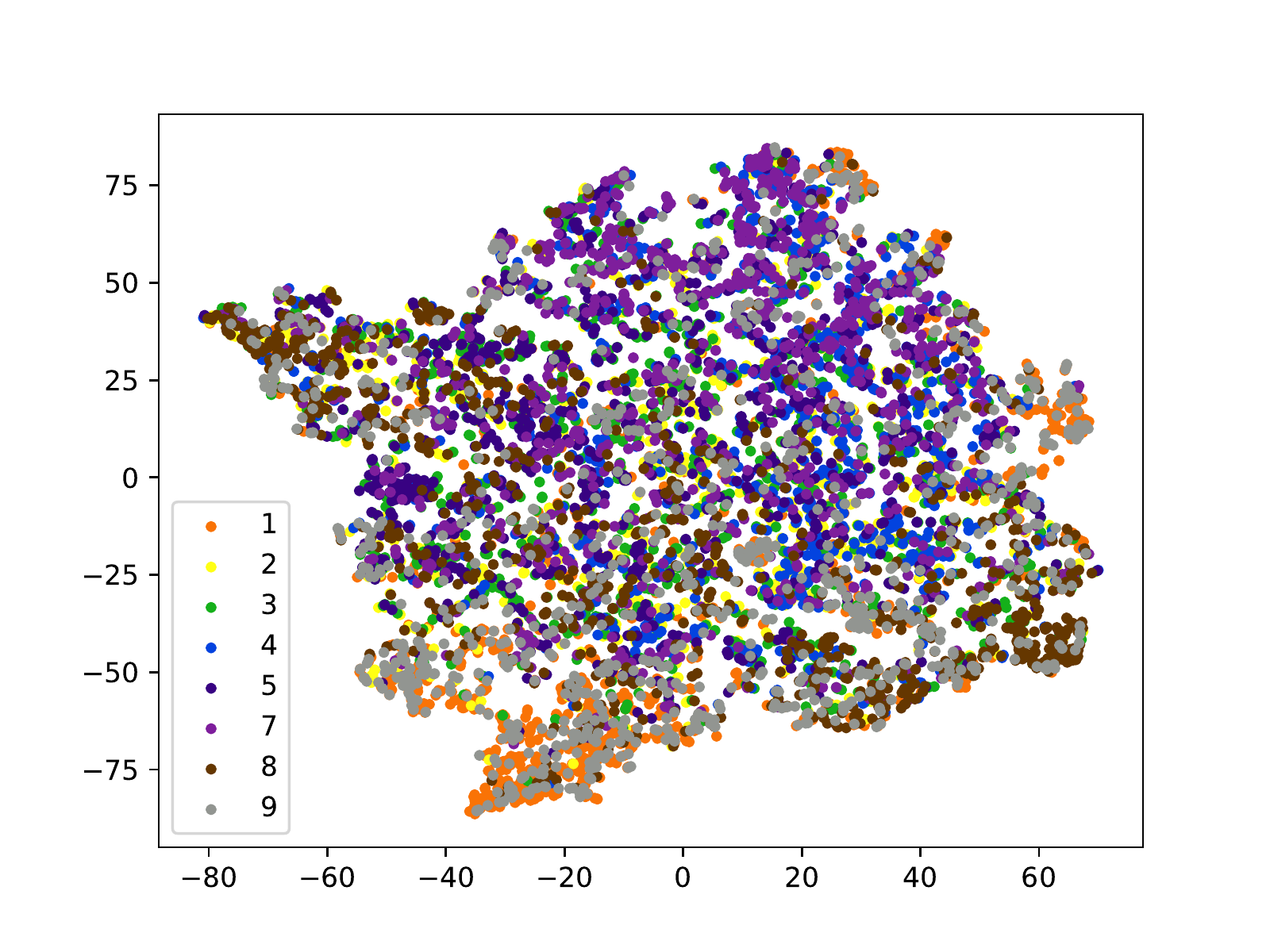}\\
		\mbox{ ({\it b-4}) {CPE-4}}
	\end{minipage}
	\begin{minipage}[h]{28mm}
		\centering
		\includegraphics[width=28mm]{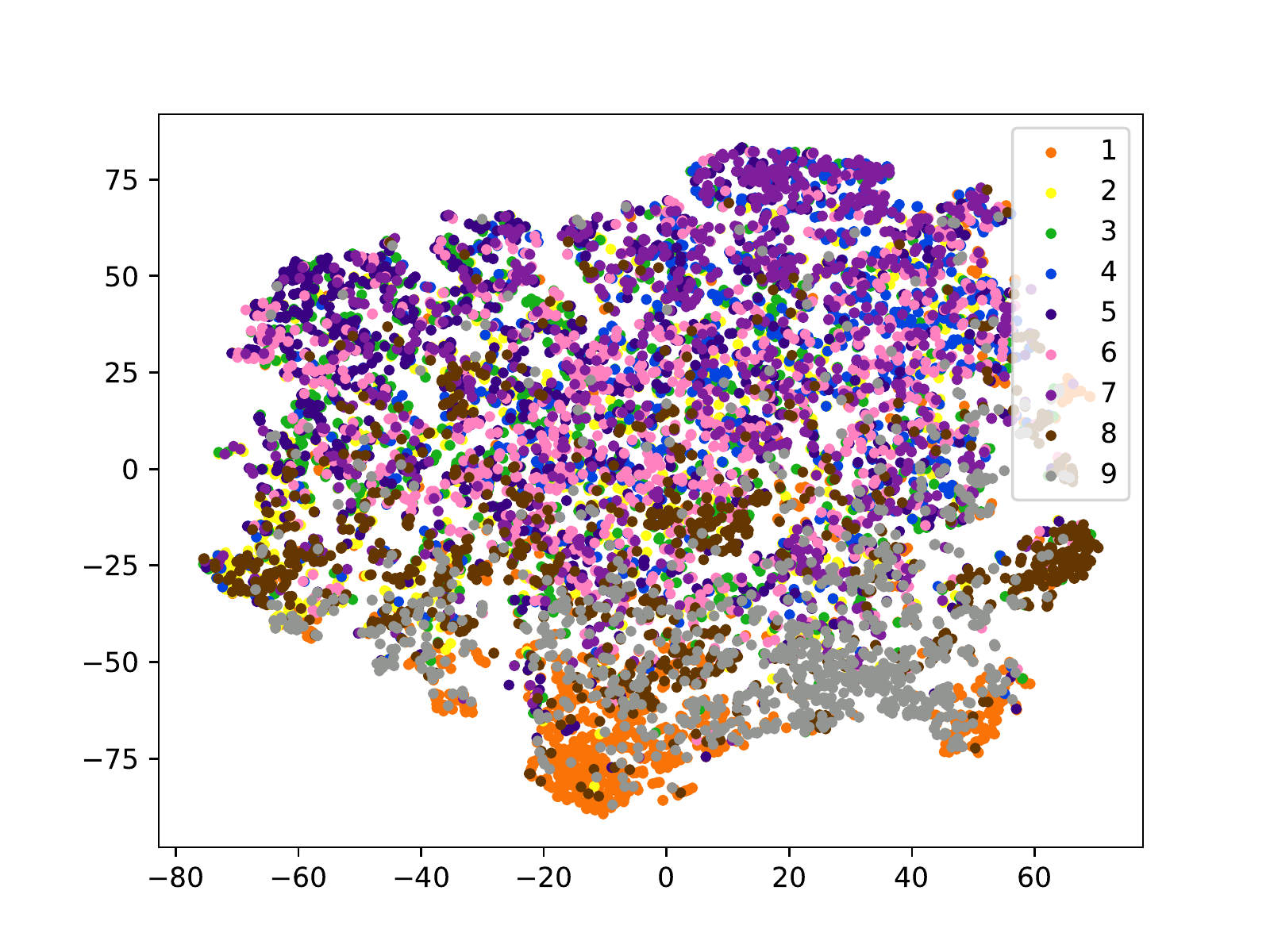}\\
		\mbox{ ({\it b-5}) {CPE-5}}
	\end{minipage}
	\begin{minipage}[h]{28mm}
		\centering
		\includegraphics[width=28mm]{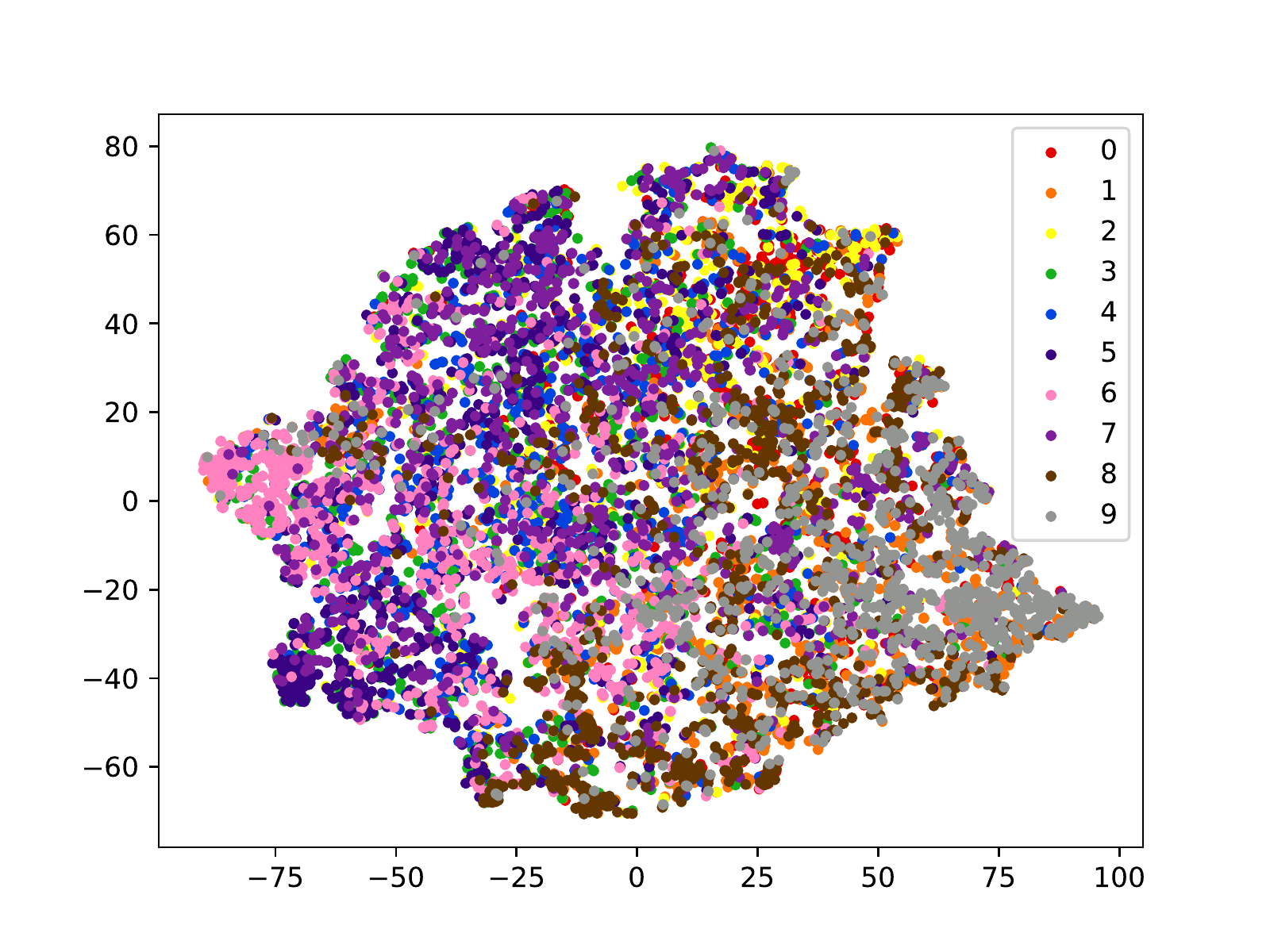}\\
		\mbox{ ({\it b-6}) {CPE-6}}
	\end{minipage}

	\begin{minipage}[h]{28mm}
		\centering
		\includegraphics[width=28mm]{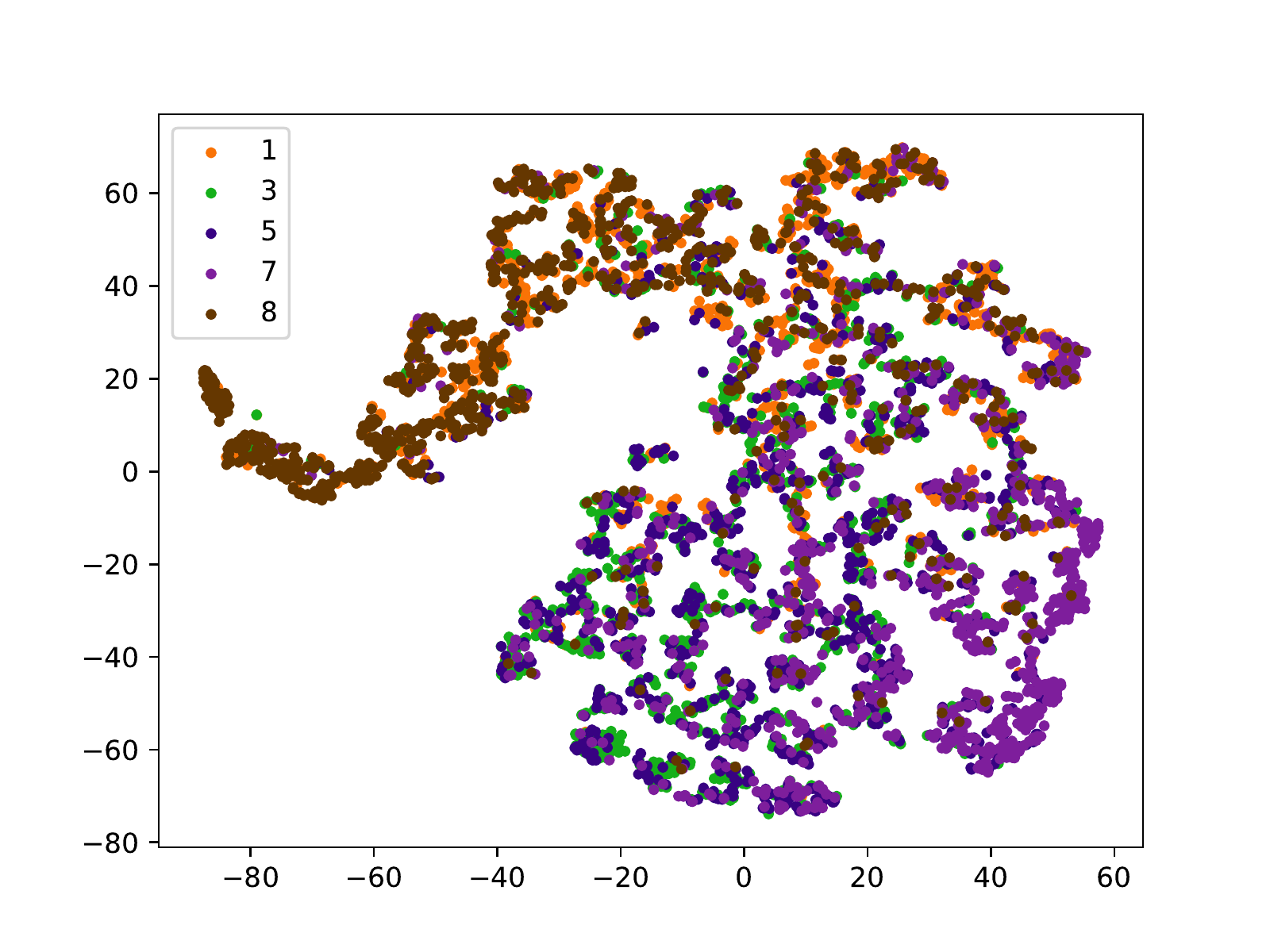}\\
		\mbox{ ({\it c-1}) {DEC-1}}
	\end{minipage}
	\begin{minipage}[h]{28mm}
		\centering
		\includegraphics[width=28mm]{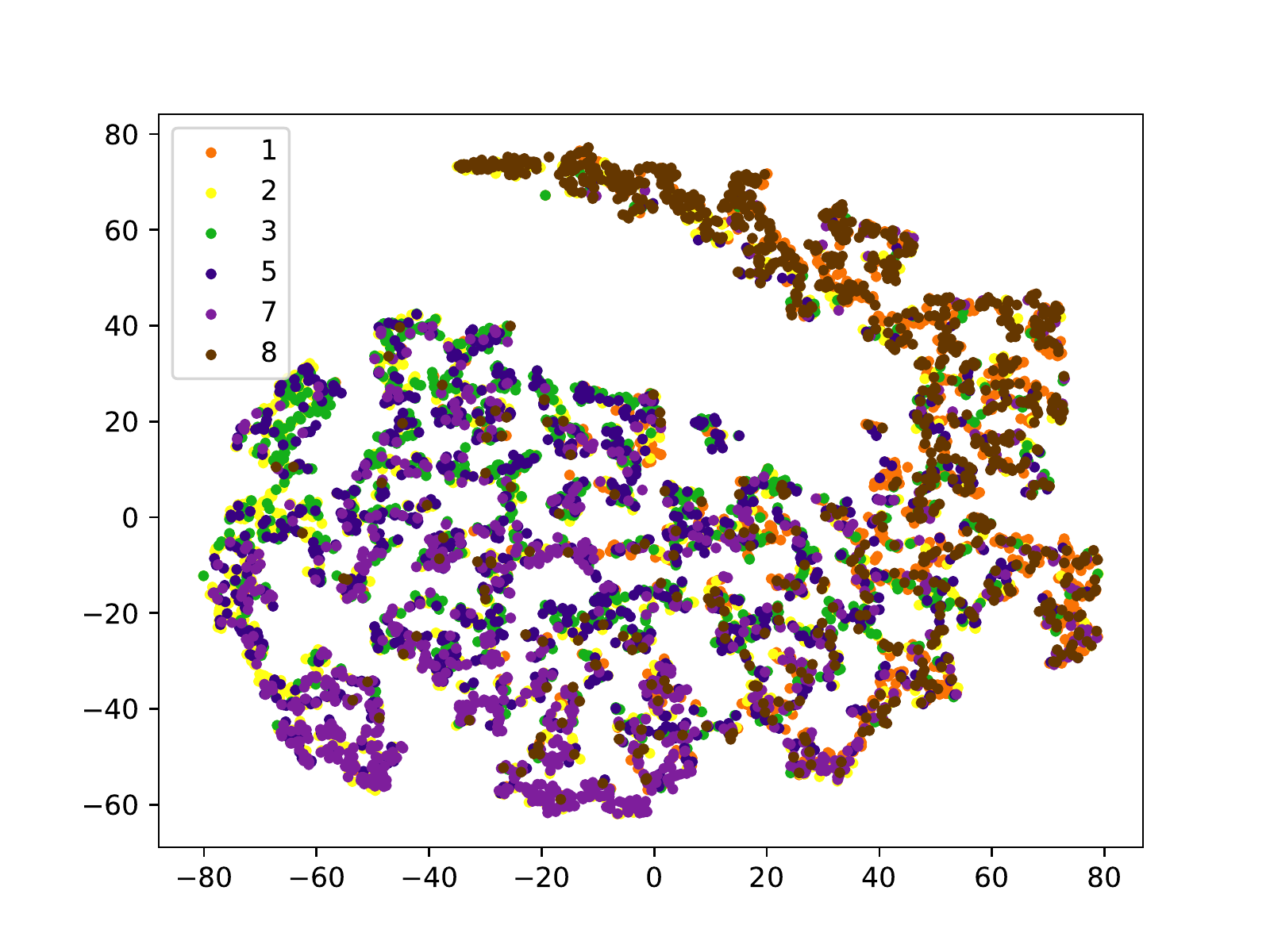}\\
		\mbox{ ({\it c-2}) {DEC-2}}
	\end{minipage}
	\begin{minipage}[h]{28mm}
		\centering
		\includegraphics[width=28mm]{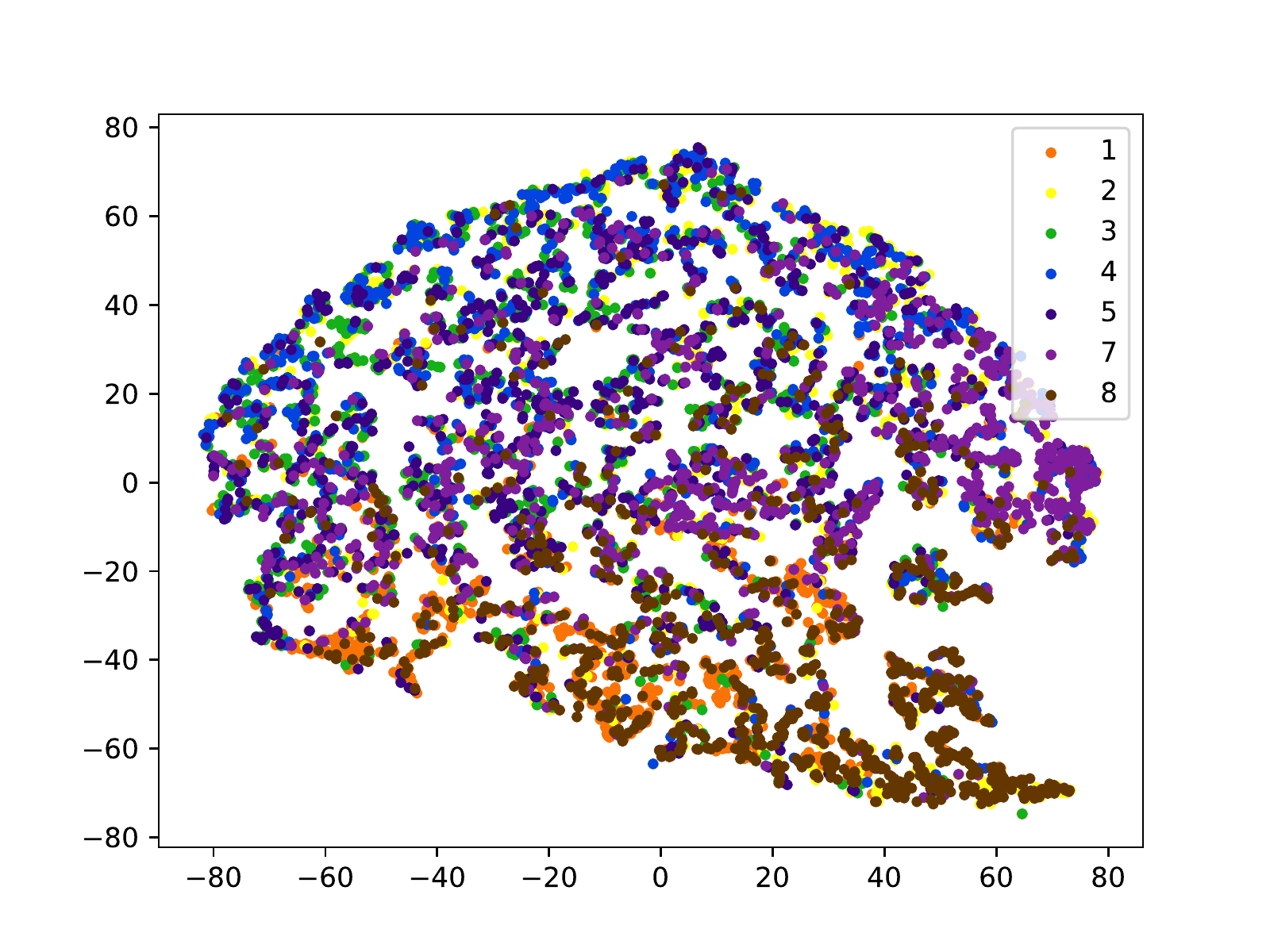}\\
		\mbox{ ({\it c-3}) {DEC-3}}
	\end{minipage}
	\begin{minipage}[h]{28mm}
		\centering
		\includegraphics[width=28mm]{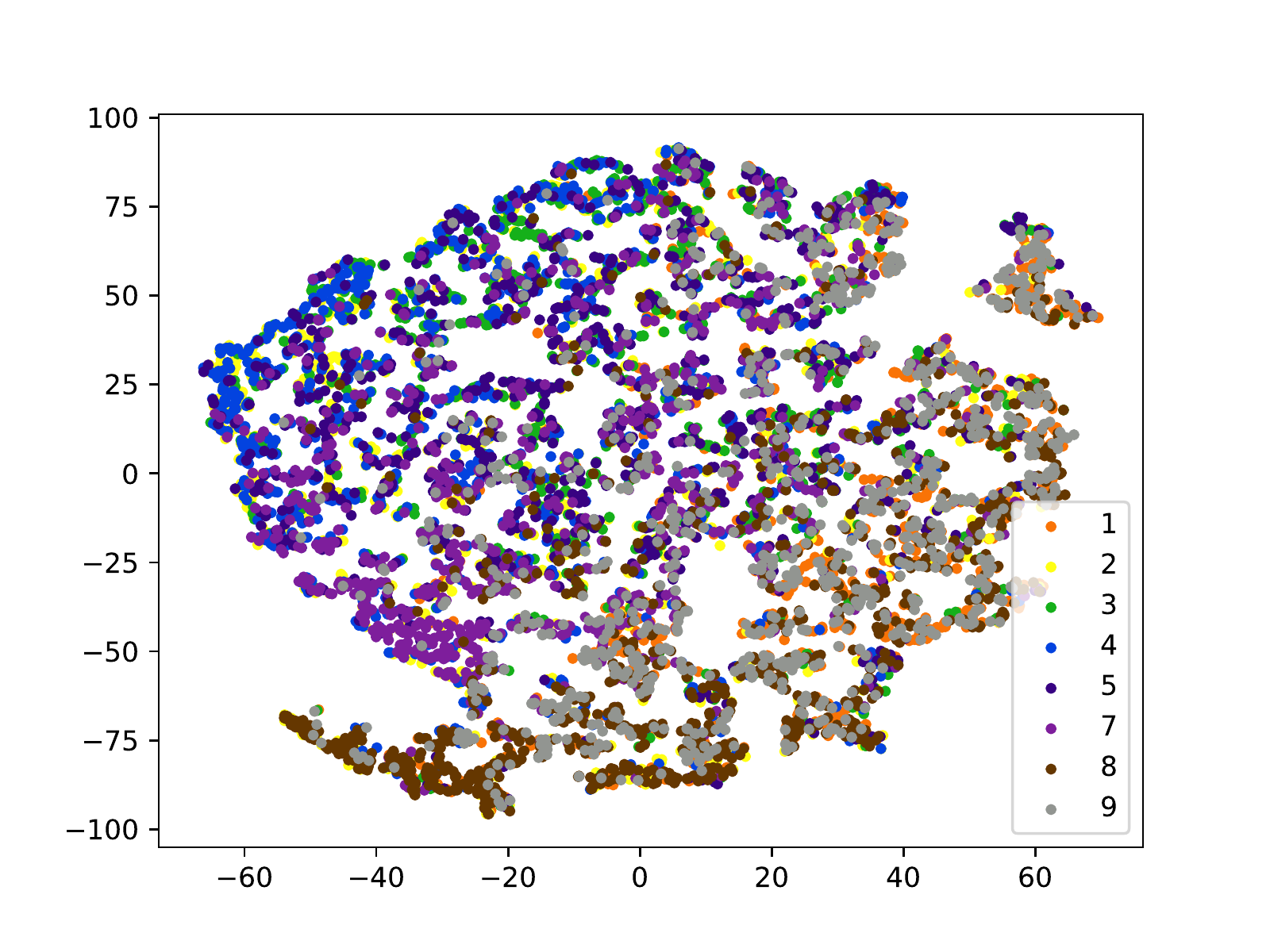}\\
		\mbox{ ({\it c-4}) {DEC-4}}
	\end{minipage}
	\begin{minipage}[h]{28mm}
		\centering
		\includegraphics[width=28mm]{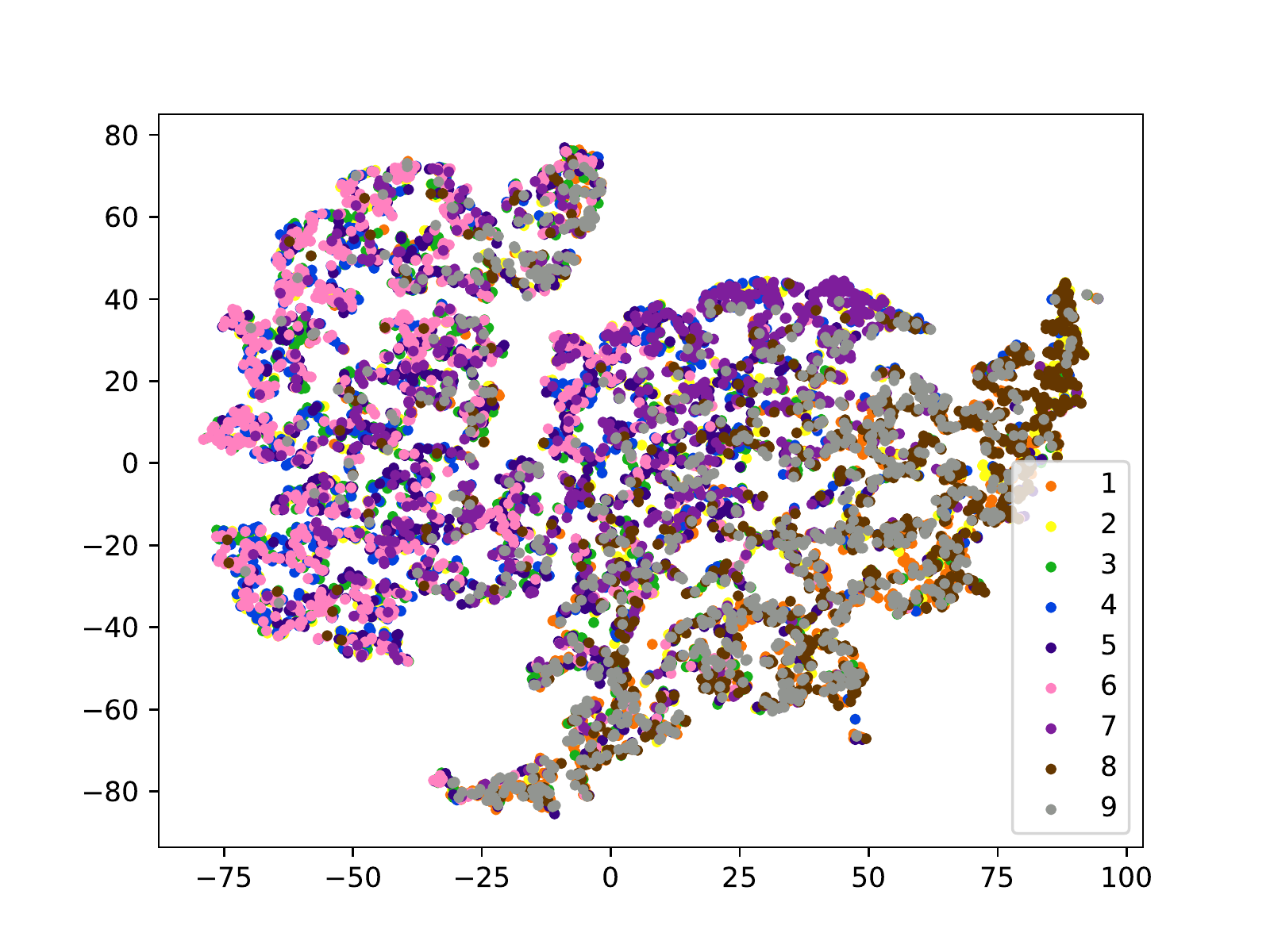}\\
		\mbox{ ({\it c-5}) {DEC-5}}
	\end{minipage}
	\begin{minipage}[h]{28mm}
		\centering
		\includegraphics[width=28mm]{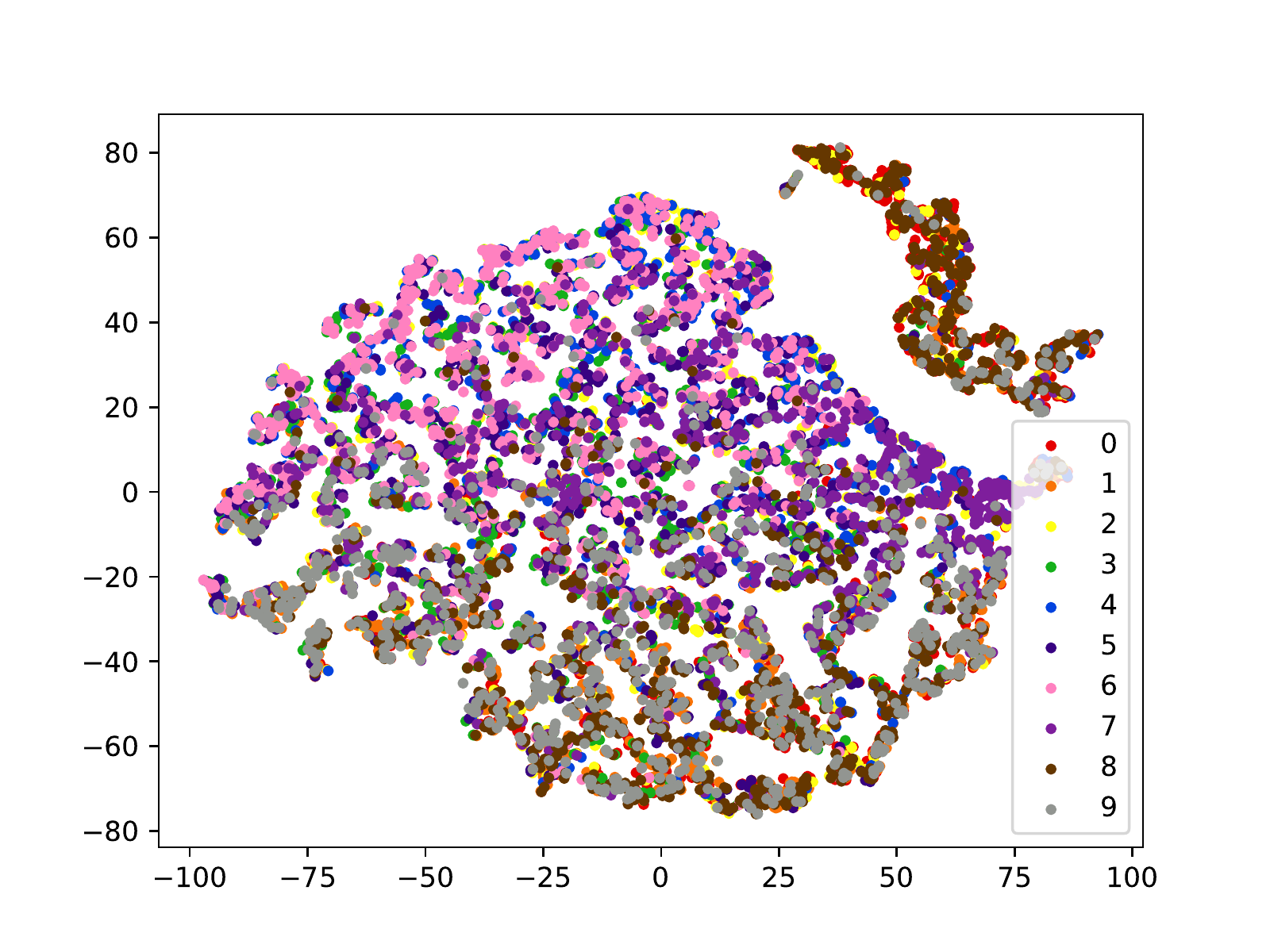}\\
		\mbox{ ({\it c-6}) {DEC-6}}
	\end{minipage}
	\begin{minipage}[h]{28mm}
		\centering
		\includegraphics[width=28mm]{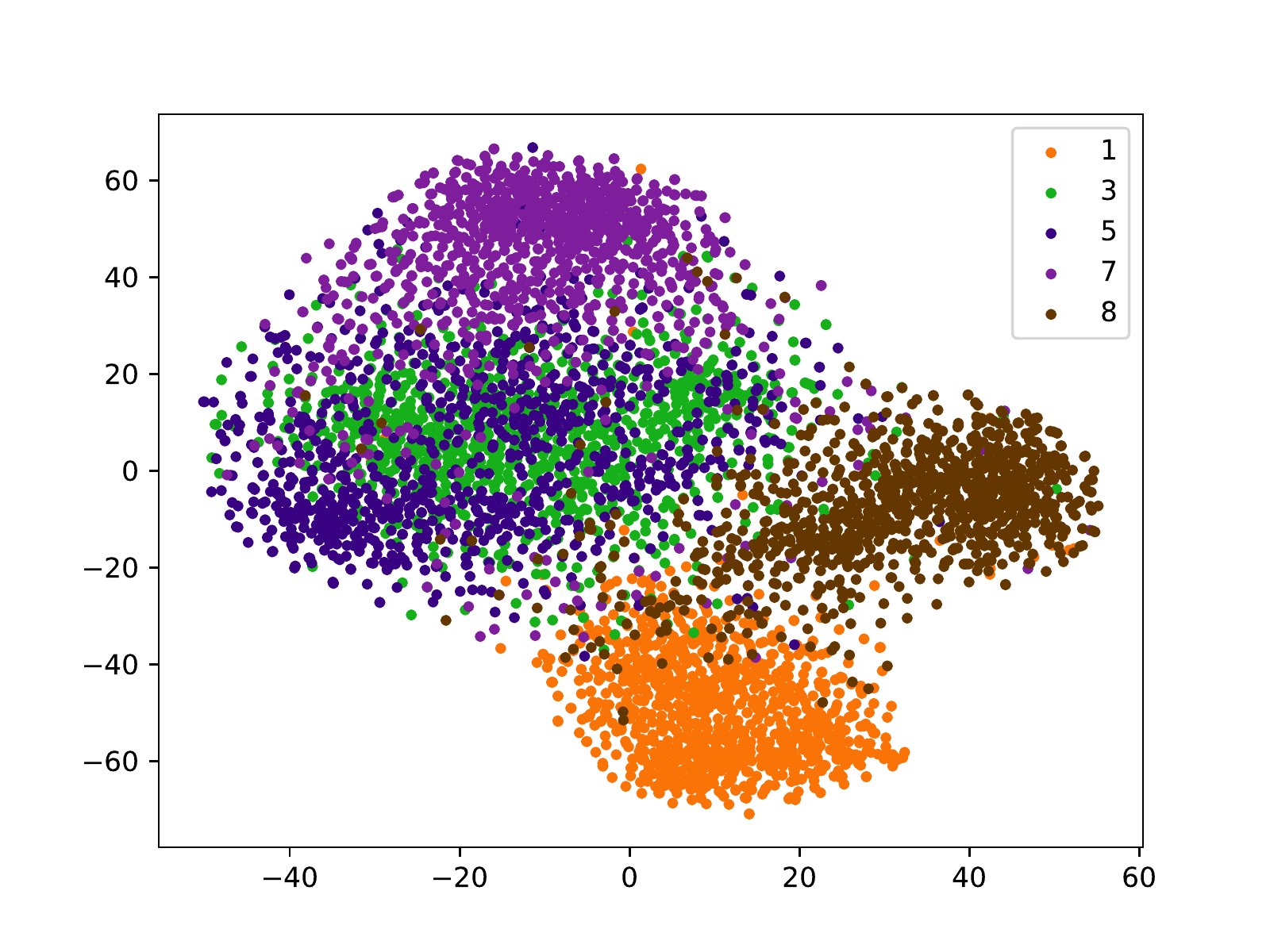}\\
		\mbox{ ({\it d-1}) {CILF-1}}
	\end{minipage}
	\begin{minipage}[h]{28mm}
		\centering
		\includegraphics[width=28mm]{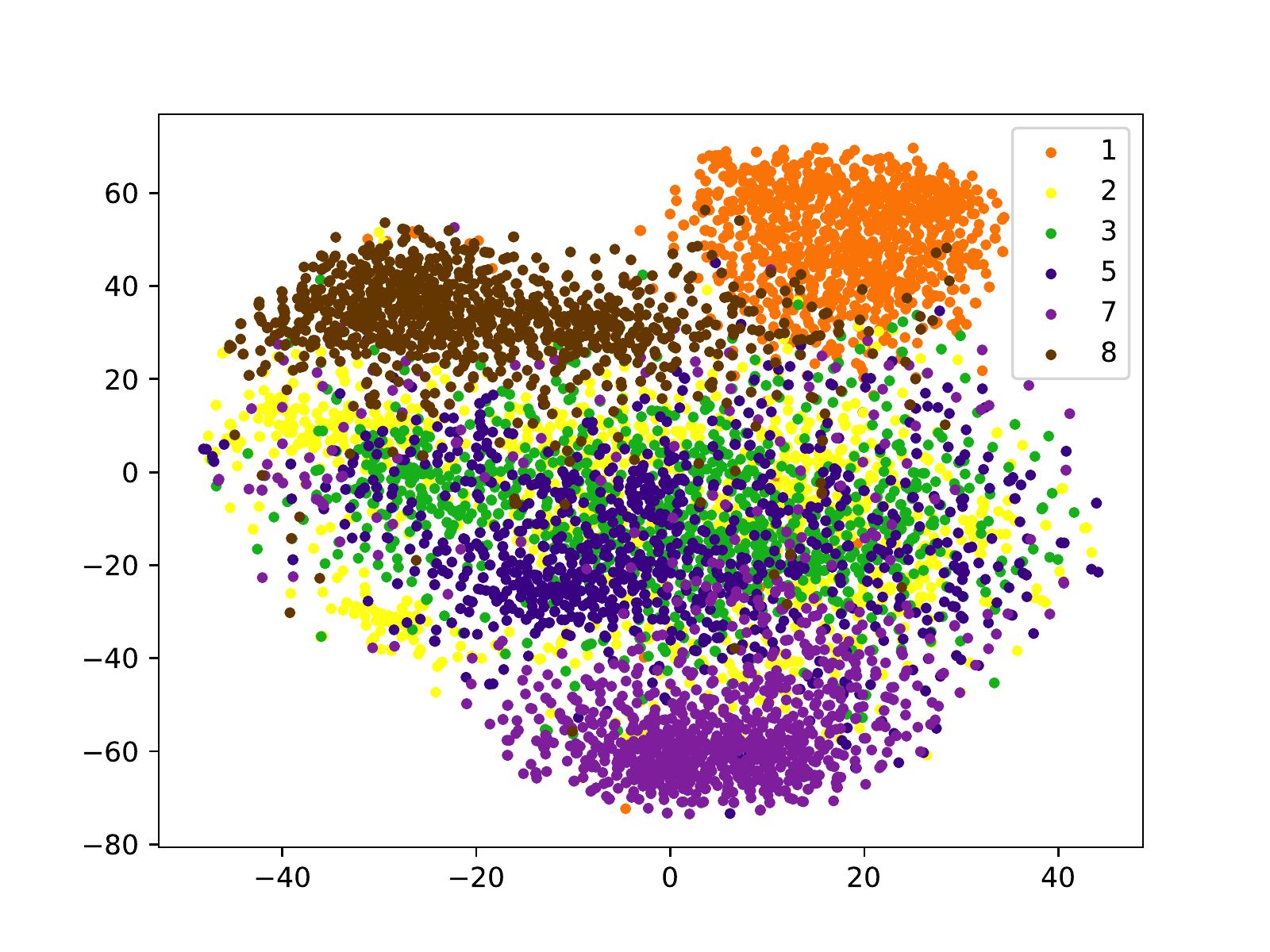}\\
		\mbox{ ({\it d-2}) {CILF-2}}
	\end{minipage}
	\begin{minipage}[h]{28mm}
		\centering
		\includegraphics[width=28mm]{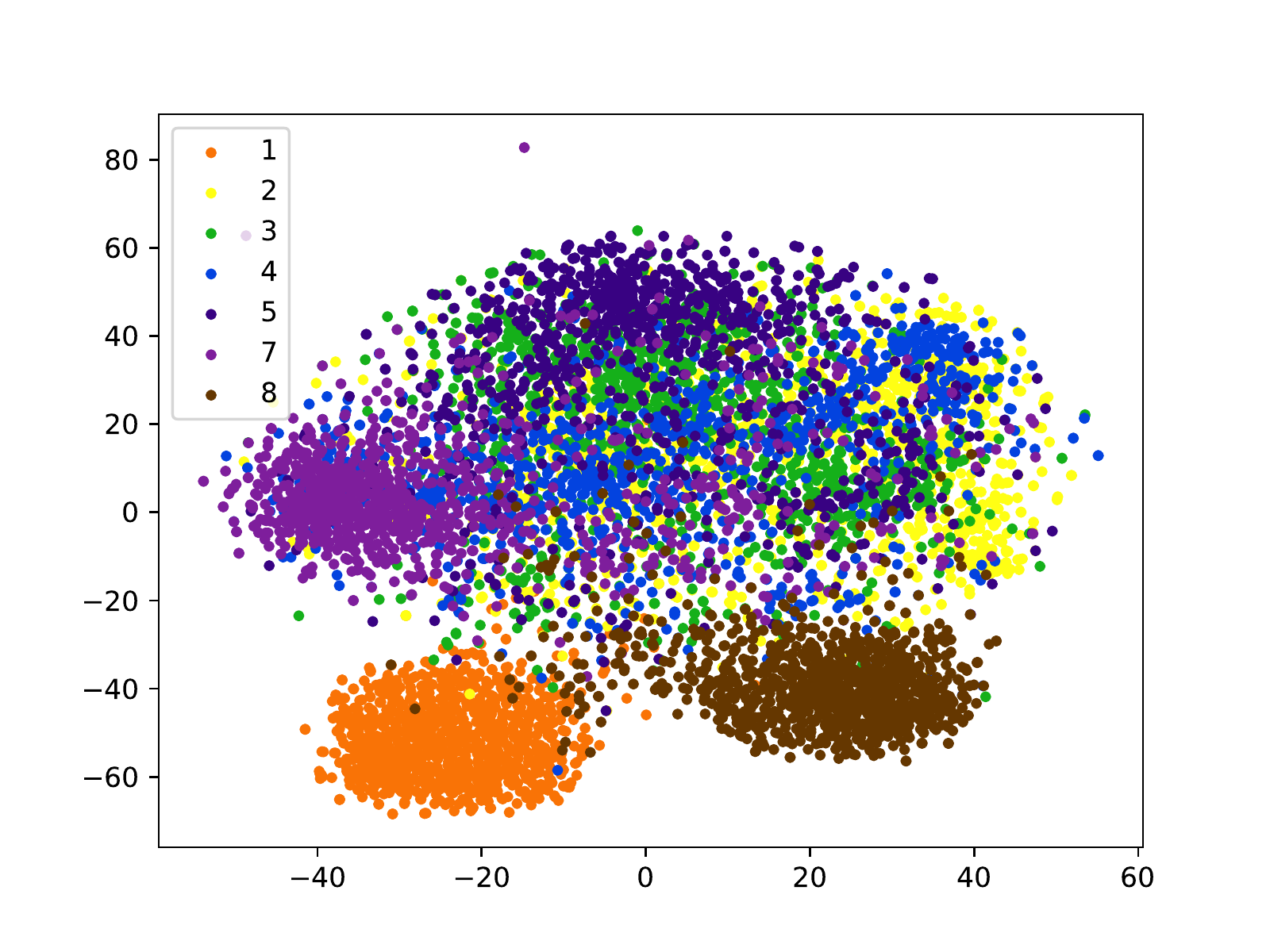}\\
		\mbox{ ({\it d-3}) {CILF-3}}
	\end{minipage}
	\begin{minipage}[h]{28mm}
		\centering
		\includegraphics[width=28mm]{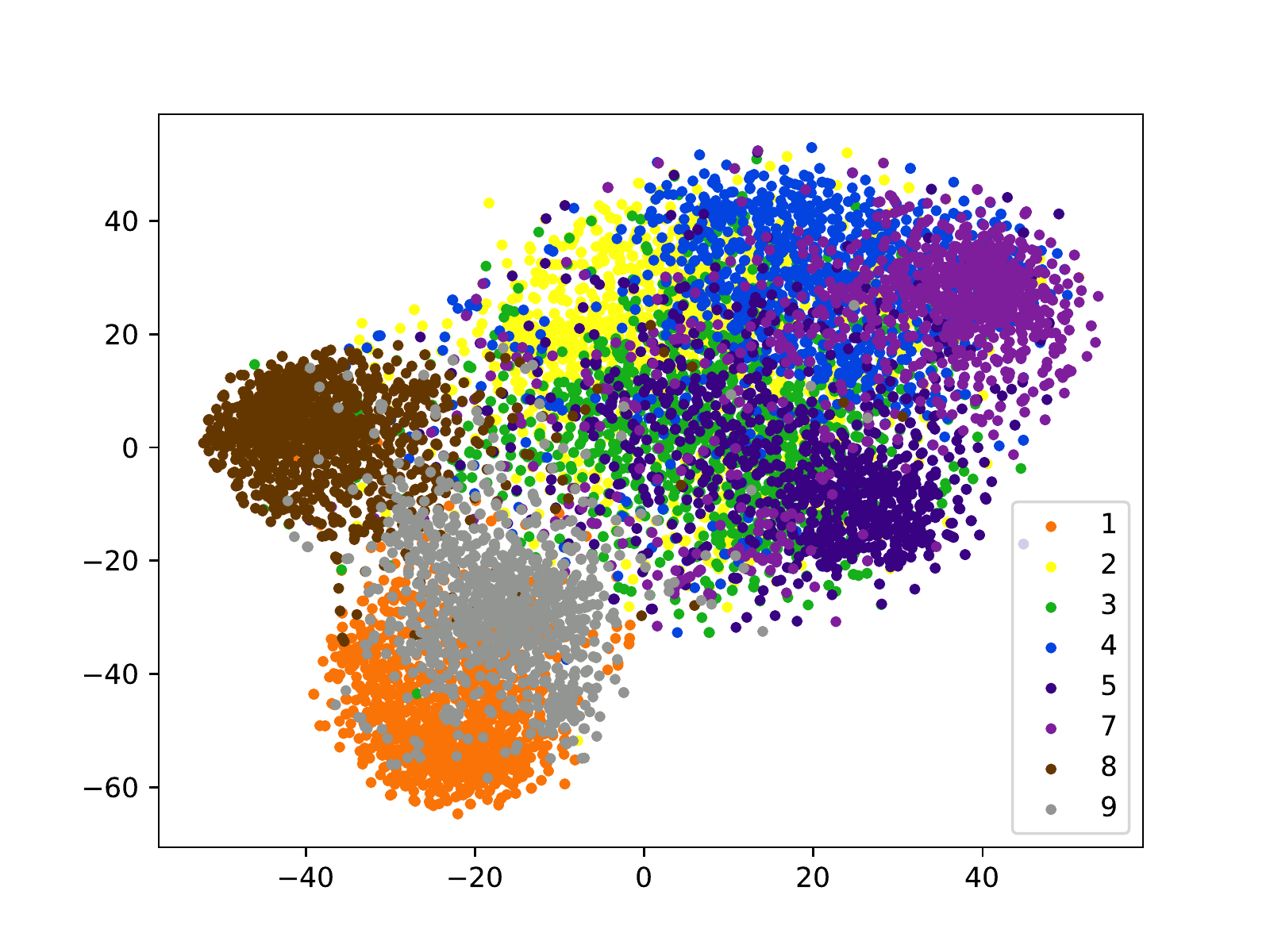}\\
		\mbox{ ({\it d-4}) {CILF-4}}
	\end{minipage}
	\begin{minipage}[h]{28mm}
		\centering
		\includegraphics[width=28mm]{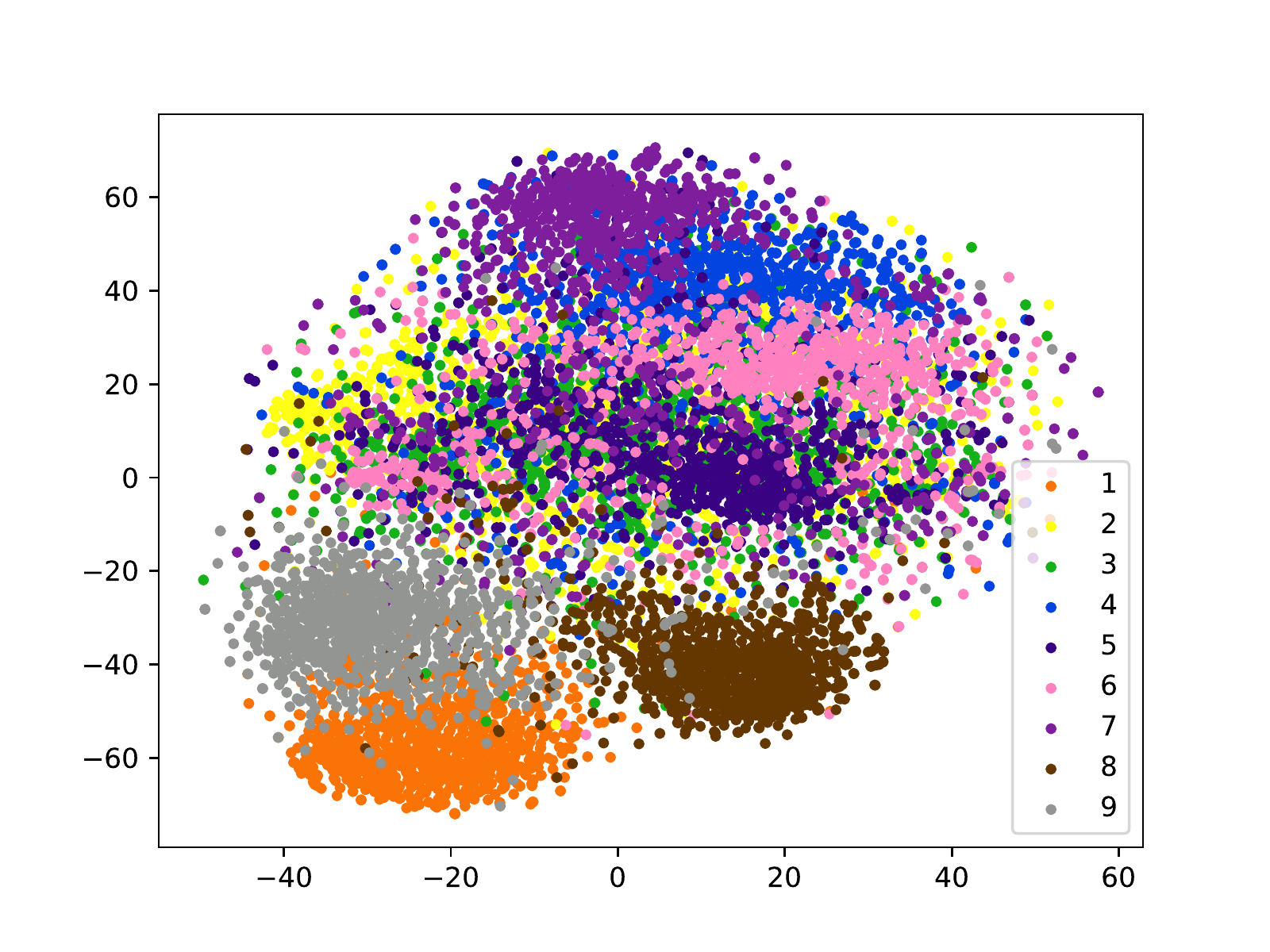}\\
		\mbox{ ({\it d-5}) {CILF-5}}
	\end{minipage}
	\begin{minipage}[h]{28mm}
		\centering
		\includegraphics[width=28mm]{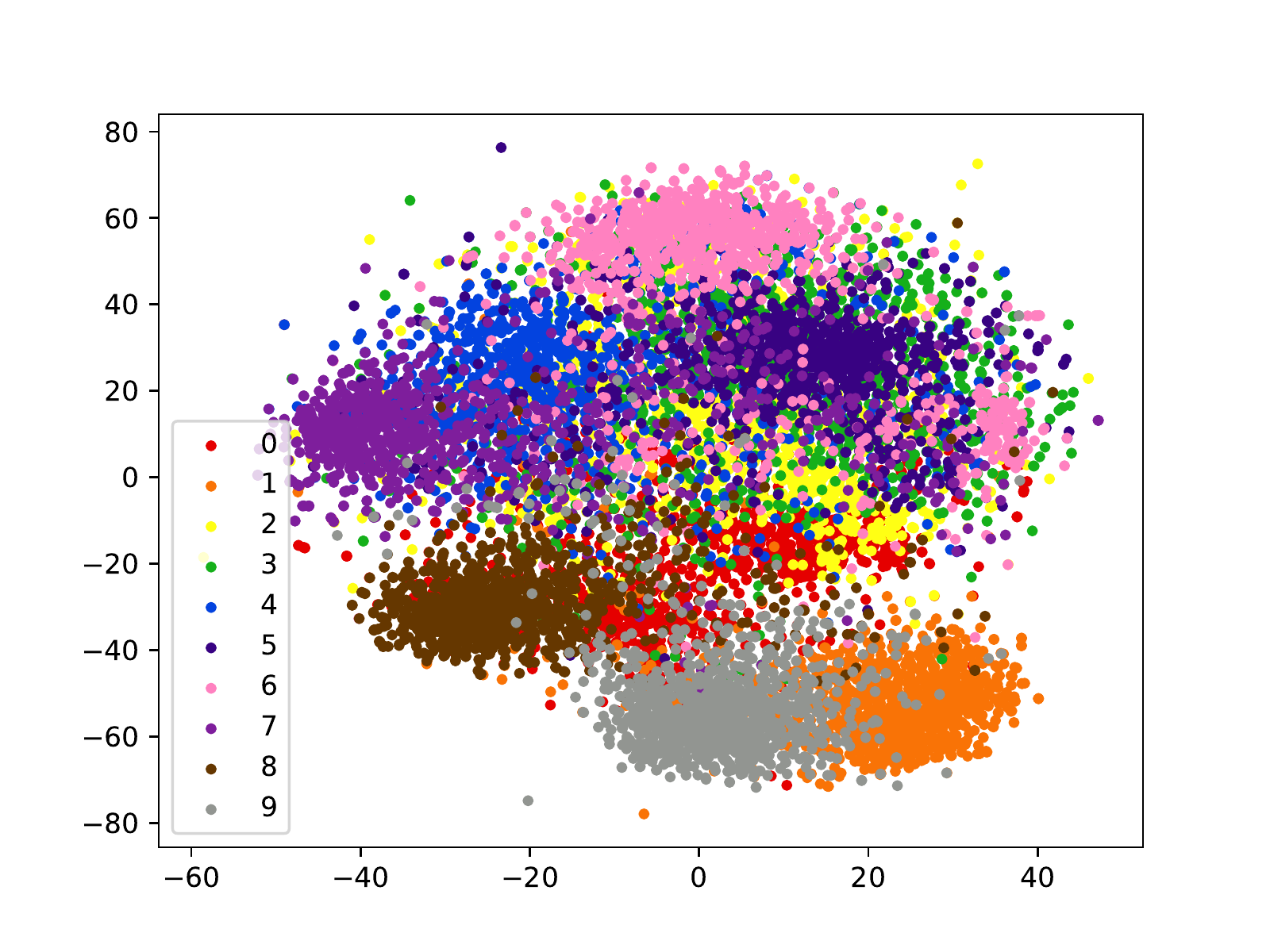}\\
		\mbox{ ({\it d-6}) {CILF-6}}
	\end{minipage}

	\end{center}
	\caption{T-SNE Visualization for both known and unknown classes on CIFAR-10 in single novel class case. (a) original feature space; (b) Learned representations through single detection method CPE~\cite{WangKCTK19}; (c) Learned representations through multi detection method DEC~\cite{HanVZ19}; (d) Learned representations through proposed CILF. Method$-t$ indicates the T-SNE of $t-$th time window of different methods.}\label{fig:f3}
\end{figure*}

\begin{table}[htb]{
		\centering
		\caption{Forgetting measure of known classes over streaming data in single novel class case. The best results are highlighted in bold.}
		\label{tab:tab2}
		\begin{tabular}{l|c|c|c|c}
			\toprule
			\multirow{2}{*}{Methods} & \multicolumn{4}{c}{Forgetting $\downarrow$} \\
			\cmidrule(l){2-5}
			& MNIST & CIFAR-10 & CIFAR-50 & CIFAR-100\\
			\midrule
			Iforest  &    .027$\pm$.006 &    .021$\pm$.002 &    .043$\pm$.001 &    .067$\pm$.001 \\
			One-SVM  &    .028$\pm$.007 &    .019$\pm$.004 &    .042$\pm$.002 &    .068$\pm$.002 \\
			LACU-SVM &    .032$\pm$.005 &    .029$\pm$.002 &    .038$\pm$.003 &    .061$\pm$.001 \\
			SENC-MAS &    .028$\pm$.005 &    .020$\pm$.001 &    .036$\pm$.002 &    .060$\pm$.001 \\
			\midrule
			ODIN-CNN &    .040$\pm$.004 &    .012$\pm$.006 &    .023$\pm$.002 &    .018$\pm$.005 \\
			CFO      &    .023$\pm$.012 &    .010$\pm$.003 &    .014$\pm$.001 &    .016$\pm$.005 \\
			CPE      &    .022$\pm$.001 &    .007$\pm$.001 &    .017$\pm$.001 &    .013$\pm$.003 \\
			\midrule
			DEC      &    .026$\pm$.005 &    .009$\pm$.001 &    .015$\pm$.002 &    .014$\pm$.001 \\
			\midrule
			DNN-Base &    .042$\pm$.005 &    .012$\pm$.007 &    .015$\pm$.003 &    .024$\pm$.006 \\
			DNN-L2   &    .032$\pm$.010 &    .011$\pm$.004 &    .012$\pm$.007 &    .029$\pm$.003 \\
			DNN-EWC  &    .023$\pm$.014 &    .016$\pm$.010 &    .014$\pm$.009 &    .016$\pm$.007 \\
			IMM      &    .030$\pm$.012 &    .009$\pm$.003 &    .016$\pm$.004 &    .025$\pm$.004 \\
			DEN      &    .024$\pm$.003 &    .007$\pm$.004 &    .011$\pm$.009 &    .017$\pm$.001 \\
			\midrule
			CILF     & \bf.020$\pm$.016 & \bf.006$\pm$.001 & \bf.010$\pm$.001 & \bf.013$\pm$.003 \\
			\bottomrule
	\end{tabular}}
\end{table}

\begin{figure}[t]
	\begin{center}
		\begin{minipage}[h]{44mm}
			\centering
			\includegraphics[width=44mm]{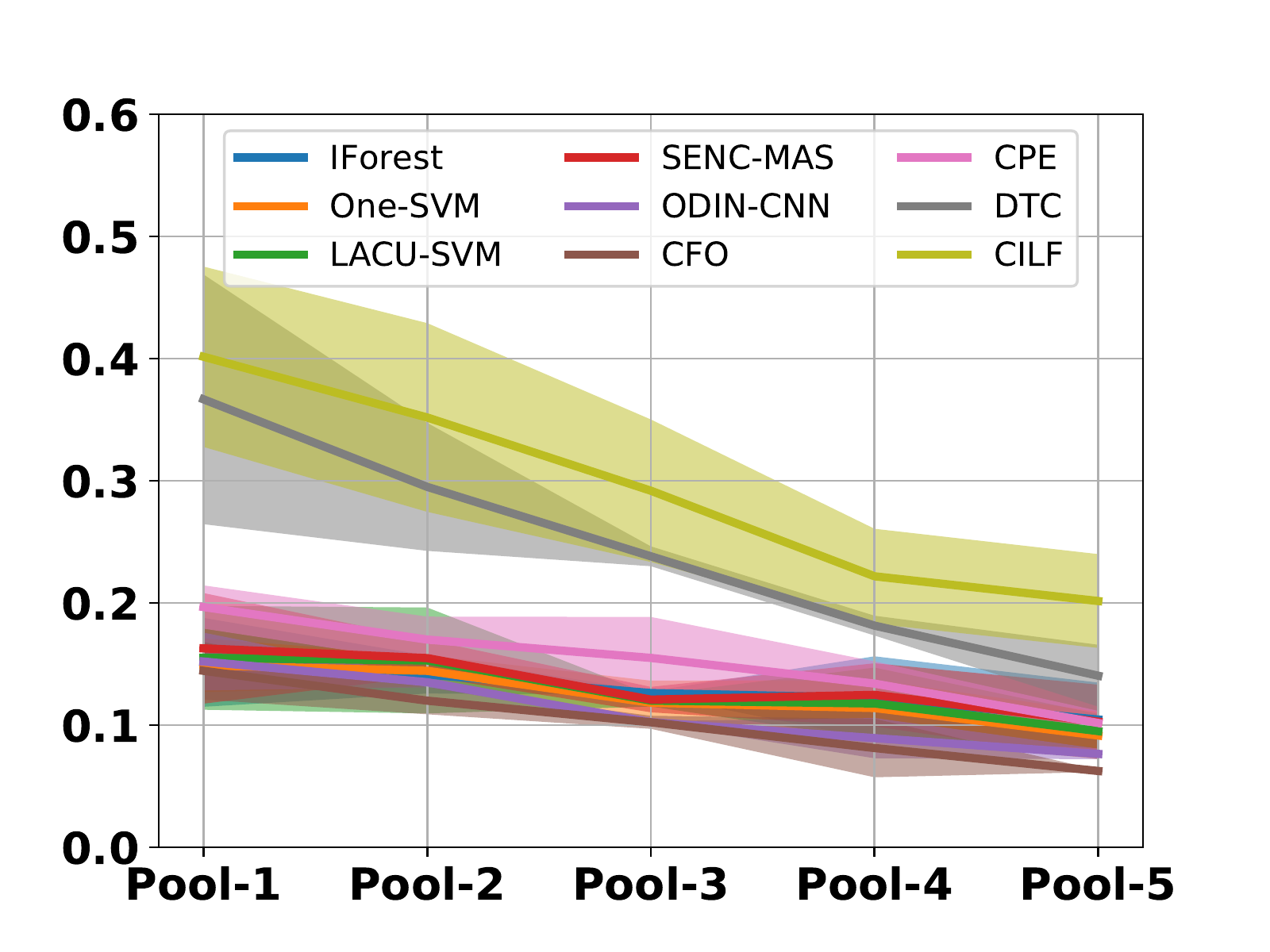}\\
			\mbox{ \;\;\;\; ({\it a}) {NA}}
		\end{minipage}
		\begin{minipage}[h]{44mm}
			\centering
			\includegraphics[width=43mm]{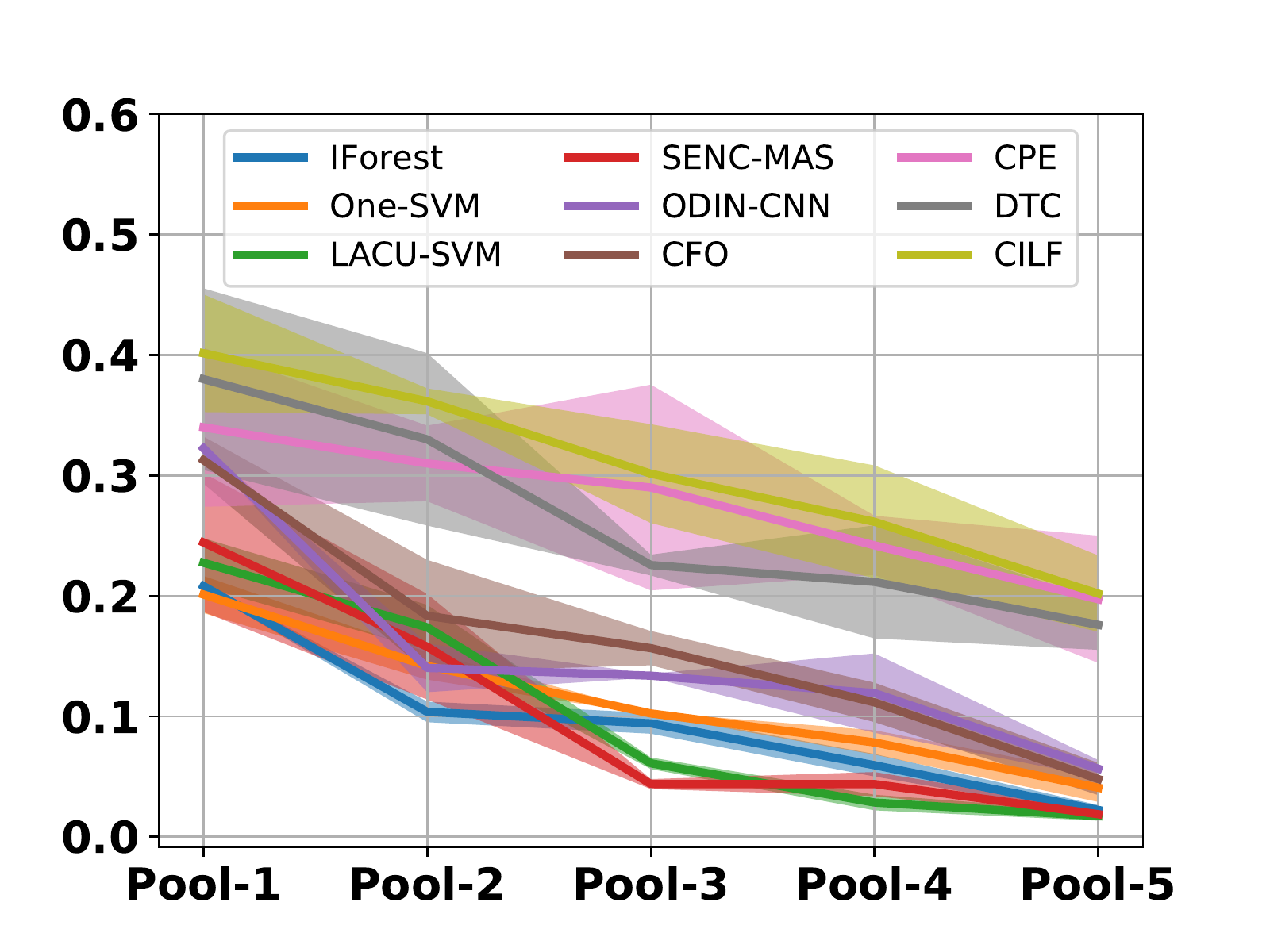}\\
			\mbox{ \;\;\;\; ({\it b}) {Macro-F-Measure}}
		\end{minipage}\\
		\begin{minipage}[h]{44mm}
			\centering
			\includegraphics[width=44mm]{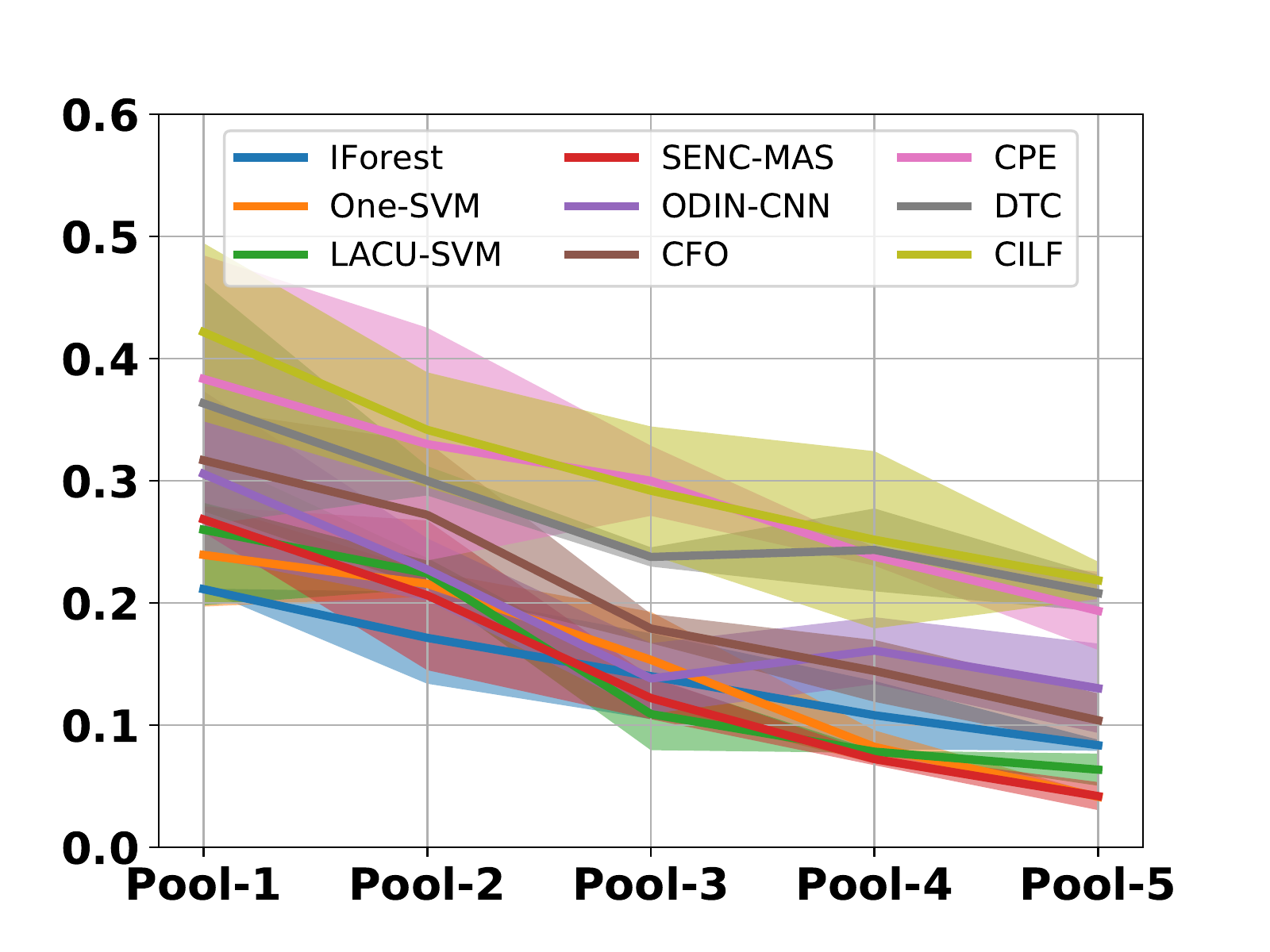}\\
			\mbox{ \;\;\;\; ({\it c}) {Micro-F-Measure}}
		\end{minipage}
		\begin{minipage}[h]{44mm}
			\centering
			\includegraphics[width=43mm]{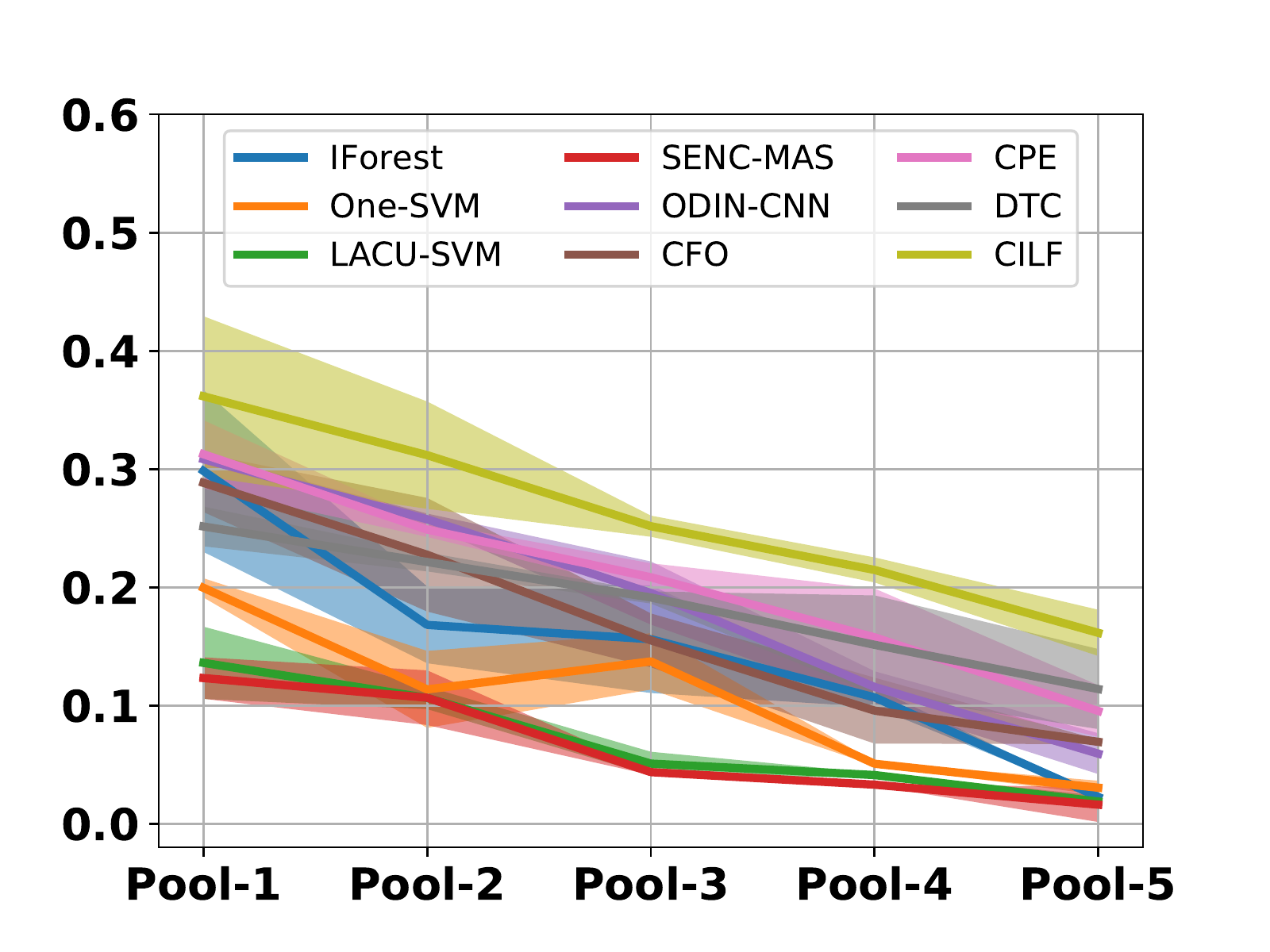}\\
			\mbox{ \;\;\;\; ({\it d}) {AUROC}}
		\end{minipage}
	\end{center}
	\caption{Performance of known classes on different time window of CIFAR-10.}\label{fig:f4}
\end{figure}

\subsection{Datasets}
We utilize three commonly used visual datasets for class-incremental scenario following~\cite{WangKCTK19,geng2018,NealOFWL18}, including  MNIST~\cite{lecun1998mnist}, CIFAR-10~\cite{krizhevsky2009learning}, CIFAR-100~\footnote{http://www.cs.toronto.edu/kriz/cifar.html}. In detail, MNIST dataset contains labeled handwritten digits images from 10 categories, where each class contains between 6313 and 7877 monochrome images; CIFAR-10 dataset has a total of 60000 color images of 32x32 pixels from 10 natural image classes; CIFAR-100 dataset is enlarged CIFAR-10, and we structure CIFAR-100 into 2 datasets: CIFAR-50 and CIFAR-100 according to~\cite{NealOFWL18}. 

Inspired from~\cite{MasudGKHT11,WangKCTK19,geng2018,NealOFWL18,chaudhry2018riemannian}, we utilize the given testing data from the raw datasets as a holdout set to evaluate forgetting, and use the given training data to generate the streaming data. Specifically, we rearrange instances in each dataset to emulate a streaming form with novel classes considering two forms: (1) single novel class each time window; (2) multiple novel classes each time window. For single novel class case, we randomly choose $C$ initial classes, and only 1 novel class may start for each time window. In order to be more in line with real-world applications, each known class may disappear randomly at the end of current time window. Specifically, we set $C=5$ for MNIST and CIFAR-10, $C=30$ for CIFAR-50, $C=50$ for CIFAR-100. Figure \ref{fig:data} (a) presents a simulated example of the CIFAR-10, i.e., we randomly choose 5 initial classes, and there are 5 time windows with 1 novel class starting for each time window. For multiple novel class case, we randomly choose $C$ initial classes, and $K^t$ novel classes (i.e., $K^t \in [2, K]$ novel classes) may randomly start for each time window. Similar to single class setting, each class may disappear randomly. Specifically, we set $C=3$ for MNIST and CIFAR-10, $C=30$ for CIFAR-50, $C=50$ for CIFAR-100. Figure \ref{fig:data} (b) presents a simulated example of the CIFAR-10, i.e., we choose 3 initial classes and there are 3 time windows, 2 novel classes start for the first time window, 3 novel classes for the second time window, 2 novel classes for the last window.

%


\begin{table*}[t]{
		\centering
		\caption{Classification of known classes and novel class detection performance over streaming data in multiple novel class case. The best results are highlighted in bold.}
		\label{tab:tab3}
		\begin{tabular*}{1\textwidth}{@{\extracolsep{\fill}}@{}l|c|c|c|c|c|c|c|c}
			\toprule
			\multirow{2}{*}{Methods} & \multicolumn{4}{c|}{Average NA $\uparrow$} & \multicolumn{4}{c}{Average Macro-F-Measure $\uparrow$}\\
			\cmidrule(l){2-9}
			& MNIST & CIFAR-10 & CIFAR-50 & CIFAR-100 & MNIST & CIFAR-10 & CIFAR-50 & CIFAR-100\\
			\midrule
			Iforest  &    .348$\pm$.069 &    .277$\pm$.062 &    .133$\pm$.021 &    .095$\pm$.014 &    .181$\pm$.062 &    .103$\pm$.004 &    .040$\pm$.004 &    .028$\pm$.002 \\
			One-SVM  &    .361$\pm$.085 &    .277$\pm$.049 &    .136$\pm$.017 &    .089$\pm$.014 &    .188$\pm$.085 &    .107$\pm$.008 &    .038$\pm$.005 &    .026$\pm$.003 \\
			LACU-SVM &    .357$\pm$.054 &    .274$\pm$.062 &    .133$\pm$.012 &    .090$\pm$.017 &    .191$\pm$.054 &    .098$\pm$.008 &    .032$\pm$.005 &    .023$\pm$.002 \\
			SENC-MAS &    .336$\pm$.075 &    .280$\pm$.050 &    .139$\pm$.021 &    .091$\pm$.013 &    .163$\pm$.071 &    .106$\pm$.002 &    .035$\pm$.004 &    .024$\pm$.002 \\
			\midrule
			ODIN-CNN &    .396$\pm$.021 &    .304$\pm$.039 &    .181$\pm$.008 &    .071$\pm$.007 &    .268$\pm$.032 &    .243$\pm$.005 &    .160$\pm$.006 &    .054$\pm$.004 \\
			CFO      &    .353$\pm$.012 &    .287$\pm$.027 &    .212$\pm$.009 &    .104$\pm$.008 &    .380$\pm$.021 &    .363$\pm$.001 &    .288$\pm$.007 &    .183$\pm$.005 \\
			CPE      &    .413$\pm$.030 &    .401$\pm$.061 &    .240$\pm$.020 &    .132$\pm$.020 &    .236$\pm$.047 &    .339$\pm$.037 &    .249$\pm$.022 &    .244$\pm$.021 \\
			\midrule
			DEC      &    .350$\pm$.142 &    .398$\pm$.084 &    .206$\pm$.012 &    .089$\pm$.016 &    .327$\pm$.315 &    .302$\pm$.217 &    .230$\pm$.017 &    .110$\pm$.011 \\
			\midrule
			CILF     & \bf.493$\pm$.051 & \bf.422$\pm$.054 & \bf.258$\pm$.040 & \bf.179$\pm$.031 & \bf.392$\pm$.045 & \bf.407$\pm$.033 & \bf.332$\pm$.035 & \bf.259$\pm$.041 \\
		\end{tabular*}
		\begin{tabular*}{1\textwidth}{@{\extracolsep{\fill}}@{}l|c|c|c|c|c|c|c|c}
			\toprule
			\multirow{2}{*}{Methods} & \multicolumn{4}{c|}{Average Micro-F-Measure $\uparrow$} & \multicolumn{4}{c}{Average AUROC $ \uparrow$}\\
			\cmidrule(l){2-9}
			& MNIST & CIFAR-10 & CIFAR-50 & CIFAR-100 & MNIST & CIFAR-10 & CIFAR-50 & CIFAR-100\\
			\midrule
			Iforest  &    .272$\pm$.079 &    .153$\pm$.009 &    .067$\pm$.002 &    .049$\pm$.003 &    .140$\pm$.015 &    .126$\pm$.019 &    .113$\pm$.023 &    .140$\pm$.009 \\
			One-SVM  &    .293$\pm$.122 &    .161$\pm$.018 &    .068$\pm$.008 &    .045$\pm$.003 &    .158$\pm$.016 &    .153$\pm$.011 &    .146$\pm$.008 &    .092$\pm$.010 \\
			LACU-SVM &    .294$\pm$.060 &    .144$\pm$.006 &    .062$\pm$.006 &    .043$\pm$.002 &    .149$\pm$.002 &    .158$\pm$.016 &    .145$\pm$.005 &    .152$\pm$.024 \\
			SENC-MAS &    .250$\pm$.082 &    .165$\pm$.012 &    .063$\pm$.008 &    .045$\pm$.004 &    .107$\pm$.092 &    .123$\pm$.006 &    .086$\pm$.051 &    .043$\pm$.032 \\
			\midrule
			ODIN-CNN &    .242$\pm$.024 &    .193$\pm$.006 &    .072$\pm$.009 &    .067$\pm$.006 &    .231$\pm$.052 &    .145$\pm$.245 &    .193$\pm$.140 &    .133$\pm$.250 \\
			CFO      &    .245$\pm$.016 &    .212$\pm$.004 &    .108$\pm$.010 &    .104$\pm$.008 &    .195$\pm$.050 &    .214$\pm$.257 &    .115$\pm$.145 &    .159$\pm$.245 \\
			CPE      &    .207$\pm$.037 &    .367$\pm$.035 &    .200$\pm$.023 &    .146$\pm$.022 &    .241$\pm$.056 &    .255$\pm$.243 & \bf.201$\pm$.140 &    .183$\pm$.254 \\
			\midrule
			DEC      &    .294$\pm$.292 &    .231$\pm$.204 &    .141$\pm$.016 &    .116$\pm$.011 &    .234$\pm$.061 &    .217$\pm$.076 &    .197$\pm$.137 &    .171$\pm$.253 \\
			\midrule
			CILF     & \bf.422$\pm$.043 & \bf.442$\pm$.043 & \bf.212$\pm$.022 & \bf.152$\pm$.016 & \bf.286$\pm$.028 & \bf.261$\pm$.022 &    .189$\pm$.034 & \bf.192$\pm$.016 \\
			\bottomrule
	\end{tabular*}}
\end{table*}

\begin{table}[htb]{
		\centering
		\caption{Forgetting measure of known classes over streaming data in multiple novel class case. The best results are highlighted in bold.}
		\label{tab:tab4}
		\begin{tabular}{l|c|c|c|c}
			\toprule
			\multirow{2}{*}{Methods} & \multicolumn{4}{c}{Forgetting $\downarrow$} \\
			\cmidrule(l){2-5}
			& MNIST & CIFAR-10 & CIFAR-50 & CIFAR-100\\
			\midrule
			Iforest  &    .014$\pm$.015 &    .006$\pm$.004 &    .035$\pm$.002 &    .048$\pm$.005 \\
			One-SVM  &    .018$\pm$.010 &    .005$\pm$.003 &    .033$\pm$.004 &    .049$\pm$.003 \\
			LACU-SVM &    .017$\pm$.010 &    .004$\pm$.002 &    .029$\pm$.003 &    .042$\pm$.004 \\
			SENC-MAS &    .011$\pm$.011 &    .006$\pm$.002 &    .032$\pm$.002 &    .041$\pm$.004 \\
			\midrule
			ODIN-CNN &    .011$\pm$.008 &    .013$\pm$.011 &    .015$\pm$.006 &    .013$\pm$.010 \\
			CFO      &    .019$\pm$.005 &    .008$\pm$.006 &    .009$\pm$.008 &    .023$\pm$.003 \\
			CPE      &    .007$\pm$.001 &    .011$\pm$.003 &    .009$\pm$.002 &    .005$\pm$.002 \\
			\midrule
			DEC      &    .002$\pm$.003 &    .009$\pm$.001 &    .006$\pm$.002 &    .023$\pm$.001 \\
			\midrule
			DNN-Base &    .022$\pm$.002 &    .023$\pm$.009 &    .032$\pm$.007 &    .039$\pm$.006 \\
			DNN-L2   &    .021$\pm$.002 &    .026$\pm$.001 &    .035$\pm$.005 &    .041$\pm$.002 \\
			DNN-EWC  &    .016$\pm$.010 &    .017$\pm$.010 &    .020$\pm$.005 &    .021$\pm$.009 \\
			IMM      &    .022$\pm$.030 &    .023$\pm$.010 &    .024$\pm$.012 &    .030$\pm$.019 \\
			DEN      &    .010$\pm$.009 &    .013$\pm$.005 &    .013$\pm$.002 &    .021$\pm$.011 \\
			\midrule
			CILF     & \bf.001$\pm$.001 & \bf.005$\pm$.001 & \bf.001$\pm$.001 & \bf.008$\pm$.001 \\
			\bottomrule
	\end{tabular}}
\end{table}

\subsection{Compared Methods}
To validate the effectiveness of proposed CILF, we compared with existing state-of-the-art novel class detection approaches and incremental learning methods. 

First, we compared CILF with existing NCD and incremental NCD methods. Including traditional anomaly detection and linear methods: Iforest~\cite{LiuTZ08}, One-Class SVM (One-SVM)~\cite{ScholkopfPSSW01}, LACU-SVM (LACU)~\cite{DaYZ14}, SENC-MAS (SENC)~\cite{MuZDLZ17}; deep methods: ODIN-CNN (ODIN)~\cite{LiangLS18}, CFO~\cite{NealOFWL18}, CPE~\cite{WangKCTK19} and DTC~\cite{HanVZ19}. Abbreviations in parentheses. DTC is clustering based methods for multiple unknown classes detection. Note that Iforest, One-SVM, LACU, ODIN, CFO, and DTC are NCD methods, SENC and CPE are incremental NCD methods. All NCD baselines except Iforest can be updated incrementally using newly labeled unknown class data and memory data.
\begin{itemize}
	\item Iforest: an ensemble tree method to detect outliers;
	\item One-Class SVM (One-SVM): a baseline for out-of-class detection and classification;
	\item LACU-SVM (LACU): a SVM-based method that incorporates the unlabeled data from open set for unknown class detection;
	\item SENC-MAS (SENC): a matrix sketching method that approximates original information with a dynamic low-dimensional structure;
	\item ODIN-CNN (ODIN): a CNN-based method that distinguishes in-distribution and out-of-distribution over softmax score;
	\item CFO: a generative method that adopts an encoder-decoder GAN to generate synthetic unknown instances;
	\item CPE: a CNN-based ensemble method, which adaptively updates the prototype for detection;
	\item DTC: an extended deep transfer clustering method for novel class detection.
\end{itemize}
Specifically, 1) Iforest, ODIN and CFO can only perform binary classifications, i.e., whether the instance is an unknown class. Thus we further conduct unsupervised clustering on both know and unknown class data for subdividing; 2) all baselines are one-class methods except DTC, i.e., they perform NCD in two steps: first detect the super-class of unknown classes, then perform unsupervised clustering; 3) all of baselines are NCD methods except LACU, SENC and CPE, but they can be applied in incremental NCD by combing memory data to update following~\cite{WangKCTK19}.

To validate the incremental model update, we also compare with state-of-the-art forgetting methods: DNN-Base, DNN-L2, DNN-EWC~\citep{KirkpatrickPRVD16}, IMM~\citep{LeeKJHZ17}, DEN~\citep{Jeongtae2017}, each time window is regarded as a task in these methods. In detail, the compared methods are: 
\begin{itemize}
	\item {\bf DNN-Base:} Base DNN with $L_2$-regularizations;
	\item {\bf DNN-L2:} Base DNN, where at each stage t, $\Theta_t$ is initialized as $\Theta_{t-1}$ and continuously trained with $L_2$-regularization between $\Theta_t$ and $\Theta_{t-1}$;
	\item {\bf DNN-EWC:} Deep network trained with elastic weight consolidation for regularization, which remembers old stages by selectively slowing down learning on the weights important for those stages;
	\item {\bf IMM:} An incremental moment matching method with two extensions: Mean-IMM and Mode-IMM, which incrementally matches the posterior distribution of the neural network trained on the previous stages;
	\item {\bf DEN:} A deep network architecture for incremental learning, which can dynamically decide its network structure with a sequence of stages and learn the overlapping knowledge sharing structure among stages.
\end{itemize}

\subsection{Evaluation Metrics}
Considering that CILF can distinguish the known and unknown classes, while mitigating the forgetting. Thus we measure the proposed method from two aspects: (1) NCD performance; (2) Forgetting performance.

Following~\cite{geng2018}, we adopt the commonly used evaluation metrics for novel class detection: 1) Normalized Accuracy (NA), which weights the accuracy for known and novel classes~\cite{Mendes-JuniorSW17}; 2) Macro-F-measure and Micro-F-measure; and 3) AUROC, which considers the NCD task as a combination of novelty detection and multi-class recognition~\cite{NealOFWL18}.  Moreover, to validate the catastrophic forgetting, we calculate the performance about forgetting profile of different learning algorithms as~\citep{chaudhry2018riemannian}, i.e., let $acc_{m,n}$ be the accuracy evaluated on the hold-out sets, i.e. the novel classes emerge on $n-$th time window ($n \leq m$), after training the network incrementally from stage 1 to $m$, the average accuracy at time $m$ is defined as: $A_m = \frac{1}{m}\sum_{n=1}^m acc_{m,n}$ ~\citep{chaudhry2018riemannian}. higher $A_m$ represents for better classifier. $Forgetting = \frac{A^{*} - mean(A)}{A^{*}}$, $A^{*}$ is the optimal accuracy with the entire data. We repeat all experiments with 5 times, and record the mean and std.

\begin{figure*}[t]
	\begin{center}
		\begin{minipage}[h]{32mm}
			\centering
			\includegraphics[width=32mm]{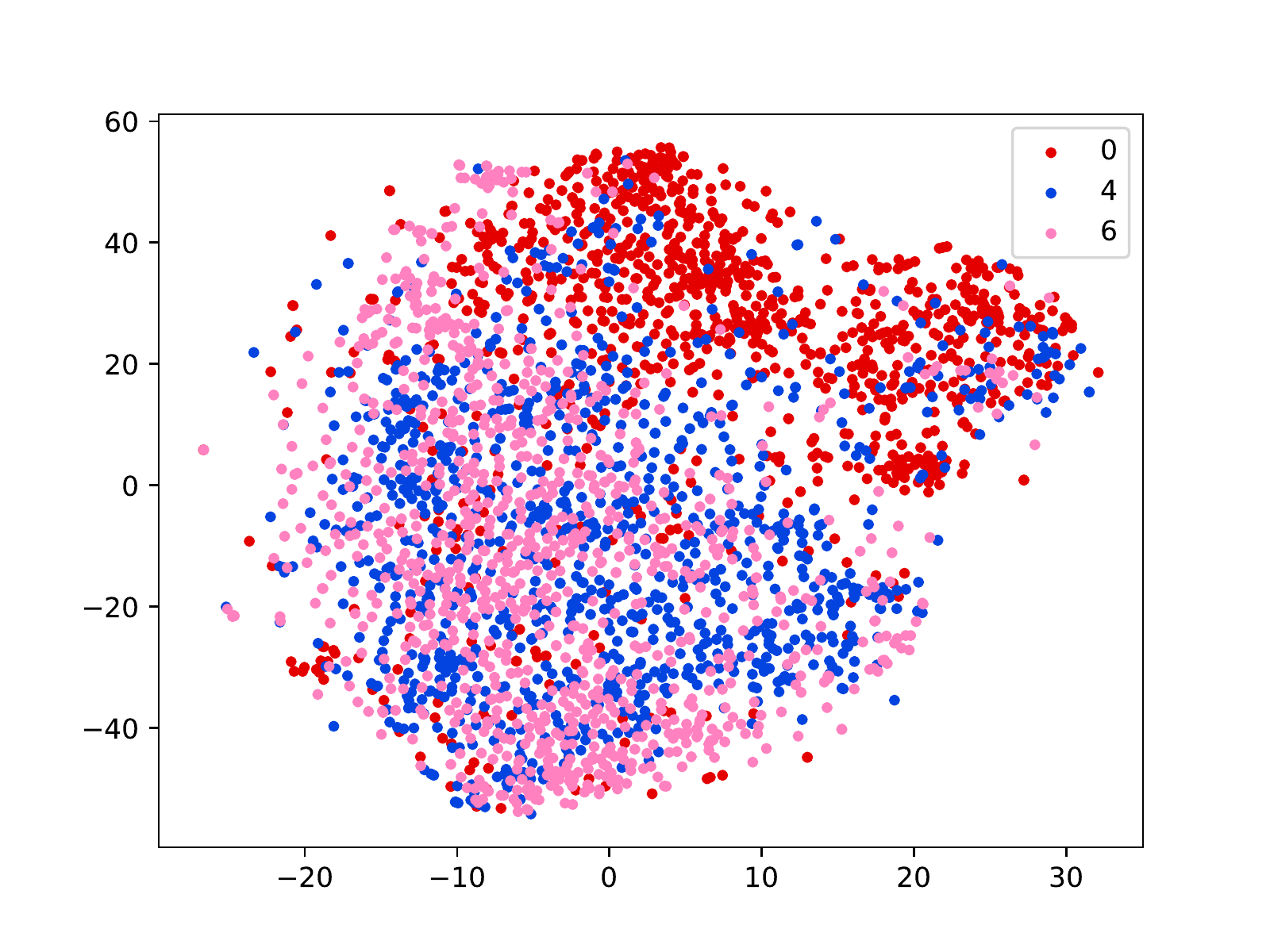}\\
			\mbox{ ({\it a-1}) {Original-1}}
		\end{minipage}
		\begin{minipage}[h]{32mm}
			\centering
			\includegraphics[width=32mm]{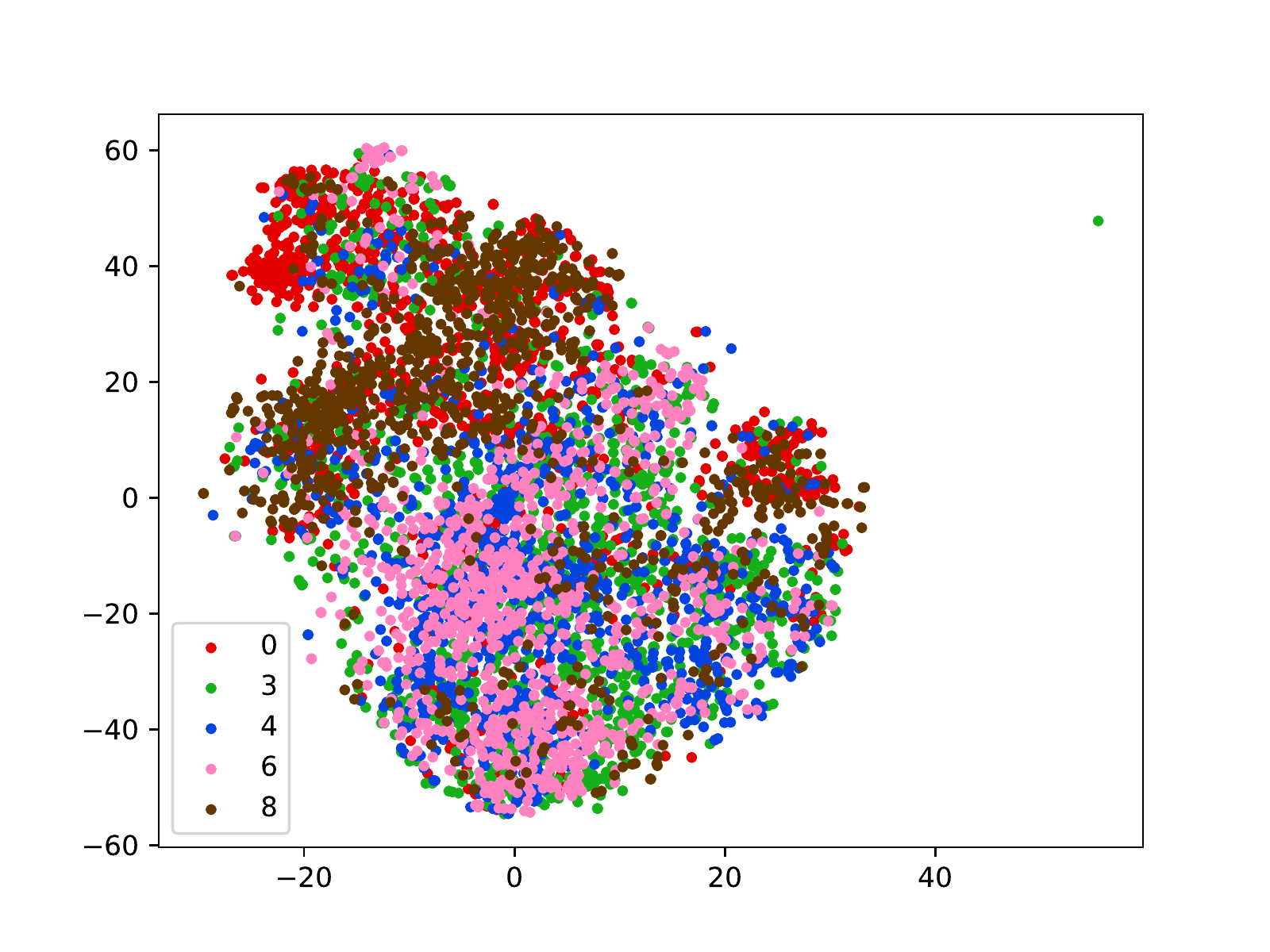}\\
			\mbox{ ({\it a-2}) {Original-2}}
		\end{minipage}
		\begin{minipage}[h]{32mm}
			\centering
			\includegraphics[width=32mm]{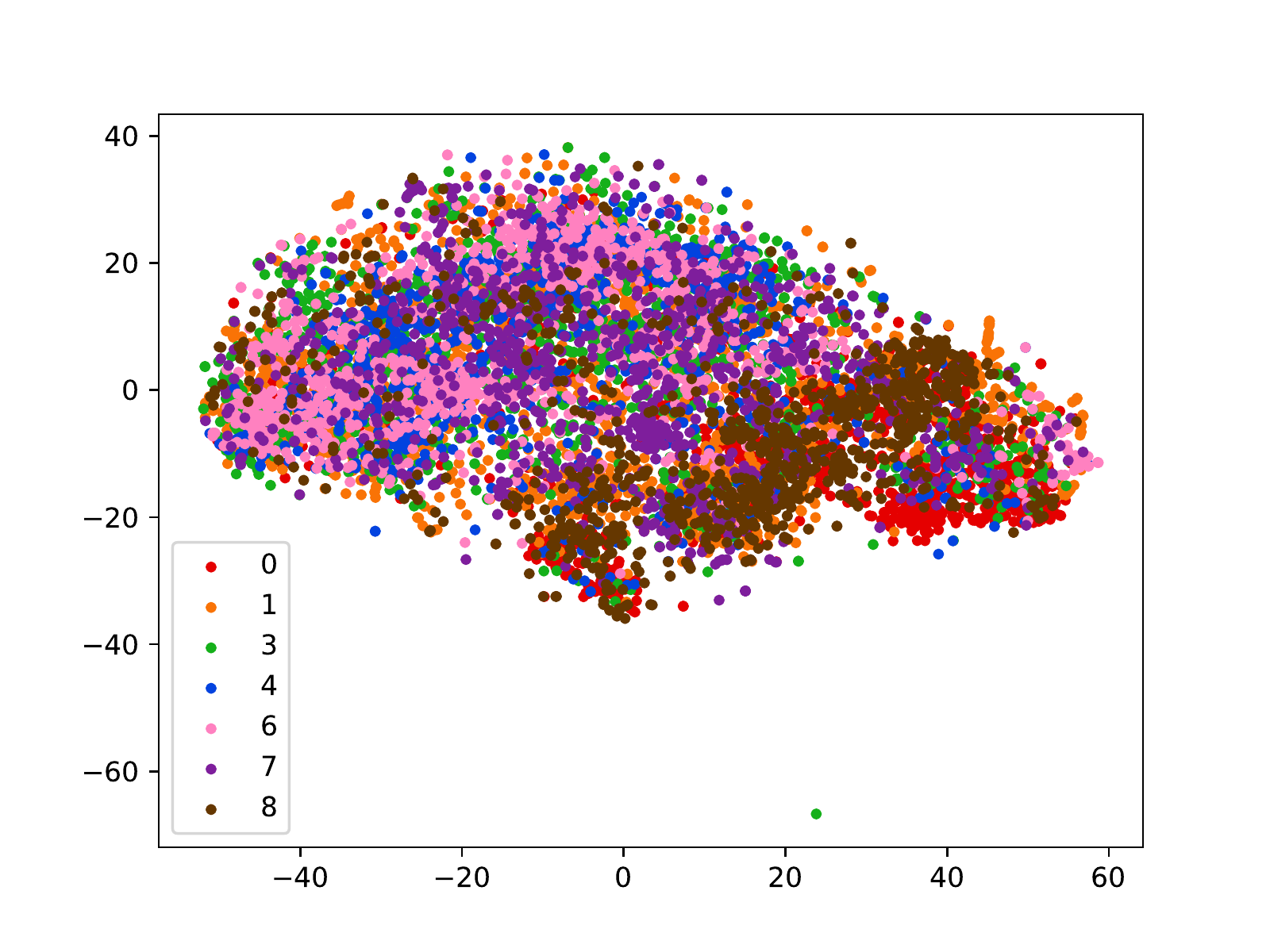}\\
			\mbox{ ({\it a-3}) {Original P3}}
		\end{minipage}
		\begin{minipage}[h]{32mm}
			\centering
			\includegraphics[width=32mm]{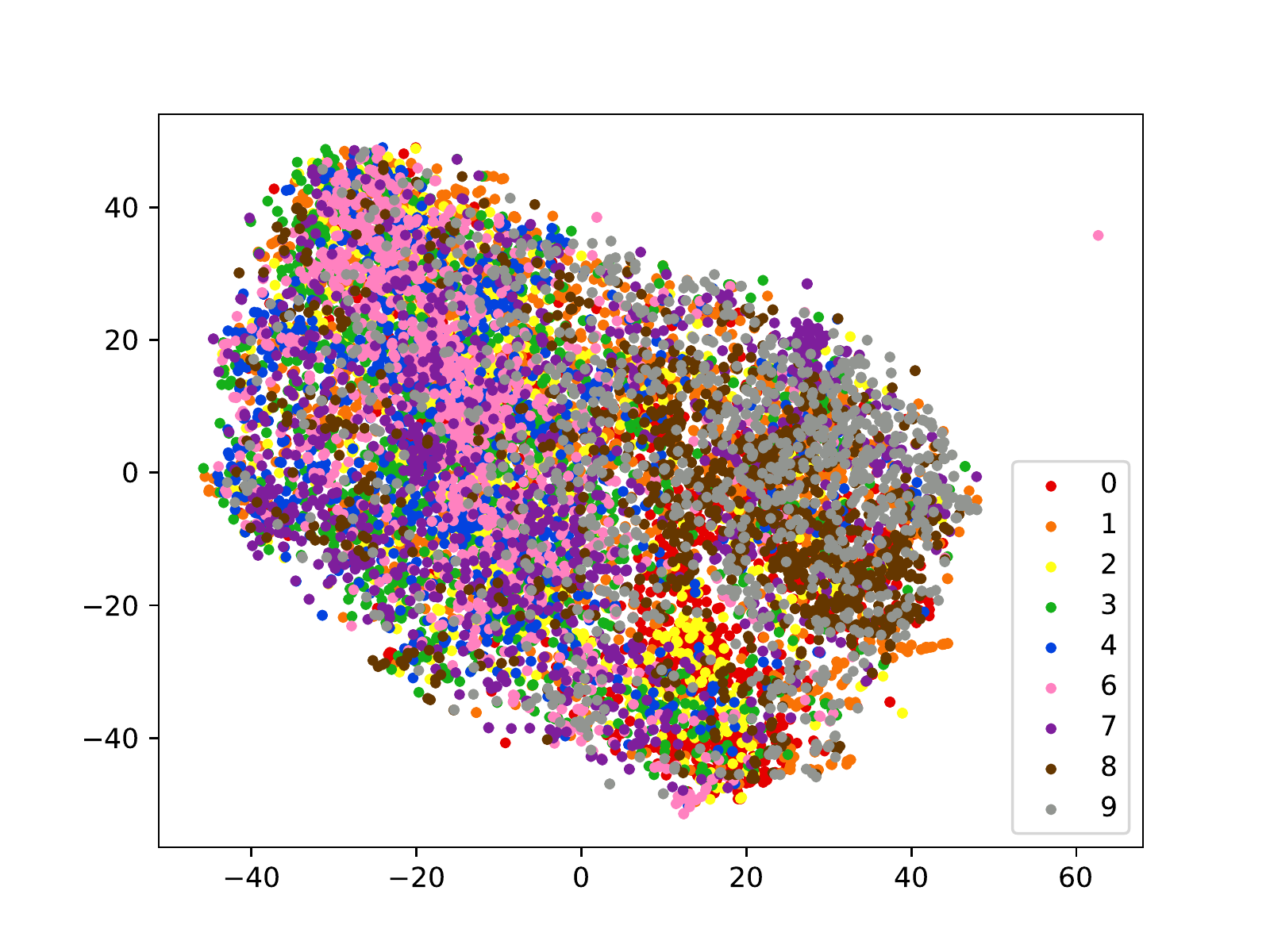}\\
			\mbox{ ({\it a-4}) {Original-4}}
		\end{minipage}
		\begin{minipage}[h]{32mm}
			\centering
			\includegraphics[width=32mm]{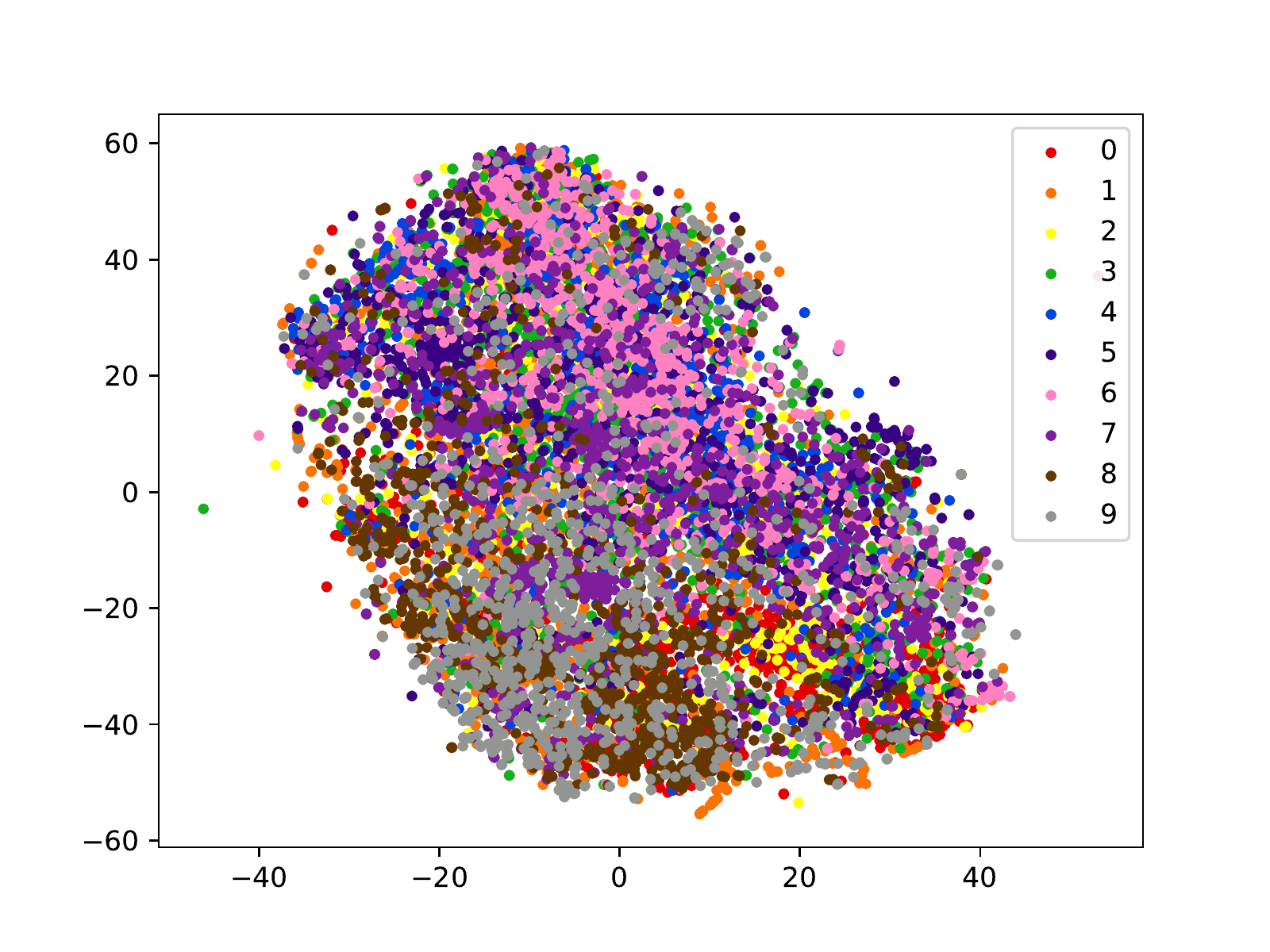}\\
			\mbox{ ({\it a-5}) {Original-5}}
		\end{minipage}		
		\begin{minipage}[h]{32mm}
			\centering
			\includegraphics[width=32mm]{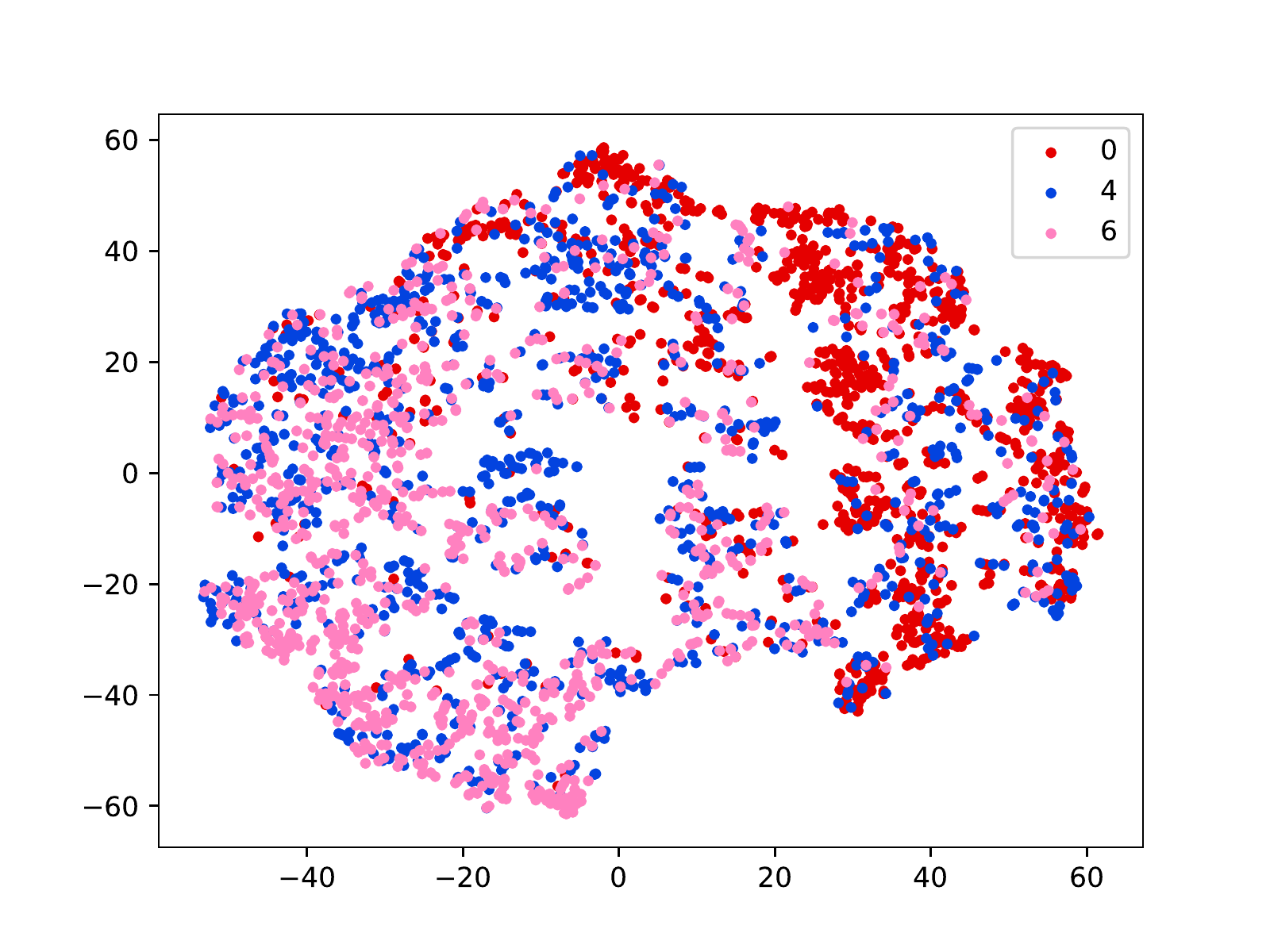}\\
			\mbox{ ({\it b-1}) {CPE-1}}
		\end{minipage}
		\begin{minipage}[h]{32mm}
			\centering
			\includegraphics[width=32mm]{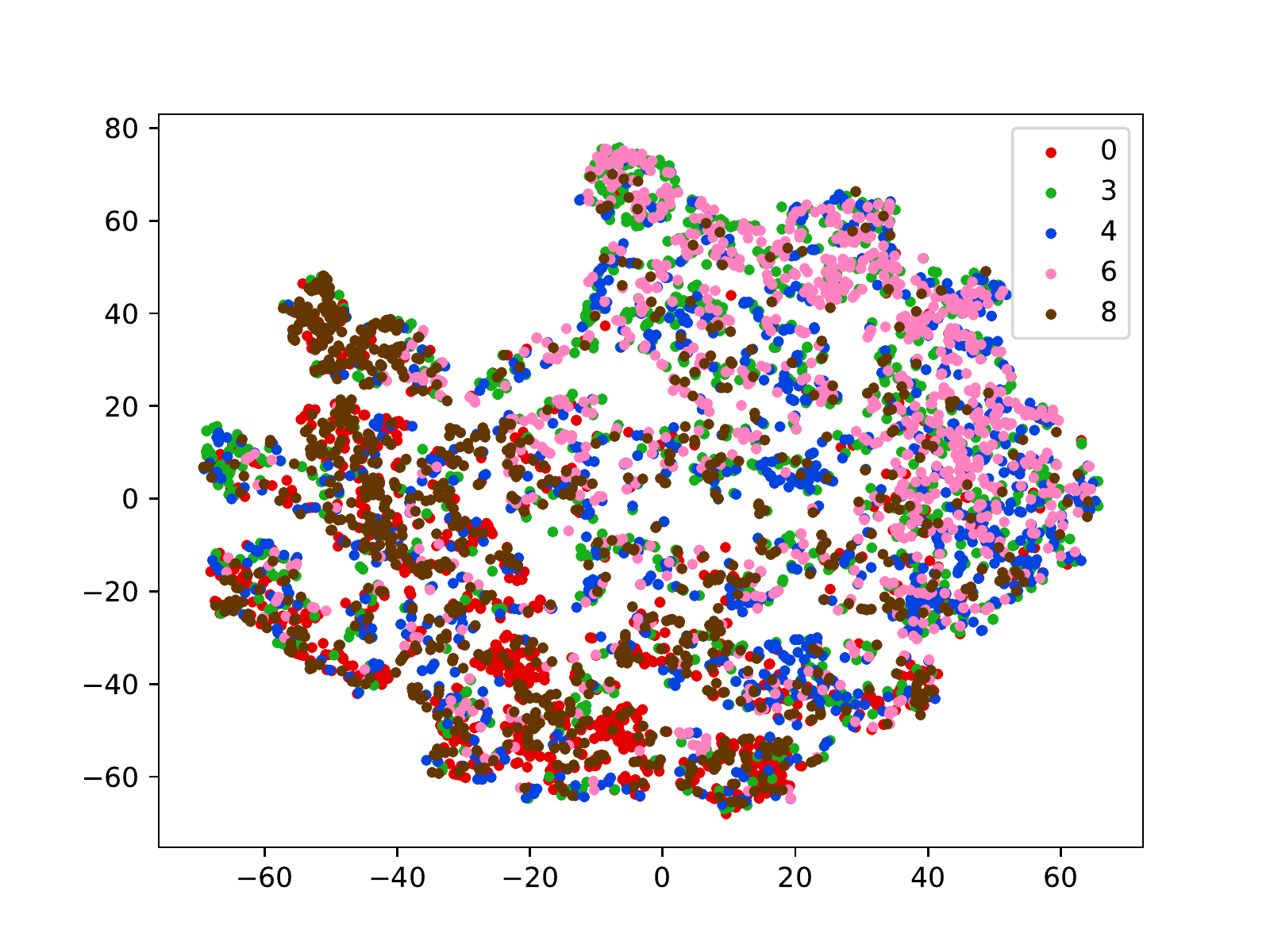}\\
			\mbox{ ({\it b-2}) {CPE-2}}
		\end{minipage}
		\begin{minipage}[h]{32mm}
			\centering
			\includegraphics[width=32mm]{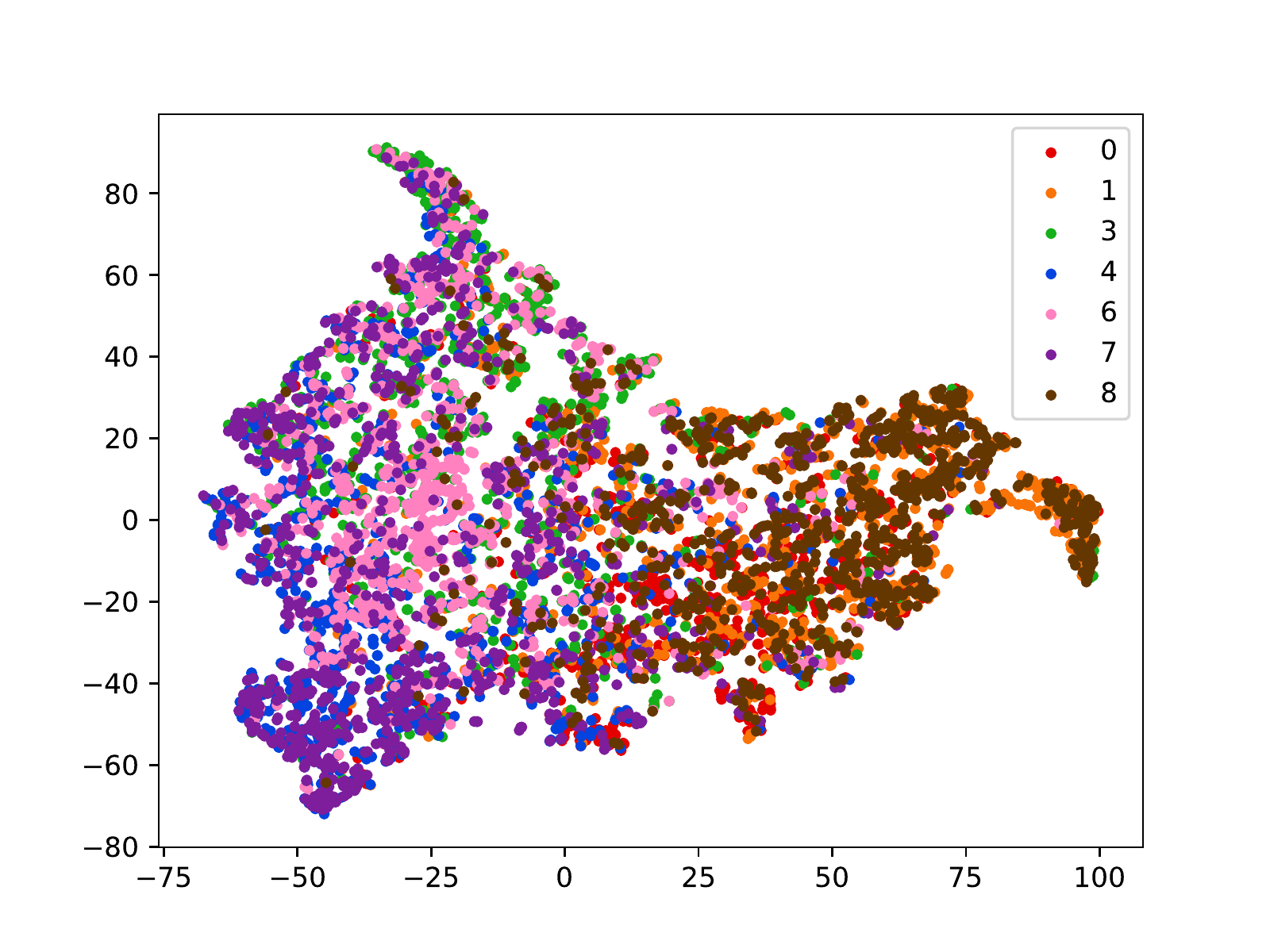}\\
			\mbox{ ({\it b-3}) {CPE-3}}
		\end{minipage}
		\begin{minipage}[h]{32mm}
			\centering
			\includegraphics[width=32mm]{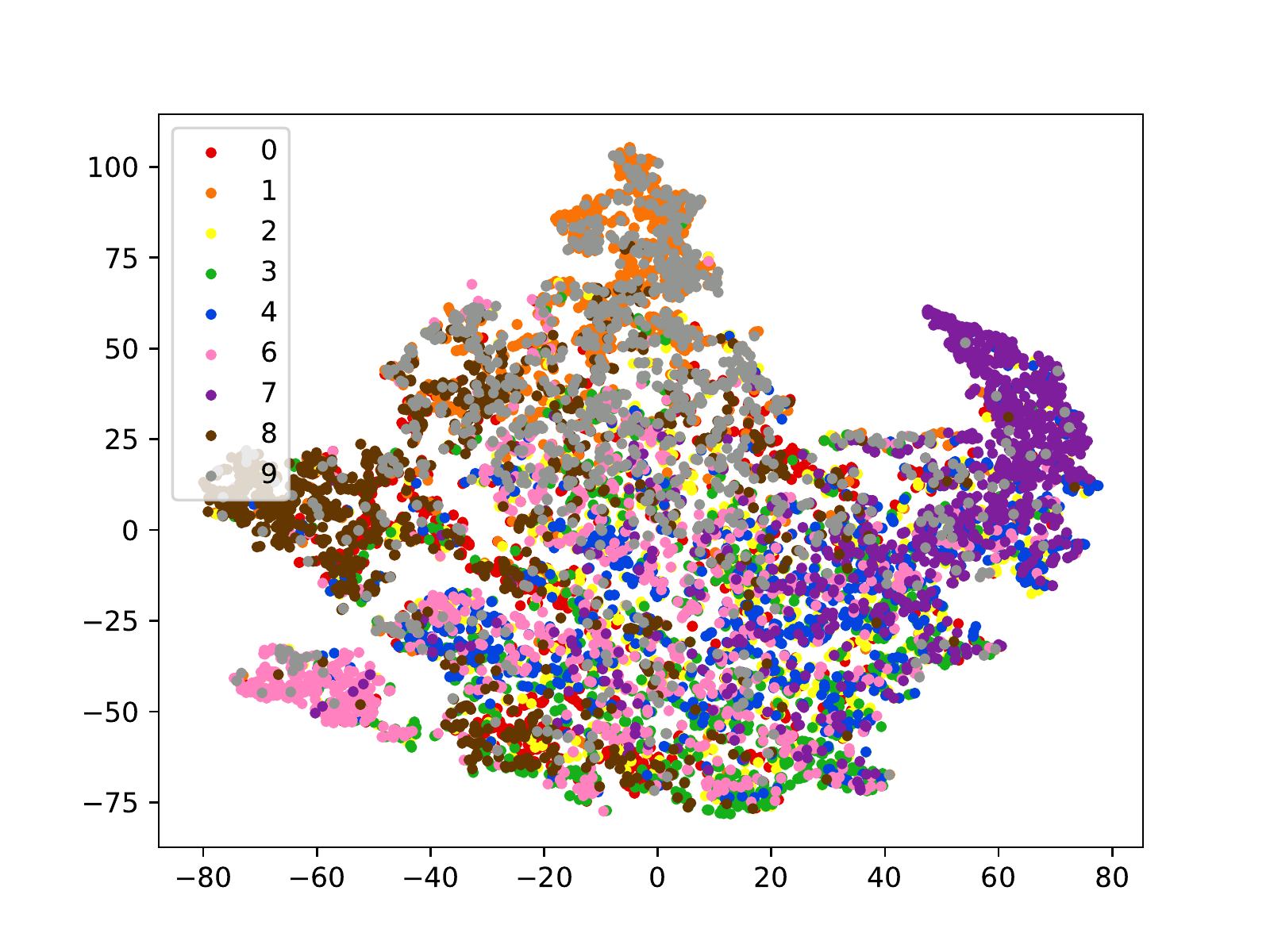}\\
			\mbox{ ({\it b-4}) {CPE-4}}
		\end{minipage}
		\begin{minipage}[h]{32mm}
			\centering
			\includegraphics[width=32mm]{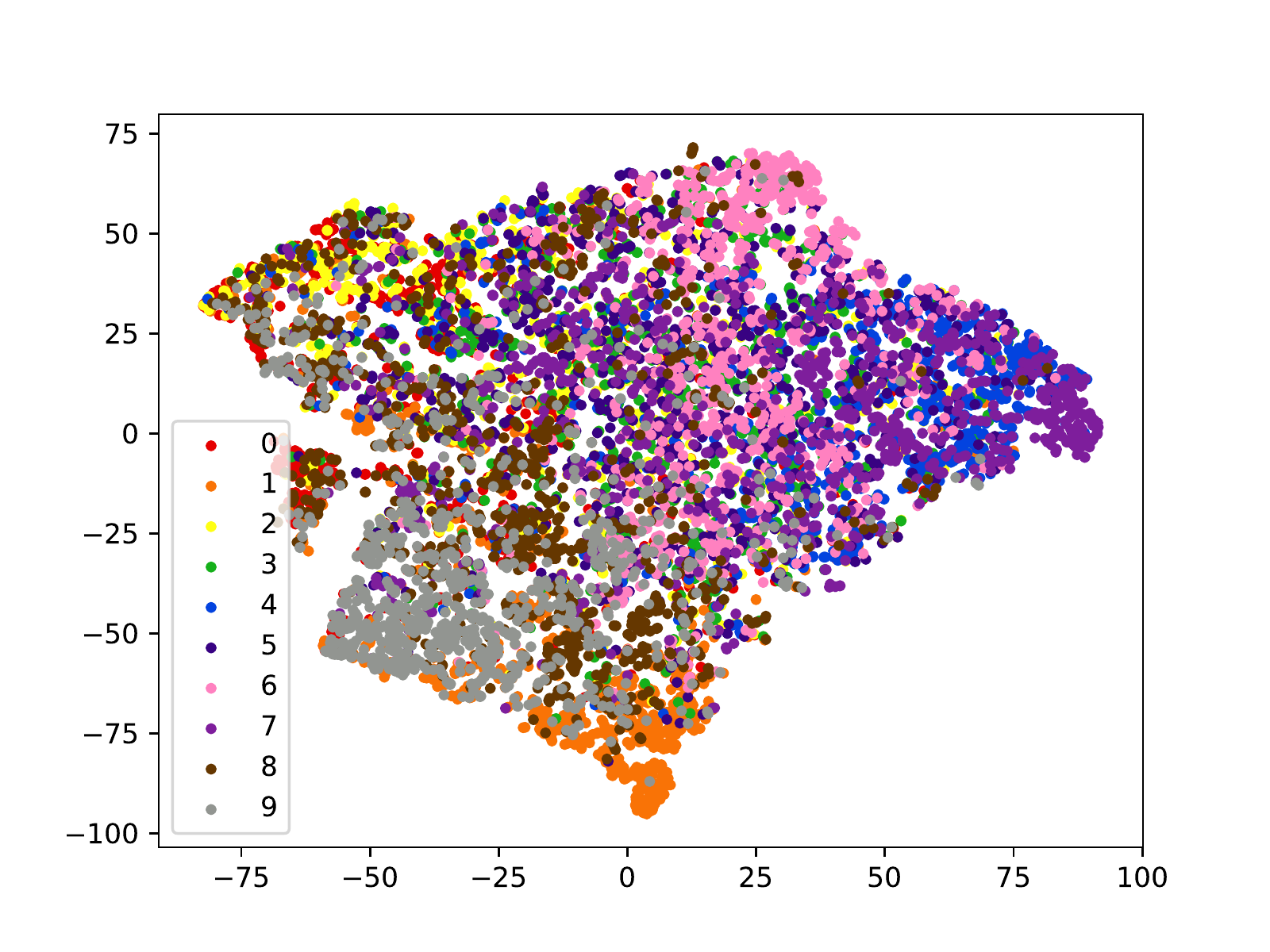}\\
			\mbox{ ({\it b-5}) {CPE-v5}}
		\end{minipage}
		
		\begin{minipage}[h]{32mm}
			\centering
			\includegraphics[width=32mm]{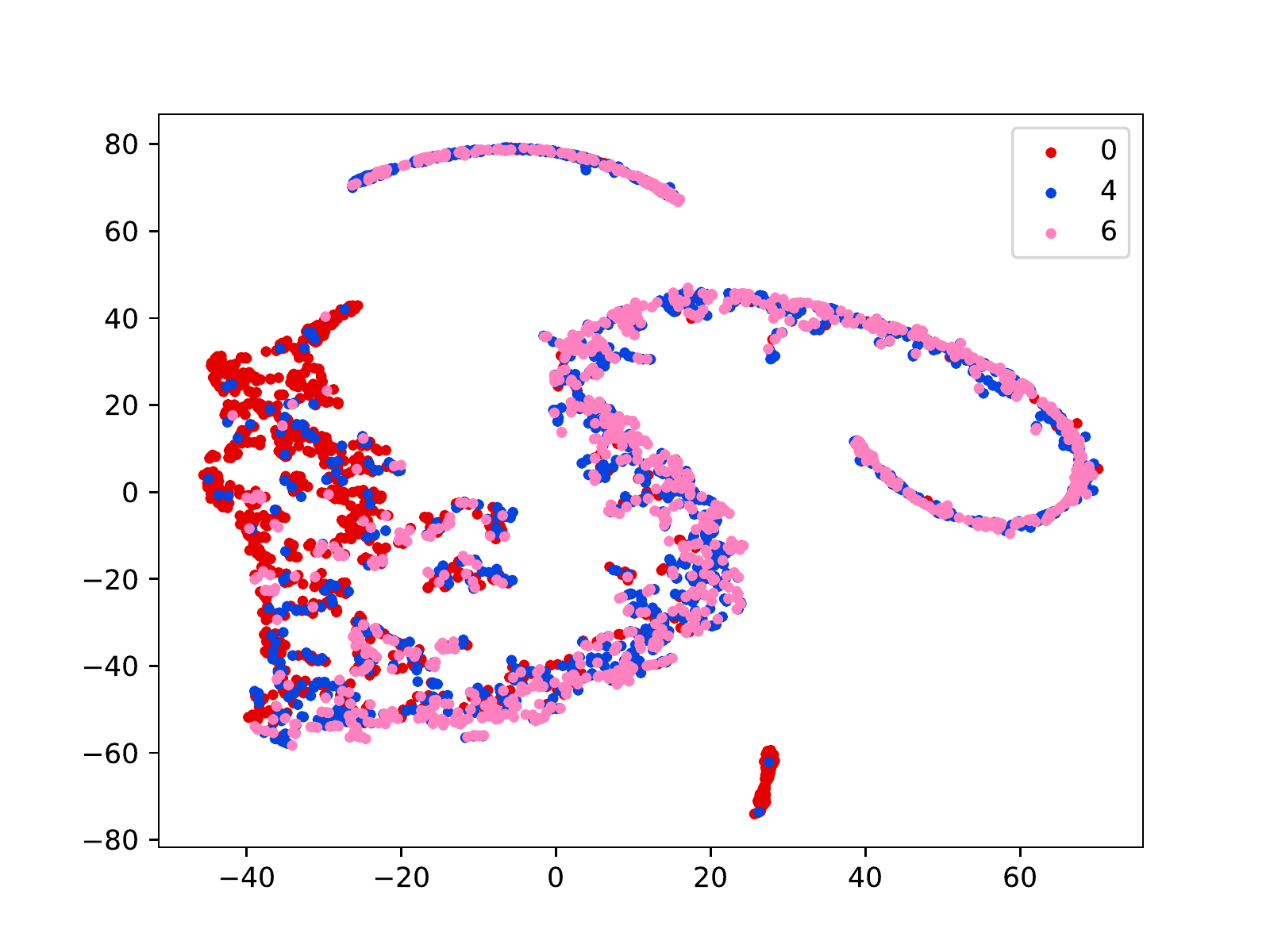}\\
			\mbox{ ({\it c-1}) {DEC-1}}
		\end{minipage}
		\begin{minipage}[h]{32mm}
			\centering
			\includegraphics[width=32mm]{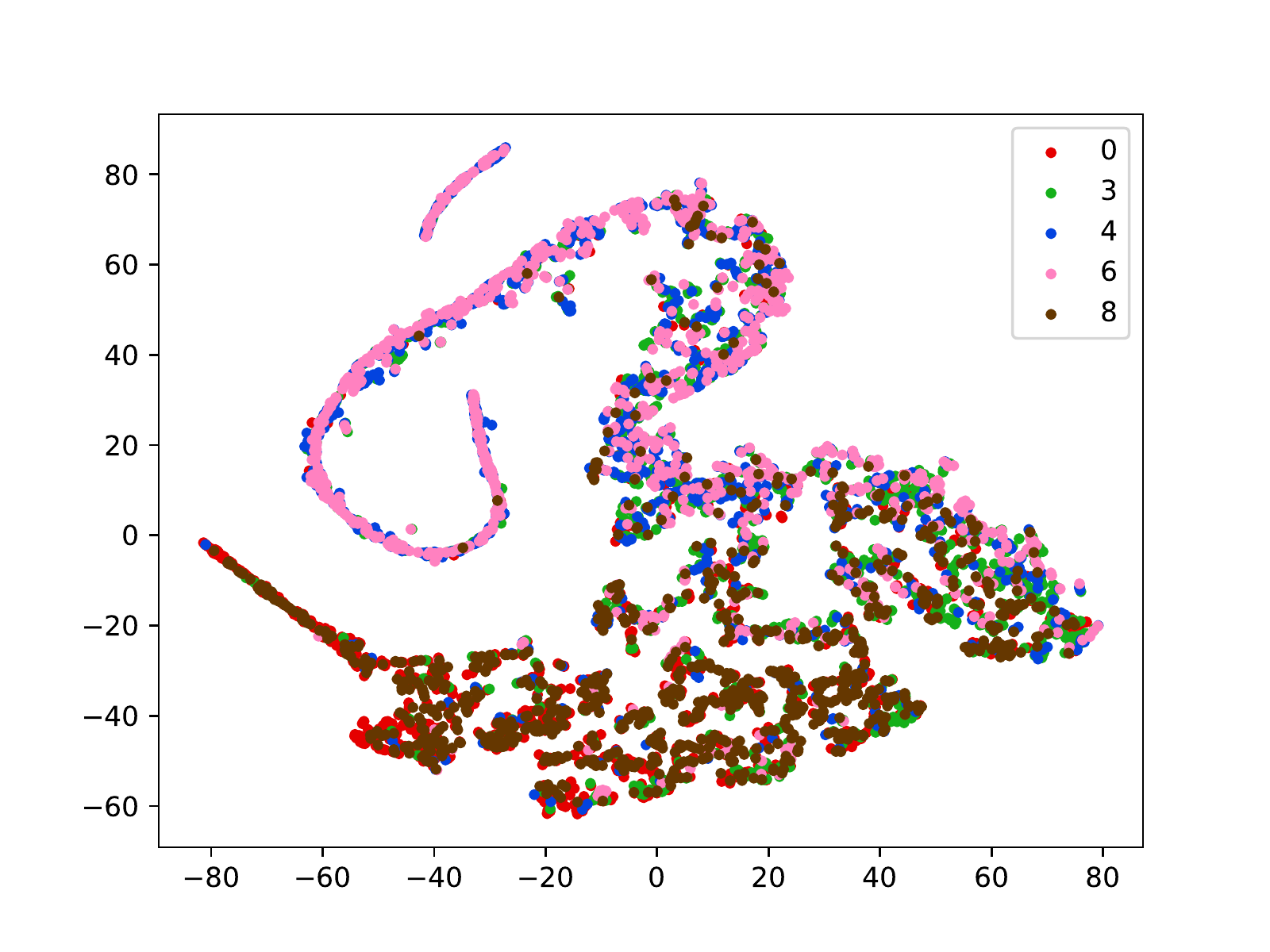}\\
			\mbox{ ({\it c-2}) {DEC-2}}
		\end{minipage}
		\begin{minipage}[h]{32mm}
			\centering
			\includegraphics[width=32mm]{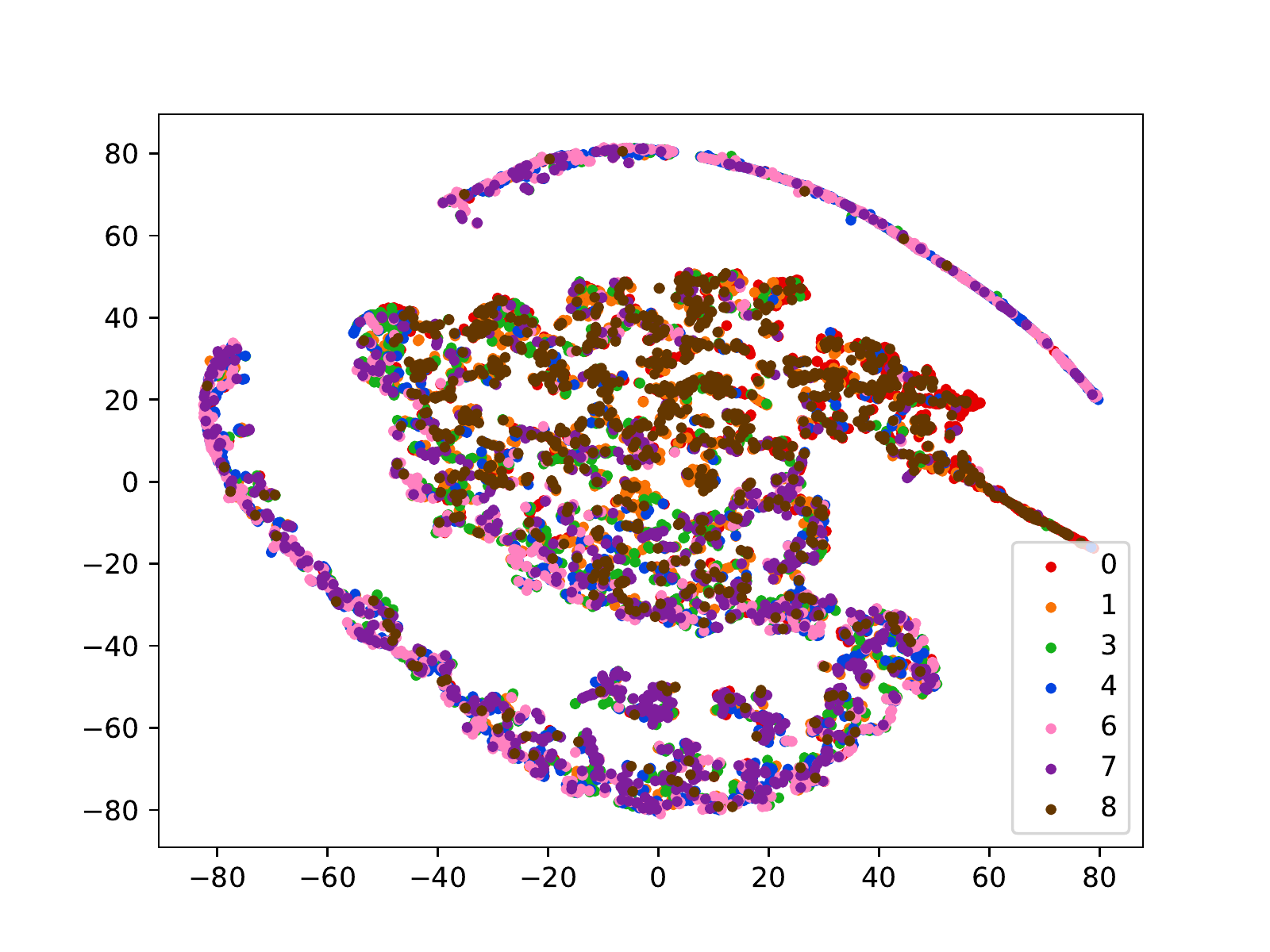}\\
			\mbox{ ({\it c-3}) {DEC-v3}}
		\end{minipage}
		\begin{minipage}[h]{32mm}
			\centering
			\includegraphics[width=32mm]{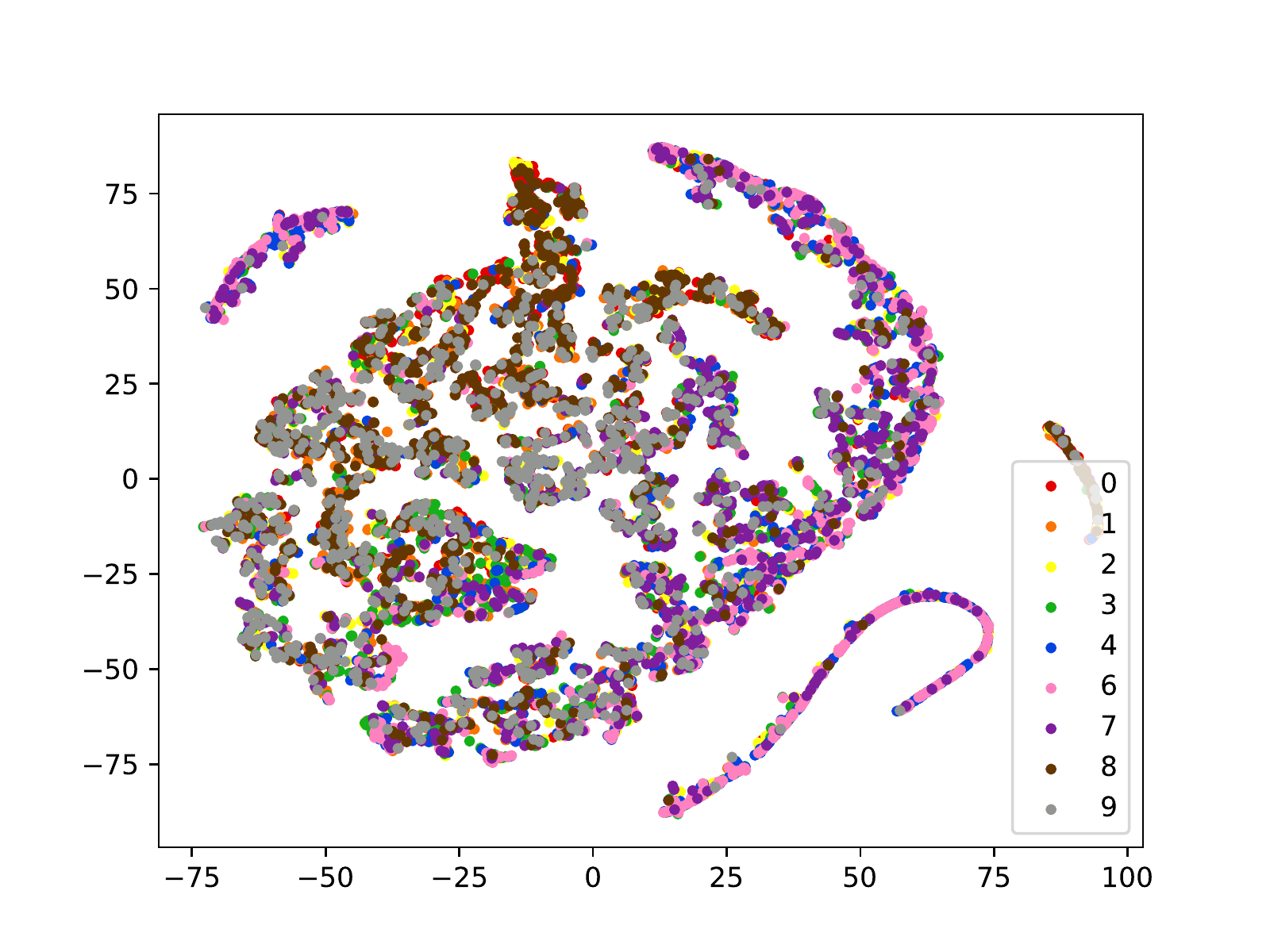}\\
			\mbox{ ({\it c-4}) {DEC-4}}
		\end{minipage}
		\begin{minipage}[h]{32mm}
			\centering
			\includegraphics[width=32mm]{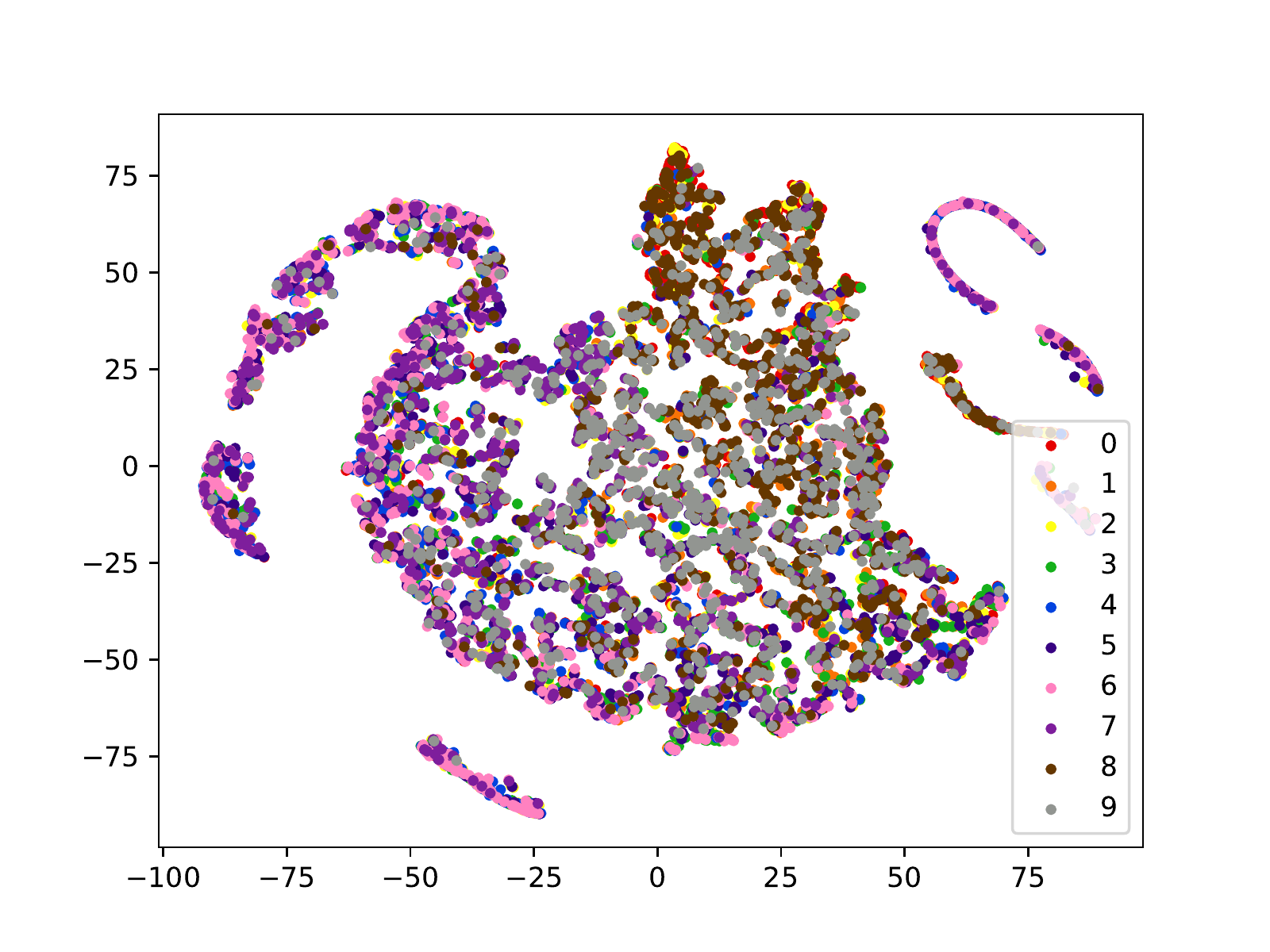}\\
			\mbox{ ({\it c-5}) {DEC-5}}
		\end{minipage}		
		\begin{minipage}[h]{32mm}
			\centering
			\includegraphics[width=32mm]{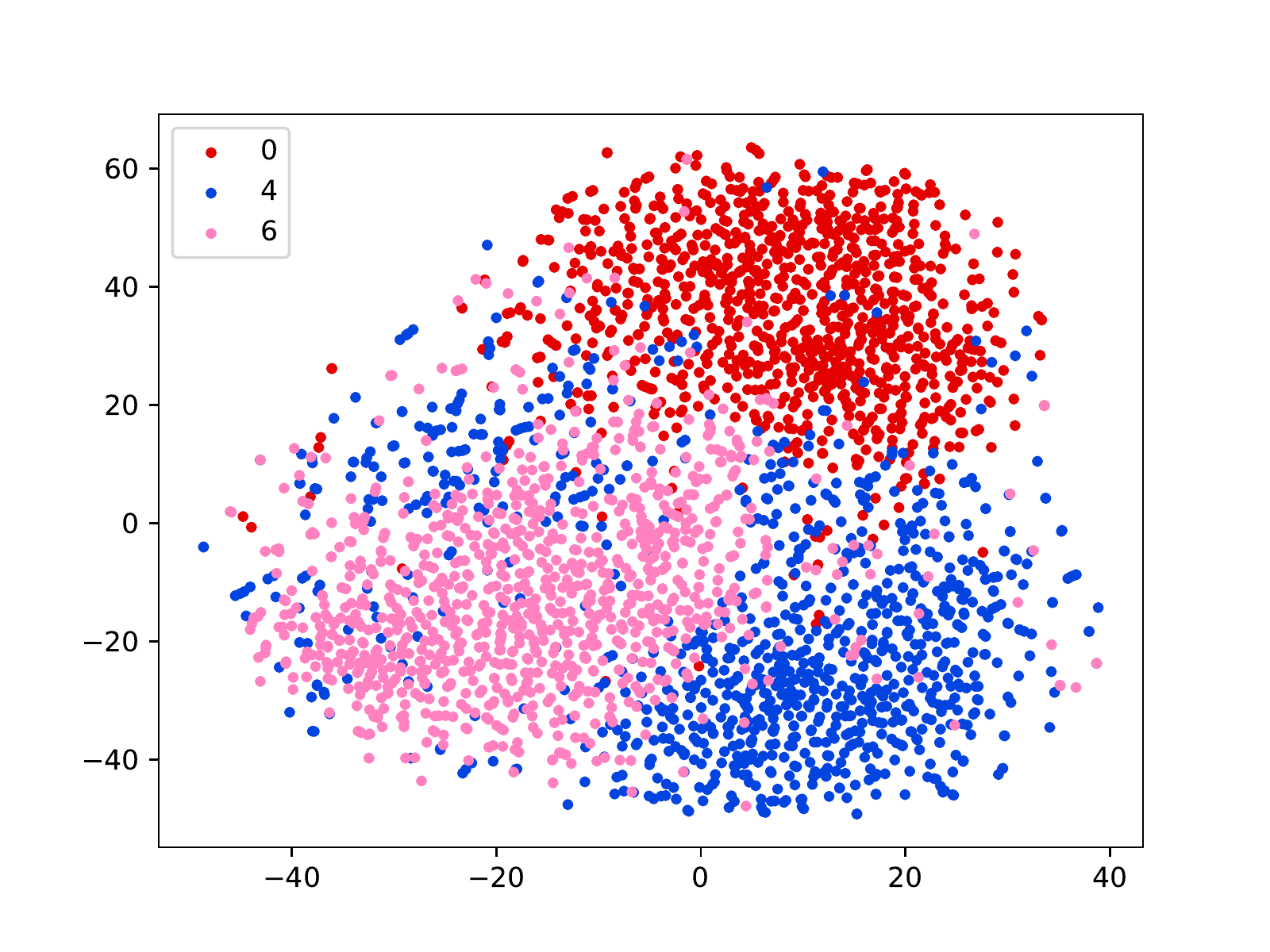}\\
			\mbox{ ({\it d-1}) {CILF-1}}
		\end{minipage}
		\begin{minipage}[h]{32mm}
			\centering
			\includegraphics[width=32mm]{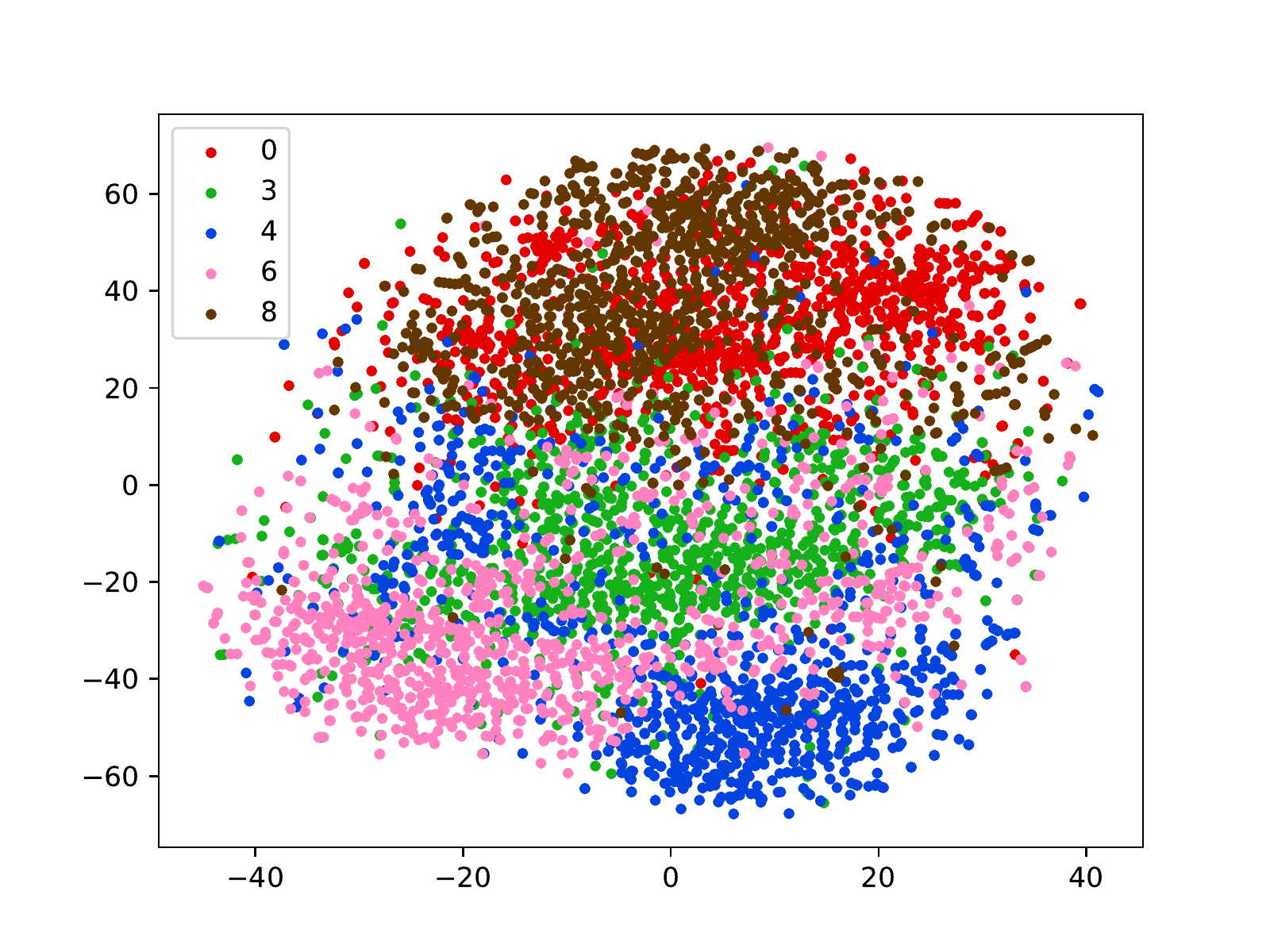}\\
			\mbox{ ({\it d-2}) {CILF-2}}
		\end{minipage}
		\begin{minipage}[h]{32mm}
			\centering
			\includegraphics[width=32mm]{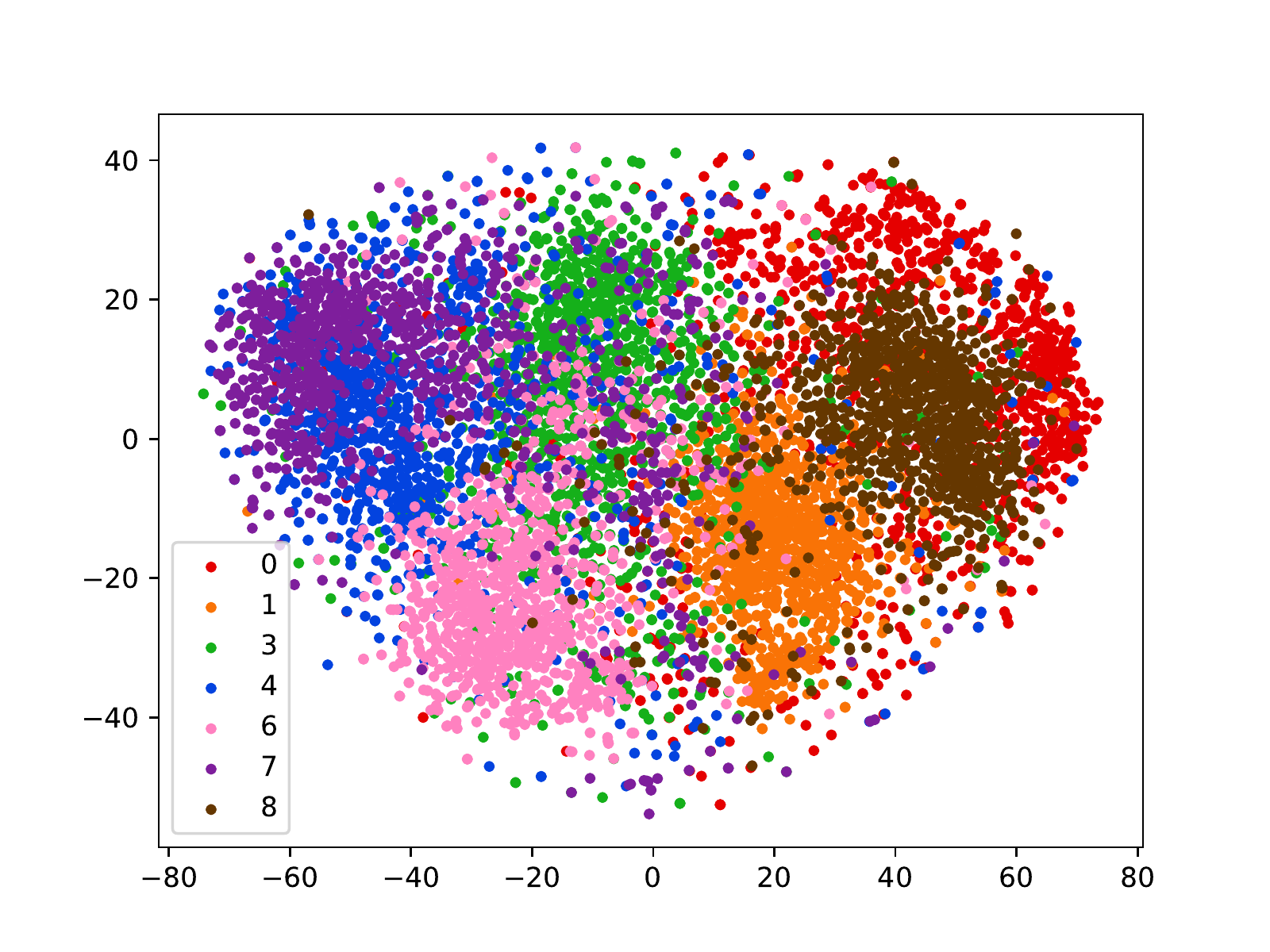}\\
			\mbox{ ({\it d-3}) {CILF-3}}
		\end{minipage}
		\begin{minipage}[h]{32mm}
			\centering
			\includegraphics[width=32mm]{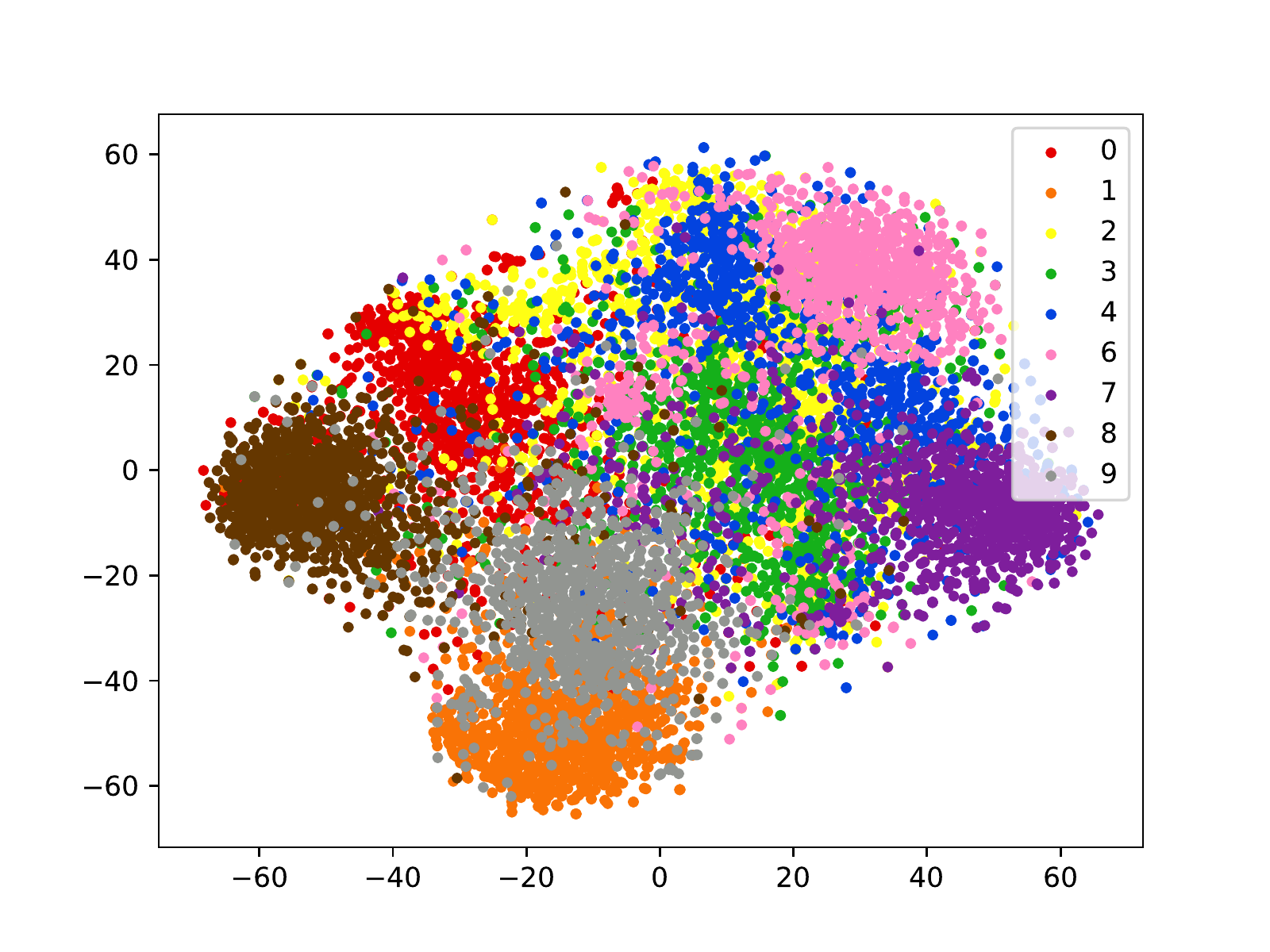}\\
			\mbox{ ({\it d-4}) {CILF-4}}
		\end{minipage}
		\begin{minipage}[h]{32mm}
			\centering
			\includegraphics[width=32mm]{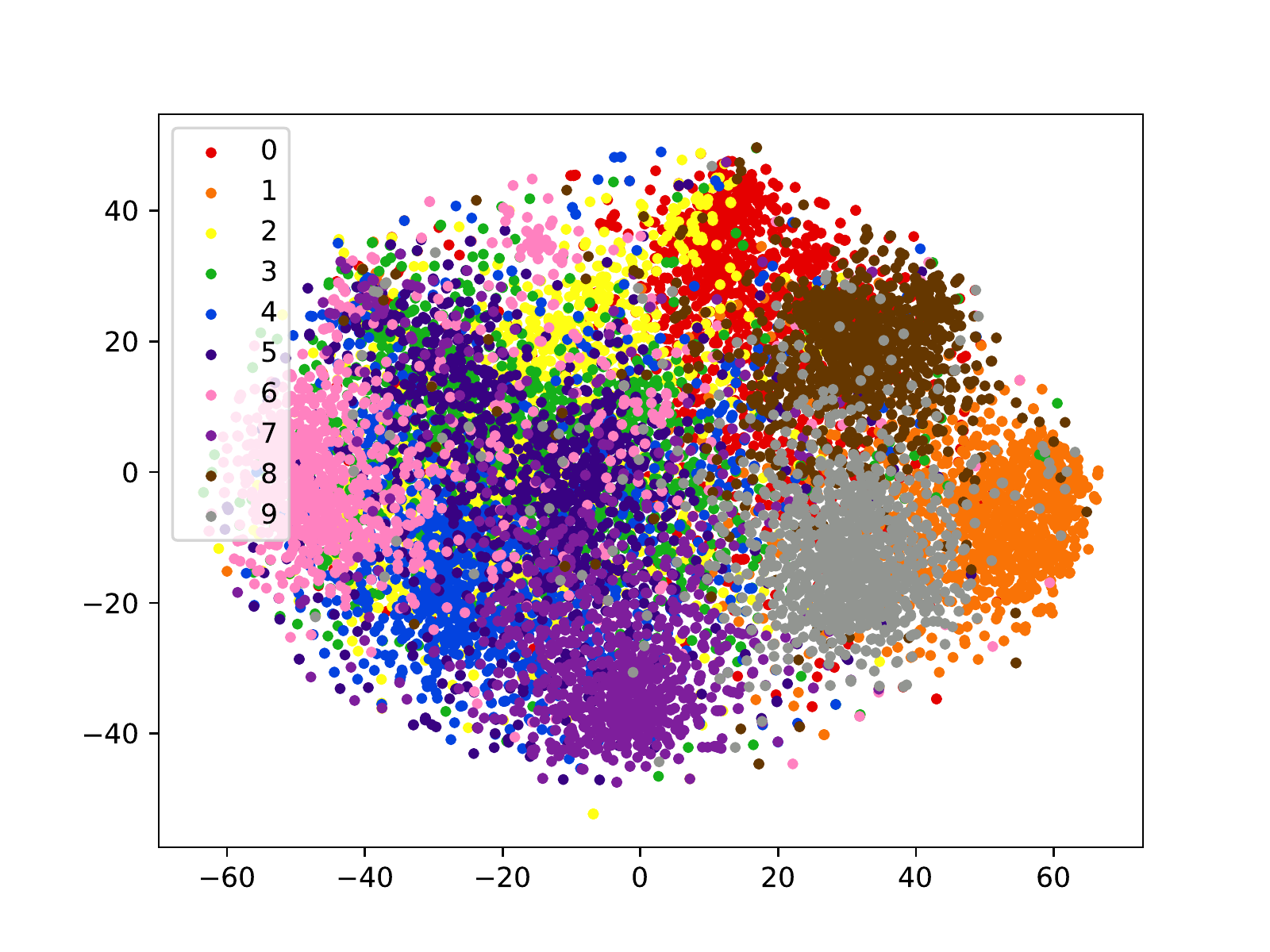}\\
			\mbox{ ({\it d-5}) {CILF-5}}
		\end{minipage}
		
	\end{center}
	\caption{T-SNE Visualization for both known and unknown classes on CIFAR-10 in multiple novel class case. (a) original feature space; (b) Learned representations through single detection method CPE~\cite{WangKCTK19}; (c) Learned representations through multi detection method DEC~\cite{HanVZ19}; (d) Learned representations through proposed CILF. Method$-t$ indicates the T-SNE of $t-$th time window of different methods.}\label{fig:f5}
\end{figure*}

\begin{figure}[t]
	\begin{center}
		\begin{minipage}[h]{44mm}
			\centering
			\includegraphics[width=44mm]{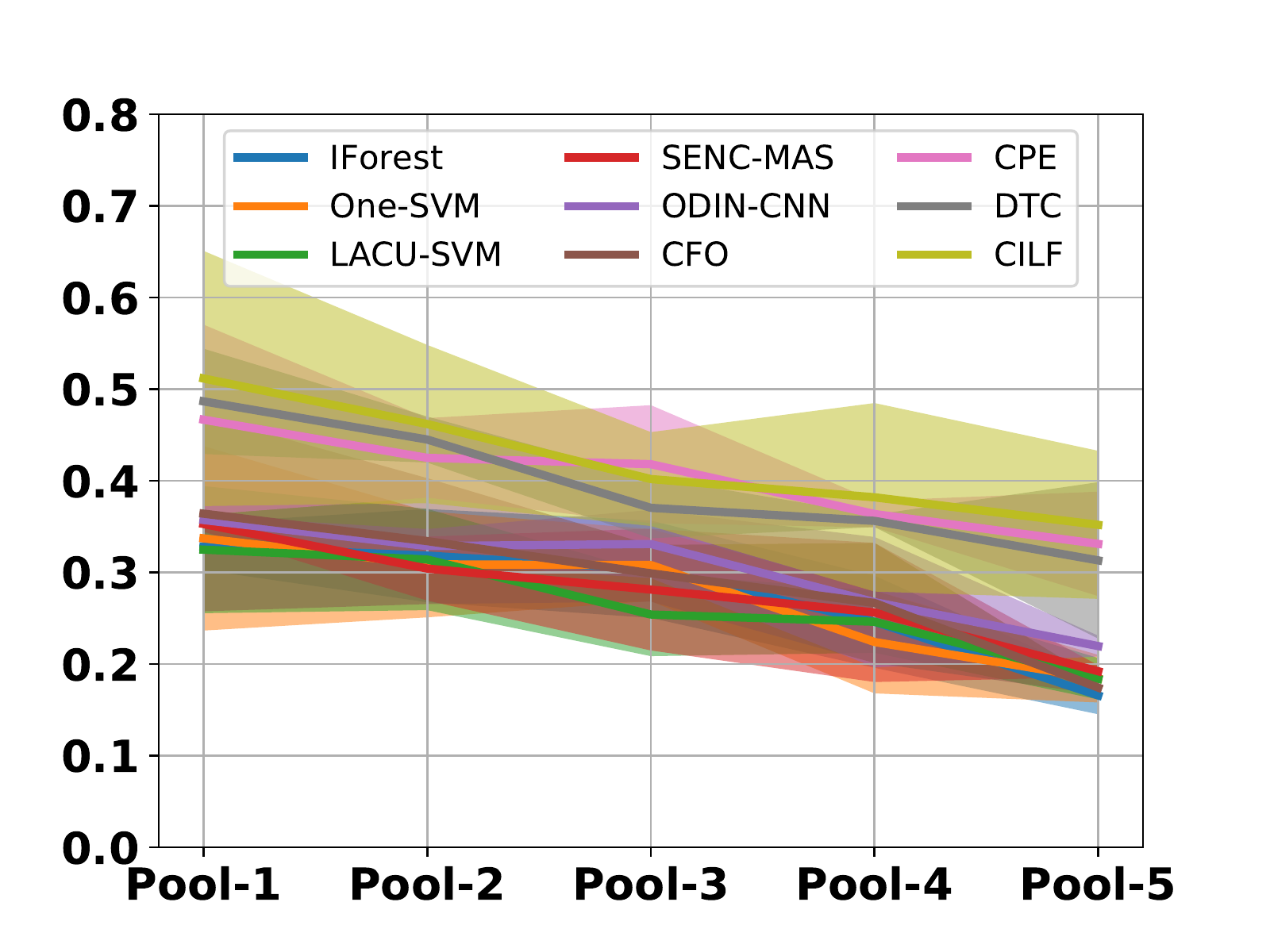}\\
			\mbox{ \;\;\;\; ({\it a}) {NA}}
		\end{minipage}
		\begin{minipage}[h]{44mm}
			\centering
			\includegraphics[width=43mm]{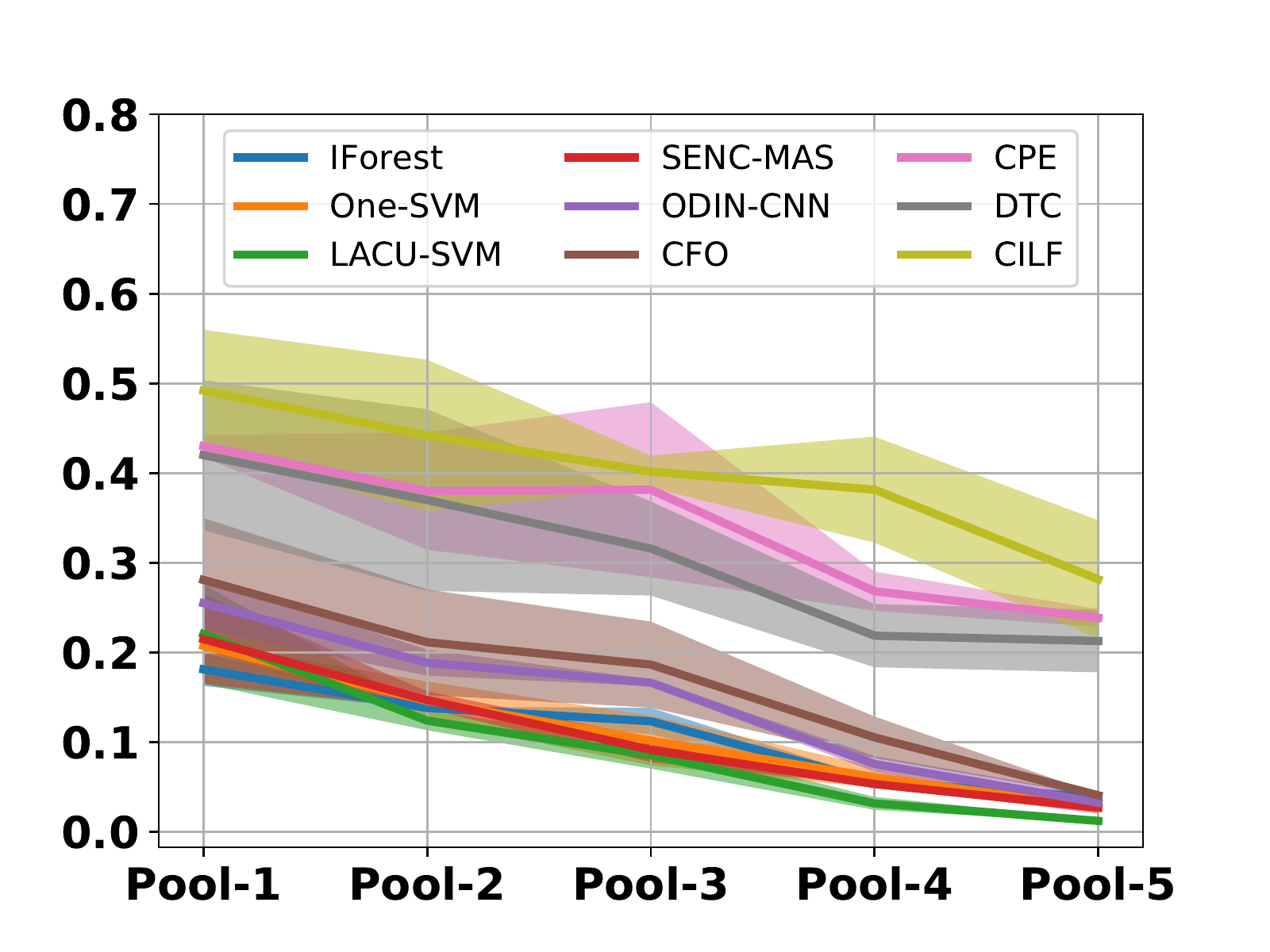}\\
			\mbox{ \;\;\;\; ({\it b}) {Macro-F-Measure}}
		\end{minipage}\\
		\begin{minipage}[h]{44mm}
			\centering
			\includegraphics[width=44mm]{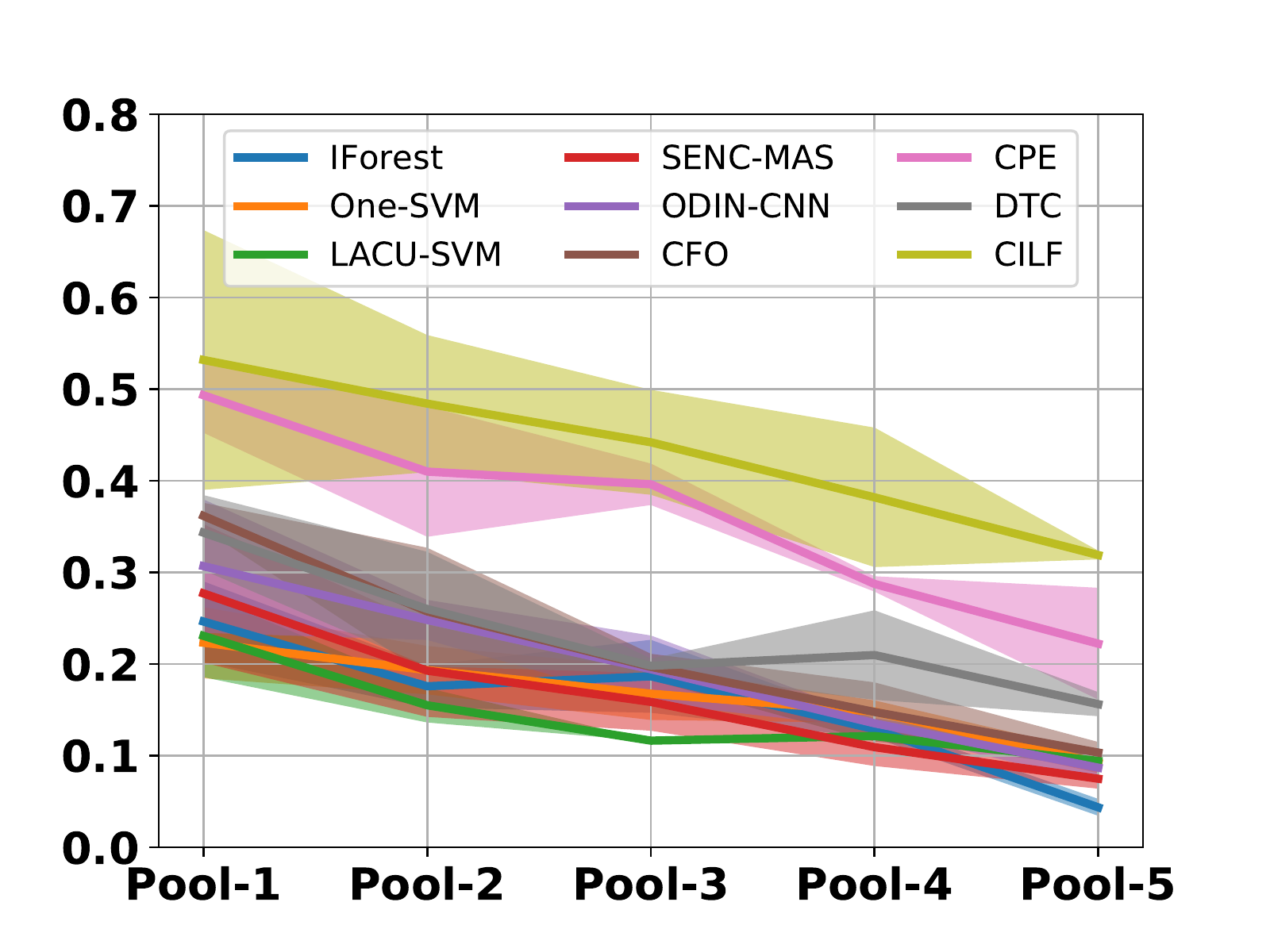}\\
			\mbox{ \;\;\;\; ({\it c}) {Micro-F-Measure}}
		\end{minipage}
		\begin{minipage}[h]{44mm}
			\centering
			\includegraphics[width=43mm]{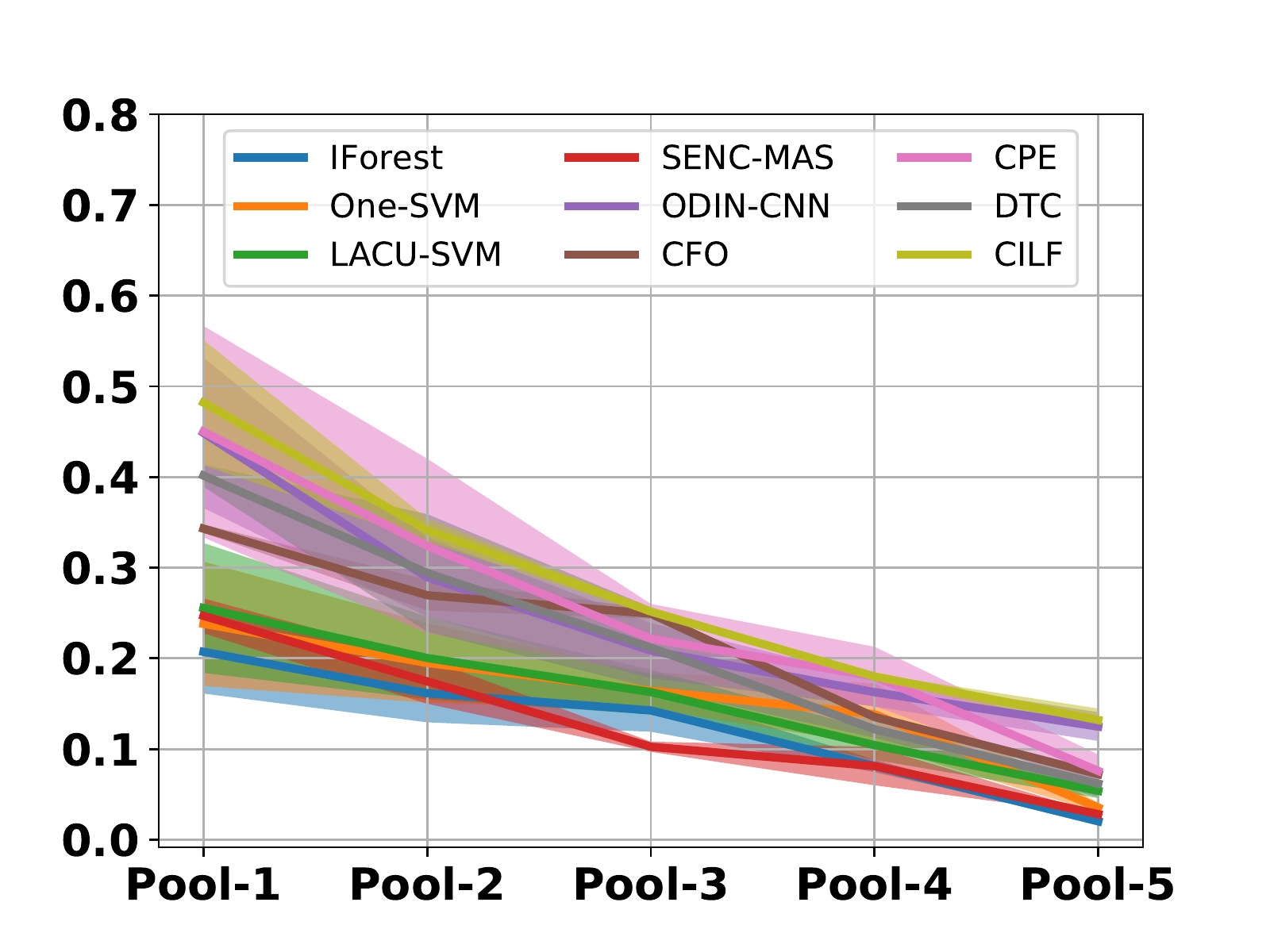}\\
			\mbox{ \;\;\;\; ({\it d}) {AUROC}}
		\end{minipage}\\
	\end{center}
	\caption{Performance criteria on different time window on CIFAR-10 in multiple novel class case.}\label{fig:f6}
\end{figure}

\subsection{Implementation}
We develop CILF based on convolutional network structure as ResNet18~\cite{HeZRS16}. Note that we use an identical set of hyperparameters ($\lambda_1 = 1$, $\lambda_2 = 1$, $\alpha = 0.3$, $\beta= 0.8$, $\upsilon = 0.2$, $\delta = 3$, $\phi = 10$). In all of our models and experiments, we adopt standard SGD with Nesterov momentum~\cite{SutskeverMDH13}, where the momentum is 0.9. We train the initial model $f$ as following: the number of epochs is 20, the batch size is 128, the learning rate is 0.01, and weight decay is 0.001. We implement all baselines and perform all experiments based on code released by corresponding authors. For CNN based methods, we use the same network architecture and parameters during training, such as optimizer, learning rate schedule, and data pre-processing. Our method is implemented on a RTX 2080TI GPU with Pytorch 0.4.06~\footnote{https://pytorch.org/}.


\subsection{Single Novel Class Detection}
Table \ref{tab:tab1} compares the detection performance of CILF with all baseline methods on each streaming data with single novel class. We observe that: (1) CNN-based methods are better than traditional detection approaches, i.e., One-SVM, LACU-SVM, SENC-MAS. This indicates that neural network provides superior feature embeddings for prediction and detection over high dimensional streaming data; (2) CILF consistently outperforms all compared CNN-based methods in all the criteria. For example, in CIFAR-10, CILF provides at least 2$\%$ improvements than other methods. This indicates the effectiveness of prototype based loss for feature embedding and curriculum clustering operator for detection; and (3) The detection performance for large-scale data sets still needs to be improved, and the results of all methods are low.

Figure \ref{fig:f3} shows feature embedding results within each time window using T-SNE~\cite{maaten2008visualizing}, in which each class randomly samples 800 instances. Note that we turn to utilize the more complex dataset CIFAR-10 as an example, rather than simpler MNIST dataset in previous methods. Clearly, the optimal discriminative method (CPE) and generative method (DEC) are greatly interfered by the embedding confusion. While the output of CILF has more distinct groups from different classes compared to other methods, which is attributed to the prototype based loss for model training. Moreover, instances from novel classes are well separated from other known clusters, which is benefit for novel class detection. The compactness of new class indicates the effectiveness of curriculum clustering operator, in which reliable prototypes are developed and the the model is fine-tuned from easy to difficult.    

Table \ref{tab:tab2} compared the forgetting performance of CILF with all baseline methods, which defines the forgetting of emerge class on a particular window, i.e., the difference between maximum knowledge gained about that window throughout learning process and we currently have about it, the lower difference the better. The results show that CILF has the least forgetting, which validates that the memory distillation and prototype regularization can mitigate the forgetting of known class data. Moreover, Figure \ref{fig:f4} gives more direct results. Due to page limitation, we only report the result of CIFAR-10. The results indicate that at different window, the performance of known classes falls slower, which shows that CILF can mitigate forgetting efficiently.

\begin{figure}[t]
	\begin{center}
		\begin{minipage}[h]{44mm}
			\centering
			\includegraphics[width=44mm]{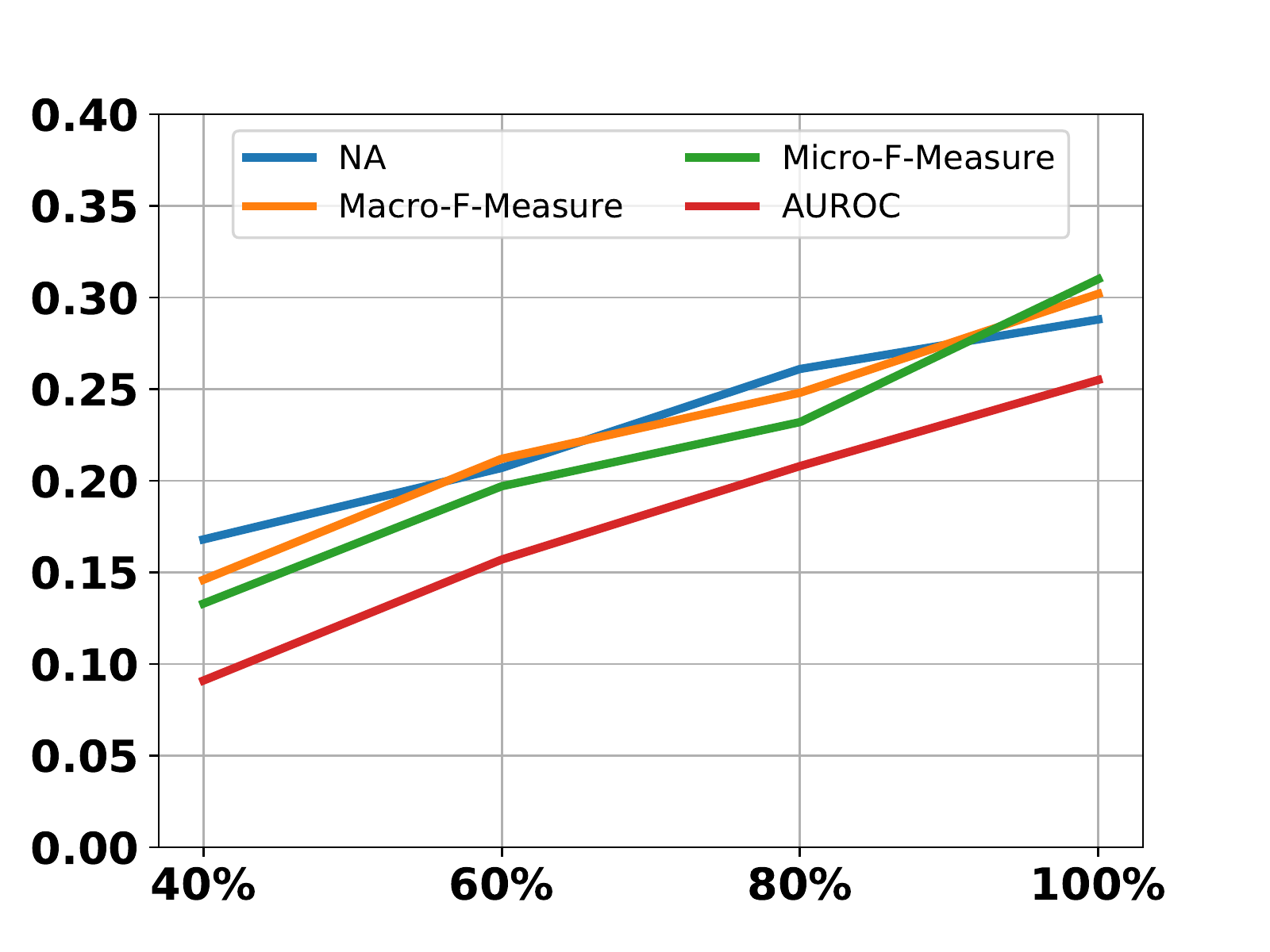}\\
			\mbox{ \;\;\;\; ({\it a}) {Single CIFAR-10}}
		\end{minipage}
		\begin{minipage}[h]{44mm}
			\centering
			\includegraphics[width=43mm]{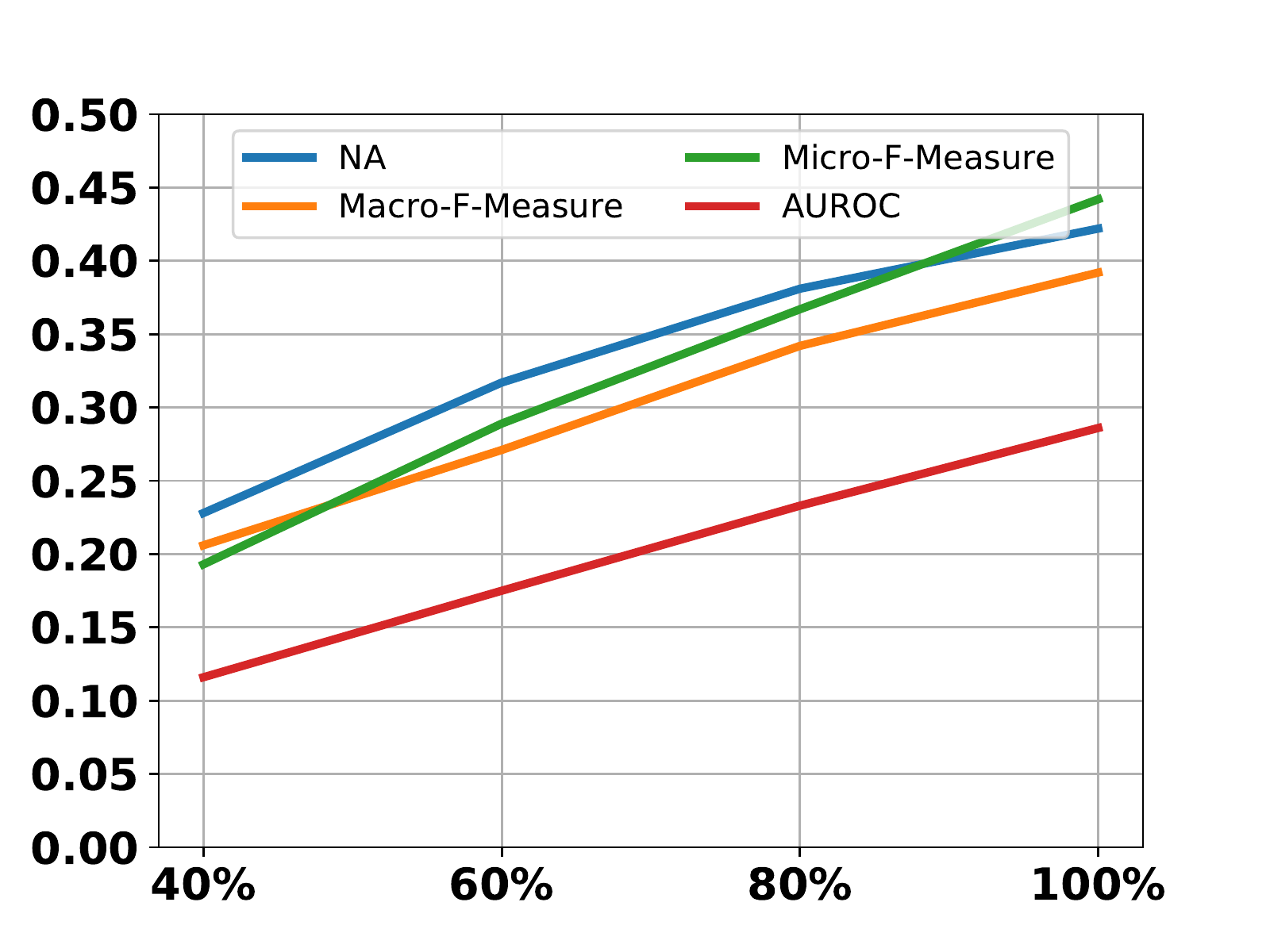}\\
			\mbox{ \;\;\;\; ({\it d}) {Multiple CIFAR-10}}
		\end{minipage}
	\end{center}
	\caption{Relationship between detection performance during stream with label request percentage for every time window.}\label{fig:f7}
\end{figure}

\subsection{Multiple Novel Class Detection}
Table \ref{tab:tab3} compares the detection performance of CILF with all baseline methods considering multiple novel classes. We observe that: (1) multiple novel class detection method, i.e., DEC, has not achieved an advantage than single novel class detection methods with subsequent clustering operator. This indicates that direct clustering method may be influenced by the embedding confusion; and (2) CILF consistently outperforms all compared CNN-based methods in all the criteria except AUROC on CIFAR-50. This further indicates the effectiveness of curriculum clustering operator for detection. 

Figure \ref{fig:f5} shows feature embedding results within each time window using T-SNE, in which each class randomly samples 800 instances. Similarly, the output of CILF has more distinct groups from different classes compared to other methods, which indicates that CILF can solve the embedding confusion effectively. Table \ref{tab:tab4} and Figure \ref{fig:f6} compared the forgetting performance of CILF with baseline methods. Identically, the results show that CILF has the least forgetting, and performance of known classes fall slower, which shows that CILF can mitigate forgetting under multiple novel classes scenario.

\subsection{Influence of Query Size}
Figure \ref{fig:f7} shows the influence of querying number about potential novel class instances, we only give the results of CIFAR-10 considering page limitation. Here, we randomly query a subset, i.e., a percent of potential instances from the current window. From the figure, it can be observed that the prediction performance improves with the increase of labeled data, which verifies the importance of ground-truths for model update.



\begin{figure}[t]
	\begin{center}
		\begin{minipage}[h]{44mm}
			\centering
			\includegraphics[width=44mm]{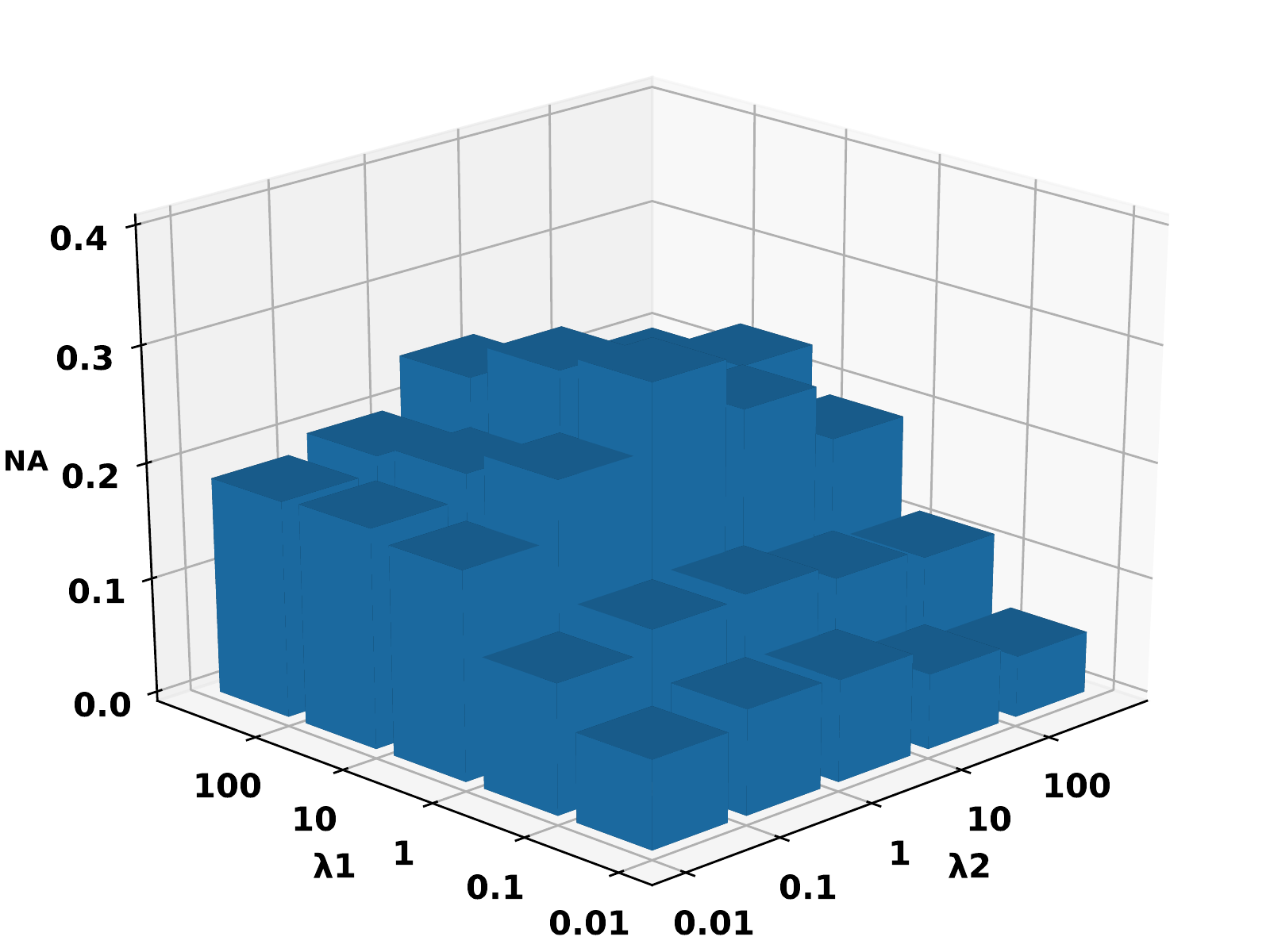}\\
			\mbox{ \;\;\;\; ({\it a}) {Single CIFAR-10}}
		\end{minipage}
		\begin{minipage}[h]{44mm}
		\centering
		\includegraphics[width=44mm]{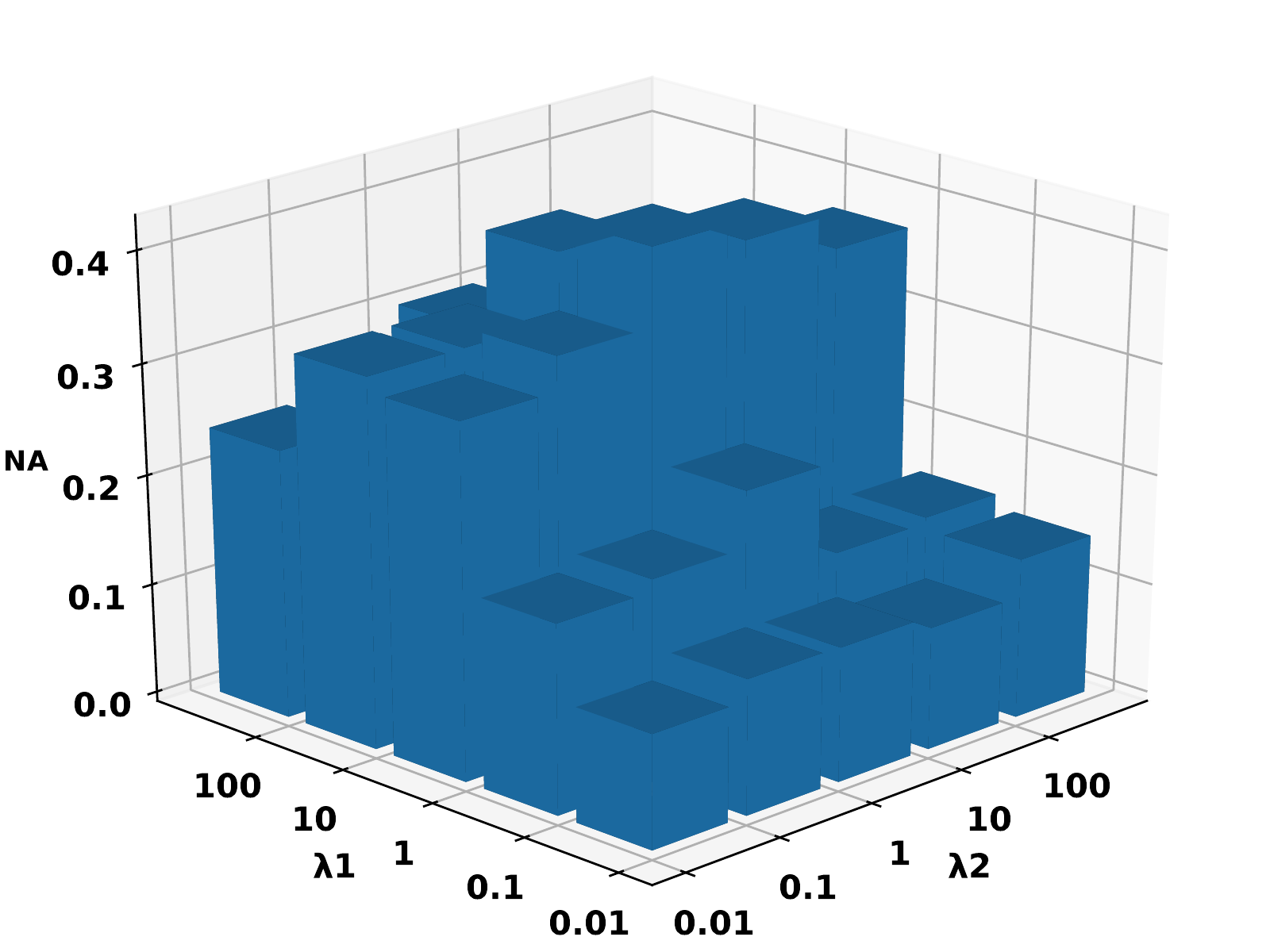}\\
		\mbox{ \;\;\;\; ({\it b}) {Multiple CIFAR-10}}
		\end{minipage}
		\end{center}
	\caption{Parameter sensitivity of $\lambda_1$ and $\lambda_2$ for the CIFAR-10 in novel detection. (a) is single novel class case, (c) is multiple novel class case.}\label{fig:f8}
\end{figure}

\subsection{Parameter Sensitivity}
The main parameters in novel class detection and model update are the $\lambda_1$ and $\lambda_2$ in Eq. \ref{fig:f8}. We vary these parameters in $\{0.01, 0.1, 1, 10, 100\}$ to study its sensitivity to classification performance and record the AUROC results in figure \ref{fig:f8}. Both the single and multiple cases indicate that the performances are higher when setting $\lambda_1$ with a larger value, i.e., larger than 1.

\begin{figure}[t]
	\begin{center}
	\begin{minipage}[h]{44mm}
		\centering
		\includegraphics[width=44mm]{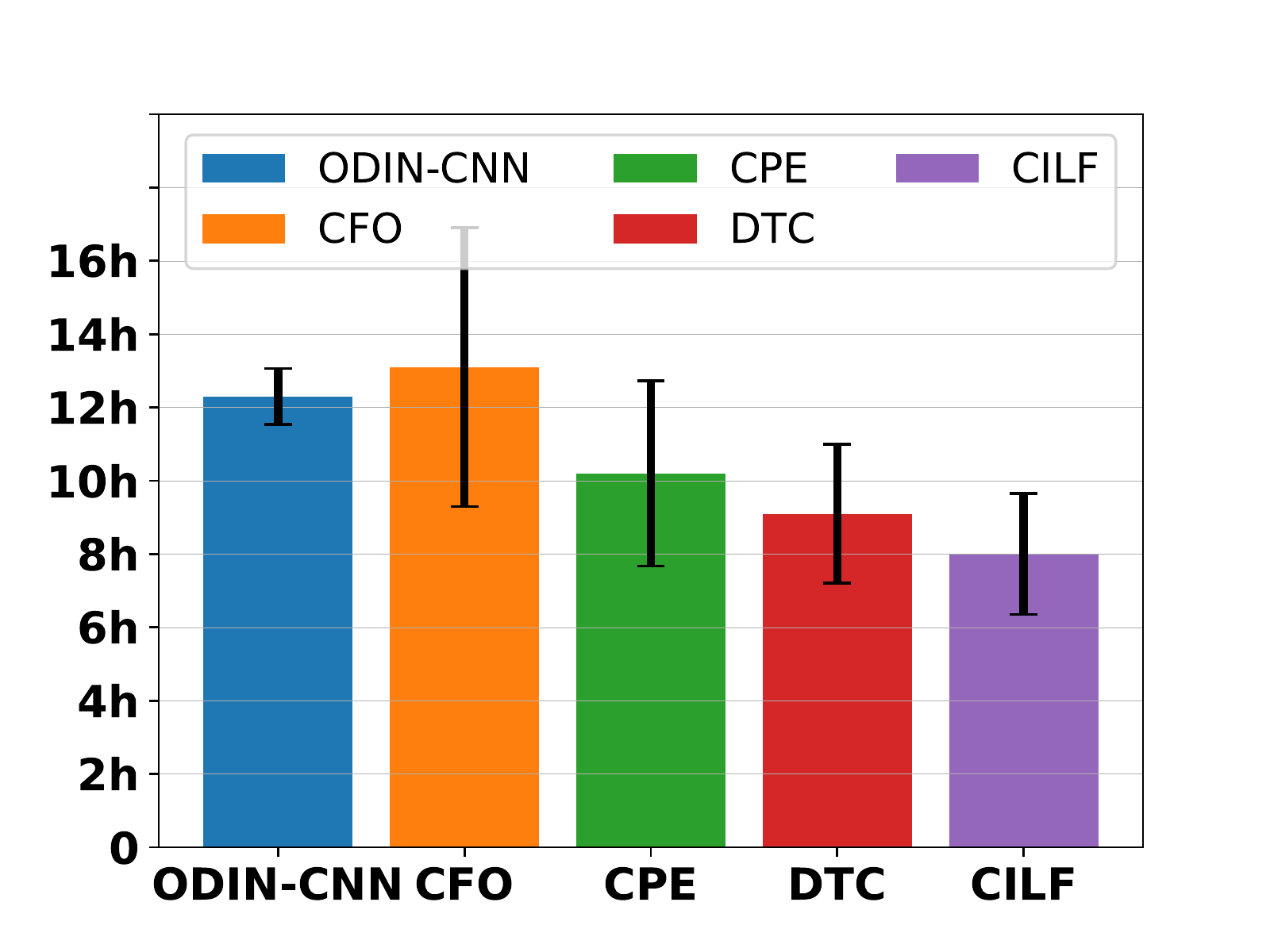}\\
		\mbox{ \;\;\;\; ({\it c}) {Multiple CIFAR-10}}
	\end{minipage}
	\begin{minipage}[h]{44mm}
		\centering
		\includegraphics[width=44mm]{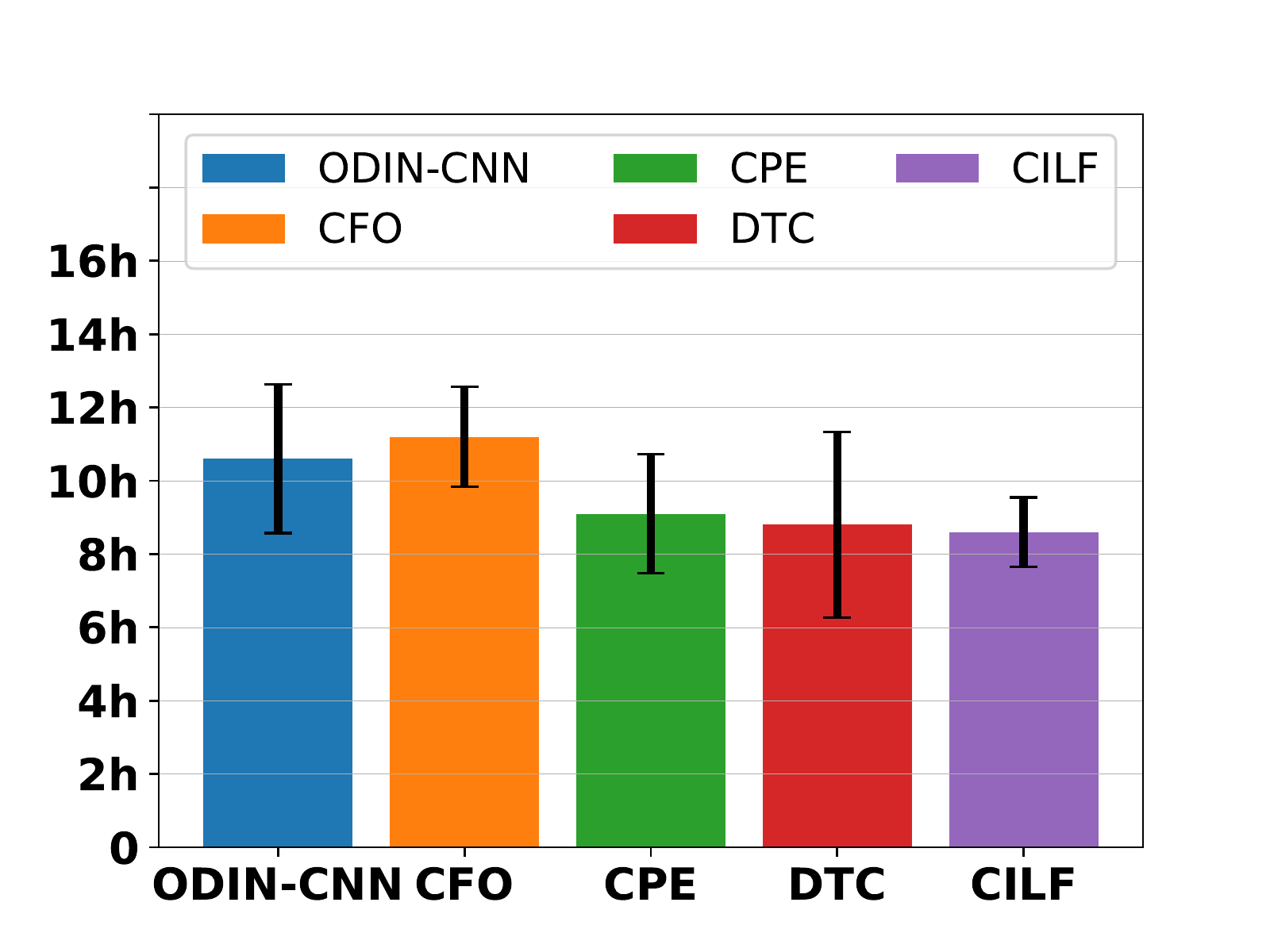}\\
		\mbox{ \;\;\;\; ({\it d}) {Multiple CIFAR-50}}
	\end{minipage}
	\end{center}
	\caption{Execution time analysis.}\label{fig:f9}
\end{figure}
\subsection{Execution Time for Model Update}
Consider our method focuses on multiple novel class detection, thus we analyze execution time for detecting and updating model with multiple novel class case. In detail, we select the five deep methods, i.e. ODIN-CNN, CFO, CPR, DTC and CILF, and record the results of multiple novel class case in Figure \ref{fig:f9}. CILF achieves the fastest results, this is because other methods require additional clustering operations, and embedding confusion will slow down the clustering convergence, which indicates the curriculum clustering can accelerate detection. 

\section{Conclusion}
Real-word application always receive the data in stream form, which emerges previously unknown classes sequentially. Incremental NCD has two main challenges: 1) Novel class detection, streaming test data will accept unknown classes; 2) Model expansion, the model needs to be effectively updated after the new class detection. However, traditional methods have not always fully considered these two challenges. To this end, we propose a Class-Incremental Learning without Forgetting (CILF) framework. CILF designed to regularize classification with decoupled prototype based loss, which can improve the intra-class and inter-class structure significantly, and acquire a compact embedding representation for novel class detection in result. Then, CILF employed a learnable curriculum clustering operator to estimate the number of semantic clusters via fine-tuning the learned network. Last, CILF updates the network effectively with robust regularization to mitigate the catastrophic forgetting. Consequently, empirical studies showed the superior performances of CILF.



\ifCLASSOPTIONcompsoc

\ifCLASSOPTIONcaptionsoff
  \newpage
\fi

\bibliographystyle{IEEEtranN}{\small
\bibliography{CILF}}



\vspace{-1.4cm}
\begin{IEEEbiography}[{\includegraphics[width=1in,height=1.25in,clip,keepaspectratio]{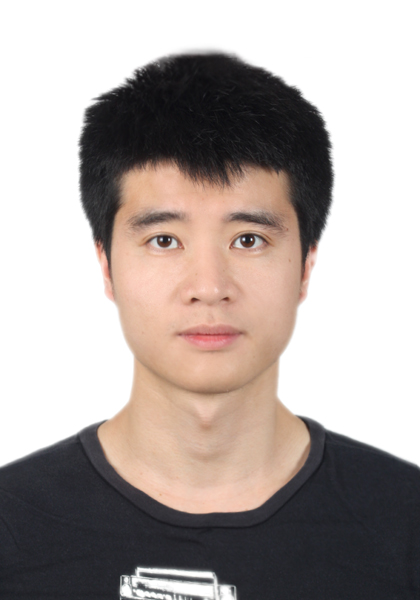}}]{Yang Yang}
received the Ph.D. degree in computer science, Nanjing University, China in 2019. At the same year, he became a faculty member at Nanjing University of Science and Technology, China. He is currently a Professor with the Computer Science and Engineering. His research interests lie primarily in machine learning and data mining, including heterogeneous learning, model reuse, and incremental mining. He serves as PC in leading conferences such as IJCAI, AAAI, ICML, NIPS, etc.
\end{IEEEbiography}
\vspace{-1.4cm}
\begin{IEEEbiography}[{\includegraphics[width=1in,height=1.25in,clip,keepaspectratio]{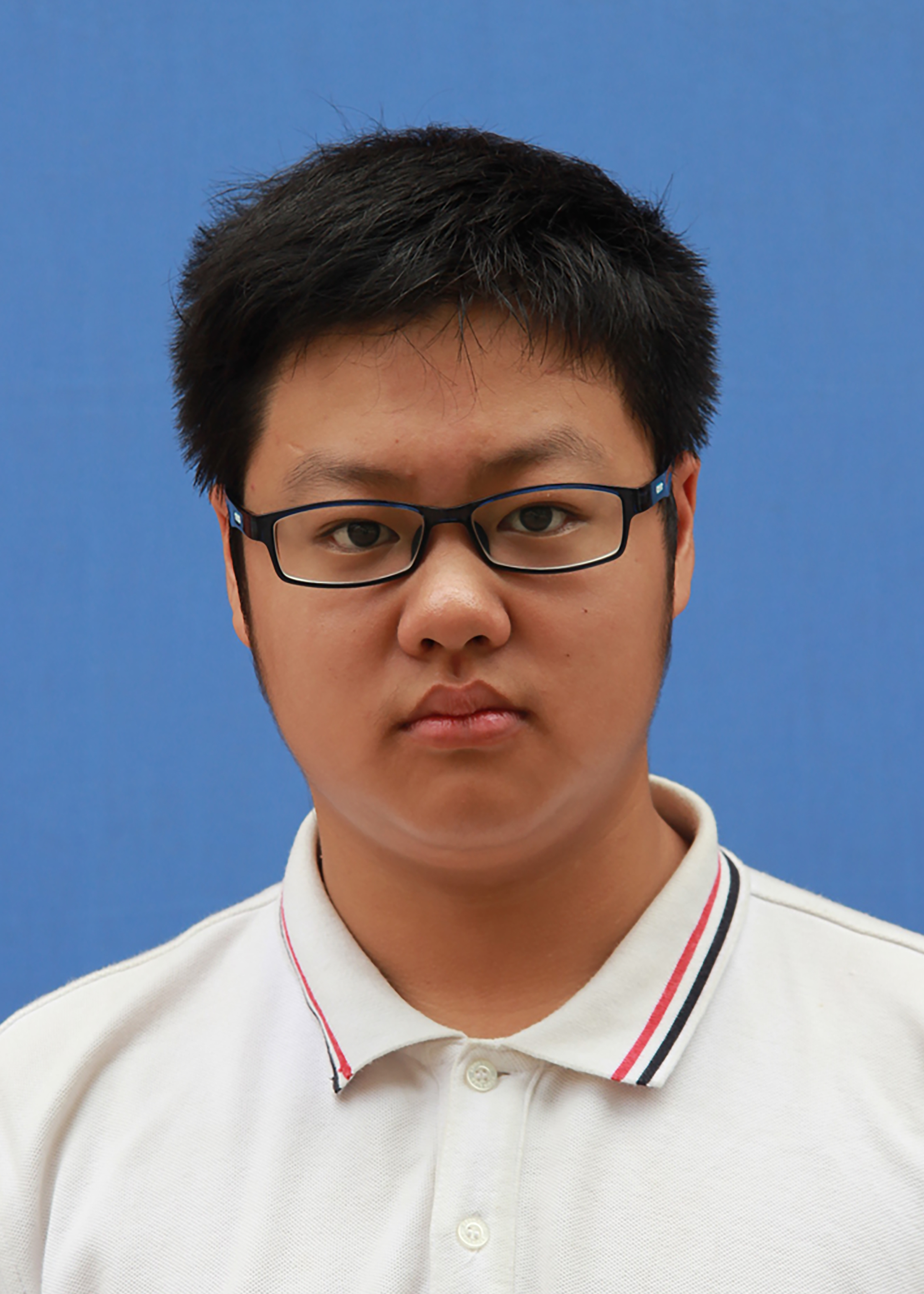}}]{Zhen-Qiang Sun}
is working towards the M.Sc. degree with the School of Computer Science $\&$ Technology in Nanjing Normal University, China. His research interests lie primarily in machine learning and data mining, including incremental learning.
\end{IEEEbiography}
\vspace{-1.4cm}
\begin{IEEEbiography}[{\includegraphics[width=1in,height=1.25in,clip,keepaspectratio]{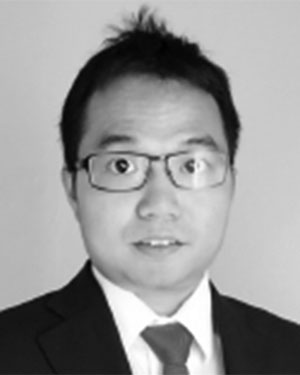}}]{Yanjie Fu}
	received the BE degree from the University of Science and Technology of China, in 2008, the ME degree from the Chinese Academy
	of Sciences, in 2011, and the PhD degree from 	Rutgers University, in 2016. He is currently an assistant professor with the Missouri University of Science and Technology. His general interests are data mining and big data analytics. He has 	published prolifically in refereed journals and conference proceedings, such as the IEEE Transactions on Knowledge and Data Engineering, the ACM Transactions on Knowledge Discovery from Data, the IEEE Transactions on Mobile Computing, and ACM SIGKDD. 
\end{IEEEbiography}
\vspace{-2cm}
\begin{IEEEbiography}[{\includegraphics[width=1in,height=1.25in,clip,keepaspectratio]{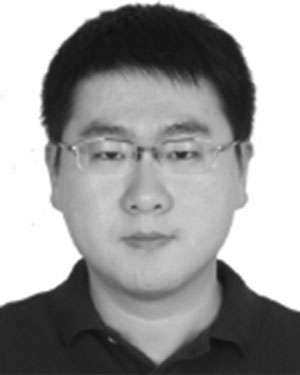}}]{HengShu Zhu}(SM’19)
	is currently a principal data scientist $\&$ architect at Baidu Inc. He received the Ph.D. degree in 2014 and B.E. degree in 2009, both in Computer Science from University of Science and Technology of China (USTC), China. His general area of research is data mining and machine learning, with a focus on developing advanced data analysis techniques for innovative business applications. He has published prolifically in refereed journals and conference proceedings, including IEEE Transactions on Knowledge and Data Engineering (TKDE), IEEE Transactions on Mobile Computing (TMC), ACM Transactions on Information Systems (ACM TOIS), ACM Transactions on Knowledge Discovery from Data (TKDD), ACM SIGKDD, ACM SIGIR, WWW, IJCAI, and AAAI. He has served regularly on the organization and program committees of numerous conferences, including as a program co-chair of the KDD Cup-2019 Regular ML Track, and a founding co-chair of the first International Workshop on Organizational Behavior and Talent Analytics (OBTA) and the International Workshop on Talent and Management Computing (TMC), in conjunction with ACM SIGKDD. He was the recipient of the Distinguished Dissertation Award of CAS (2016), the Distinguished Dissertation Award of CAAI (2016), the Special Prize of President Scholarship for Postgraduate Students of CAS (2014), the Best Student Paper Award of KSEM-2011, WAIM-2013, CCDM-2014, and the Best Paper Nomination of ICDM-2014. He is the senior member of IEEE, ACM, and CCF. 
\end{IEEEbiography}
\vspace{-1.2cm}
\begin{IEEEbiography}[{\includegraphics[width=1in,height=1.25in,clip,keepaspectratio]{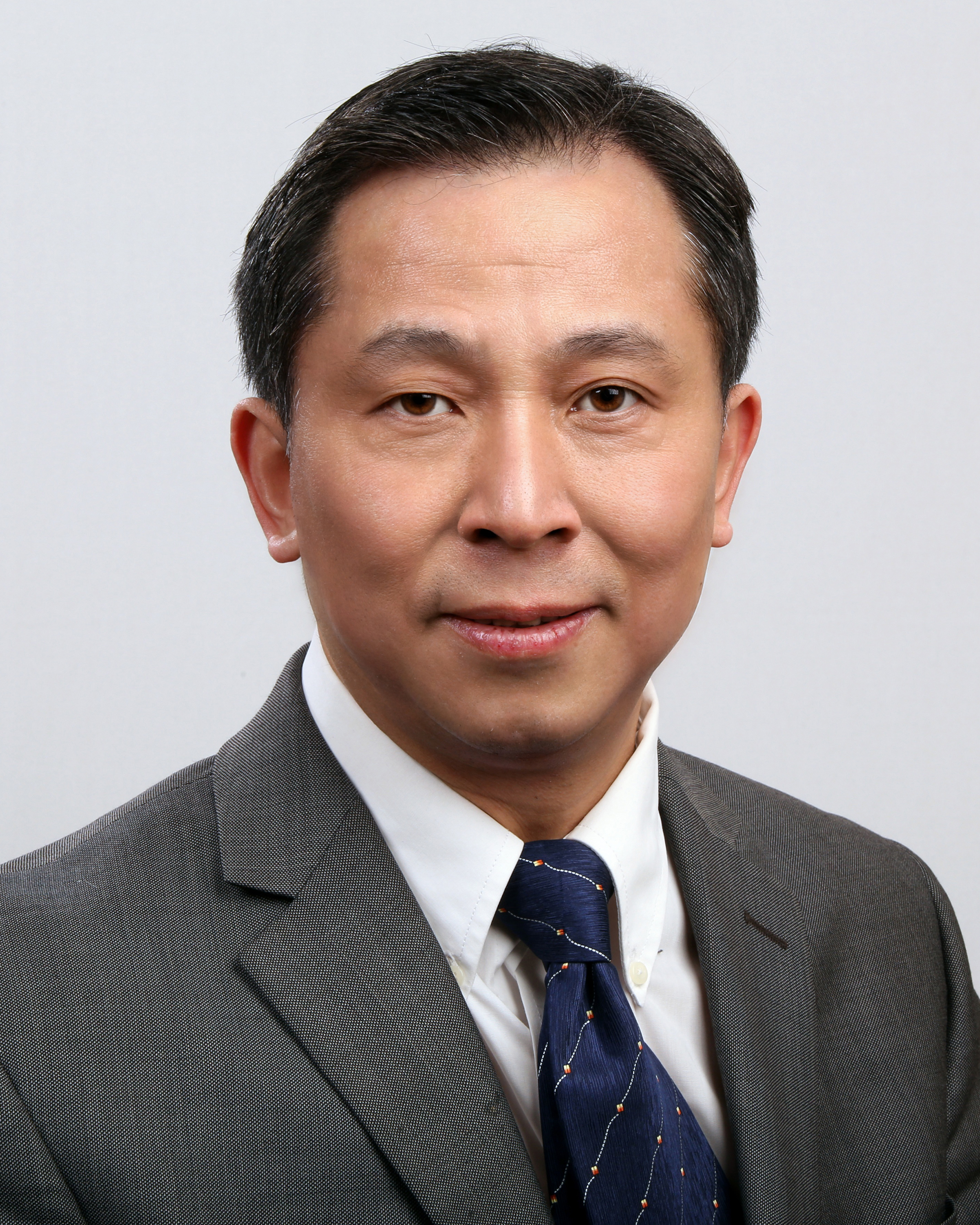}}]{Hui Xiong} (SM'07)
is currently a Full Professor at the Rutgers, the State University of New Jersey, where he received the 2018 Ram Charan Management Practice Award as the Grand Prix winner from the Harvard Business Review, RBS Dean’s Research Professorship (2016), the Rutgers University Board of Trustees Research Fellowship for Scholarly Excellence (2009), the ICDM Best Research Paper Award (2011), and the IEEE ICDM Outstanding Service Award (2017). He received the Ph.D. degree from the University of Minnesota (UMN), USA. He is a co-Editor-in-Chief of Encyclopedia of GIS, an Associate Editor of IEEE Transactions on Big Data (TBD), ACM Transactions on Knowledge Discovery from Data (TKDD), and ACM Transactions on Management Information Systems (TMIS). He has served regularly on the organization and program committees of numerous conferences, including as a Program Co-Chair of the Industrial and Government Track for the 18th ACM SIGKDD International Conference on Knowledge Discovery and Data Mining (KDD), a Program Co-Chair for the IEEE 2013 International Conference on Data Mining (ICDM), a General Co-Chair for the IEEE 2015 International Conference on Data Mining (ICDM), and a Program Co-Chair of the Research Track for the 2018 ACM SIGKDD International Conference on Knowledge Discovery and Data Mining. He is an IEEE Fellow and an ACM Distinguished Scientist.
\end{IEEEbiography}
\vspace{-1.2cm}
\begin{IEEEbiography}[{\includegraphics[width=1in,height=1.25in,clip,keepaspectratio]{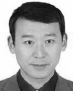}}]{Jian Yang} (M'08)
received the Ph.D. degree in pattern recognition and intelligence systems from the Nanjing University of Science and Technology
(NUST), Nanjing, China, in 2002. In 2003, he was a Post-Doctoral Researcher with the University of Zaragoza, Zaragoza, Spain. From
2004 to 2006, he was a Post-Doctoral Fellow with the Biometrics Centre, The Hong Kong Polytechnic University, Hong Kong. From 2006 to 2007, he was a Post-Doctoral Fellow with the Department of Computer Science, New Jersey Institute of Technology, Newark, NJ, USA. He is currently a Chang-Jiang Professor with the School of Computer Science and Engineering, NUST. He has authored more than 200 scientific papers in pattern recognition and computer vision. His papers have been cited more than 6000 times in the Web of Science and 15,000 times in the Scholar Google. His current research interests include pattern recognition, computer vision, and machine learning. Dr. Yang is a Fellow of IAPR. He is currently an Associate Editor of Pattern Recognition, Pattern Recognition Letters, the IEEE TRANSACTIONS ON NEURAL NETWORKS AND LEARNING SYSTEMS, and Neurocomputing.
\end{IEEEbiography}
\vspace{-1.2cm}

\end{document}
